\documentclass[10pt,journal]{IEEEtran}

\usepackage{ifpdf}
\usepackage{cite}
\usepackage{amsmath,amssymb,amsfonts}
\usepackage{algorithmic}
\usepackage{graphicx}
\usepackage{textcomp}
\usepackage{xcolor}
\usepackage{multirow}
\usepackage{float}
\usepackage{caption} 
\usepackage{subcaption}
\usepackage{siunitx}
\usepackage{mathtools}
\usepackage{longtable}
\usepackage[utf8]{inputenc}
\usepackage[pagebackref=true,breaklinks=true,colorlinks,bookmarks=false]{hyperref}
\hypersetup{
    colorlinks=true,
    filecolor=magenta,      
    urlcolor=magenta,
}

\def\BibTeX{{\rm B\kern-.05em{\sc i\kern-.025em b}\kern-.08em
    T\kern-.1667em\lower.7ex\hbox{E}\kern-.125emX}}

\ifCLASSOPTIONcompsoc
  \usepackage[nocompress]{cite}
\else
  \usepackage{cite}
\fi

\usepackage{amsmath}

\usepackage{array}

\begin{document}




\title{AttX: Attentive Cross-Connections for Fusion of Wearable Signals in Emotion Recognition}

\author{Anubhav Bhatti, Behnam Behinaein, Paul Hungler, Ali Etemad,~\IEEEmembership{Senior Member,~IEEE}
\IEEEcompsocitemizethanks{\IEEEcompsocthanksitem A. Bhatti, B. Behinaein, and A. Etemad are with the Department of Electrical and Computer Engineering and Ingenuity Research Labs, Queen's University, Kingston, Ontario, Canada.
	E-mails: \{anubhav.bhatti, 9hbb, ali.etemad\}@queensu.ca
	\IEEEcompsocthanksitem P. Hungler is with Ingenuity Research Labs, Queen's University, Kingston, Ontario, Canada.
	E-mails: {paul.hungler}@queensu.ca}
}

\IEEEtitleabstractindextext{%
\begin{abstract}
We propose cross-modal attentive connections, a new dynamic and effective technique for multimodal representation learning from wearable data. Our solution can be integrated into any stage of the pipeline, i.e., after any convolutional layer or block, to create intermediate connections between individual streams responsible for processing each modality. Additionally, our method benefits from two properties. First, it can share information uni-directionally (from one modality to the other) or bi-directionally. Second, it can be integrated into multiple stages at the same time to further allow network gradients to be exchanged in several touch-points. We perform extensive experiments on three public multimodal wearable datasets, WESAD, SWELL-KW, and CASE, and demonstrate that our method can effectively regulate and share information between different modalities to learn better representations. Our experiments further demonstrate that once integrated into simple CNN-based multimodal solutions (2, 3, or 4 modalities), our method can result in superior or competitive performance to state-of-the-art and outperform a variety of baseline uni-modal and classical multimodal methods. 
\end{abstract}

}

\maketitle

\IEEEdisplaynontitleabstractindextext

%
\IEEEpeerreviewmaketitle

\section{Introduction}
Affective computing is an emerging field of study for developing advanced computing systems that recognize, model, interpret, and respond to human affective states (e.g., moods and emotions) \cite{picard2003affective}. It can be defined as an interdisciplinary field that involves psychology, computer science, and biomedical engineering \cite{marin2018affective, calvo2010affect}. In general, affective computing systems enhance the quality of human-machine interaction by automatically recognizing and responding to user's emotional states, thereby making the machine interface more usable and effective. Human emotions have physiological fingerprints and can be recognized by analyzing bio-signals that can be monitored using wearables. As a result, several studies have focused on machine learning and deep learning to analyze physiological signals towards classification or quantification of human emotional states \cite{plarre2011continuous, sarkar2020self, healey2005detecting, zhang2022parse}. Recent applications of affect-sensitive systems include stress and anxiety detection \cite{ healey2005detecting, koldijk2014swell}, human monitoring systems \cite{taylor2017personalized, shashikumar2017deep}, healthcare (primarily mental health) \cite{koldijk2016detecting, sarkar2021detection}, education \cite{ruberto2021future}, adaptive gaming, marketing, etc. \cite{marin2018affective}.

The literature on affective computing have utilized various modalities to automatically classify emotional states such as stress. These modalities include electrocardiagram (ECG) \cite{agrafioti2011ecg, sarkar2020self}, electrodermal activity (EDA), also known as, galvanic skin response \cite{zhai2006stress}, voice \cite{kwon2003emotion}, posture and gait \cite{etemad2014classification}, facial expressions \cite{sepas2019deep, sepas2020facial}, electrooculogram (EOG) \cite{zhang2019capsule}, electroencephalogram (EEG) \cite{zhang2022parse, zhang2020rfnet}, respiration (RESP) \cite{plarre2011continuous}, and blood volume pulse (BVP) \cite{bota2020emotion}.
In such solutions, different deep learning models are often used to learn features from different modalities that are then fed to classifiers for predicting the emotion class. 

\begin{figure}[!t]
    \centering
    \centerline{\includegraphics[width=\columnwidth]{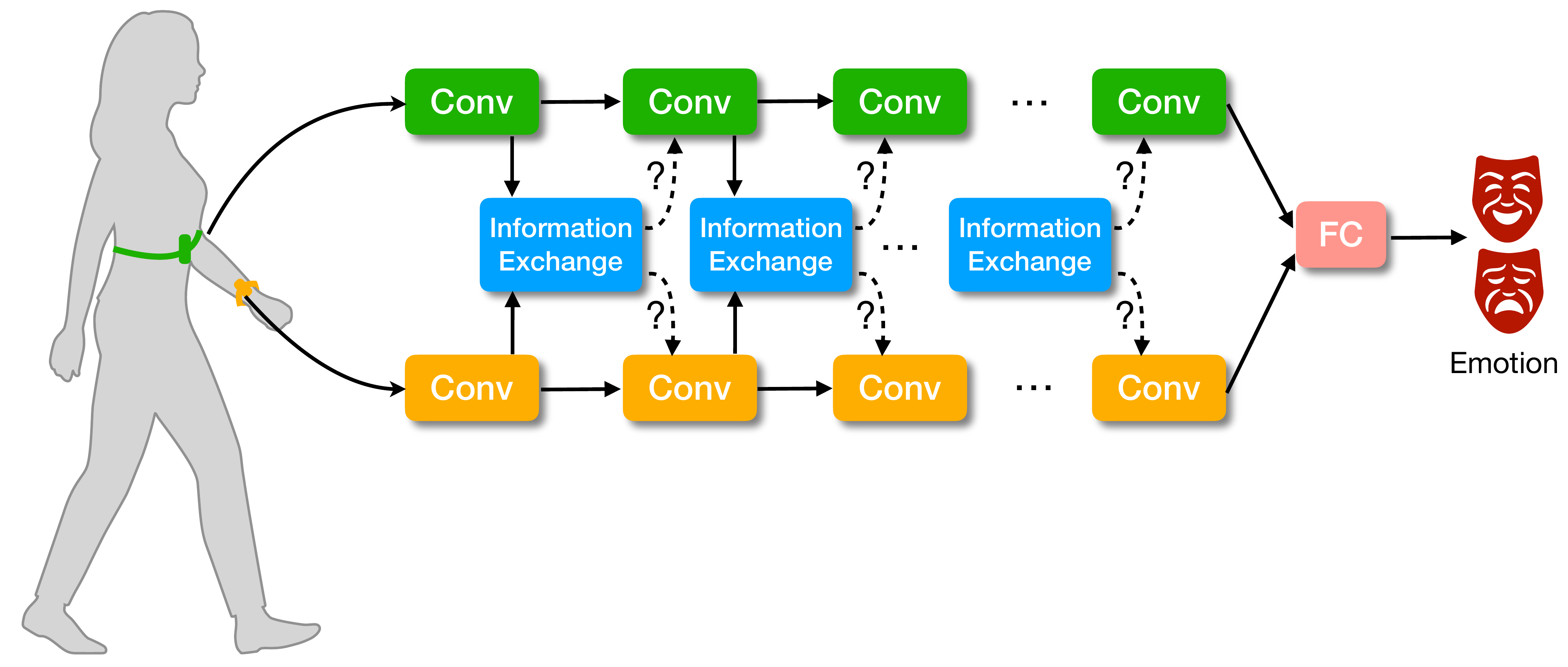}}
    \caption{An overview of our study is presented where attentive cross-modal connections are proposed, and their integration in multimodal settings is studied for emotion recognition.}
    \label{fig:1}
\end{figure}

In emotion recognition, different physiological modalities comprise different information that can be combined to boost classification performance. To enhance performance, the different modalities, their learned features, or the respective decisions made by individual machine learning models, are often fused. Traditionally, the different modalities are combined using two fusion techniques, i.e., feature-level fusion also referred to as early fusion, and decision-level fusion (or score-level fusion) also called late fusion. In feature-level fusion, high-level embeddings from different modalities, extracted by the respective feature extractors, are concatenated, and then fed to a classifier for discriminative tasks \cite{lin2019explainable}. In decision-level fusion, modalities are provided to their respective classifiers, and the decision of each classifier is combined to obtain a final decision. Combining multiple physiological signals for emotion recognition has been the focus of many recent studies \cite{plarre2011continuous, katsigiannis2017dreamer, ross2019toward, alzoubi2012detecting, das2016emotion, schmidt2018introducing}. Several studies also focus on using deep learning pipelines to extract high-level embeddings from the modalities and then fuse them to classify emotion classes \cite{siddharth2019utilizing, bota2020emotion, lin2019explainable, yang2019attribute, ross2020unsupervised}. These studies use handcrafted features or extract features automatically using deep learning and then use early or late fusion to fuse the features for emotion recognition.

Recent studies have shown that besides using early or late fusion, the learned representations from different hidden layers of the deep learning networks can also be combined via information-sharing branches referred to as \textit{cross-connections} \cite{velivckovic2016xcnn,cangea2019xflow}.
These cross-connections can help exploit the complementary information between the different modalities effectively, leading to better performance. However, the cross-connections are generally connected directly between the hidden layers to share the learned representations in these methods. As a result, these cross-connections are not capable of performing any additional processing on the shared information to enhance the information. Also, these works do not present exhaustive experiments to show the impact of adding these information-sharing cross-connections at different stages of the networks on the overall performance.

In this paper, we introduce a novel attentive cross-modal connection (AttX) comprising an attentive feedforward network to learn enhanced shared representations for multimodal affective computing. This allows the shared information to be weighted based on its importance before being shared between the available modalities. To control the flow of information, we define three types of AttX connections: 
Type I shares information from the first modality to the second; Type II shares information from the second modality to the first; and Type III simultaneously shares information between both modalities. In addition, we study the behaviour of overall model when AttX is integrated in different locations (referred to as stages) in our deep learning pipeline for emotion classification. A broad overview of our work is depicted in Figure \ref{fig:1}. We use three publicly available datasets to evaluate our proposed solution: WESAD \cite{schmidt2018introducing}, SWELL-KW \cite{koldijk2014swell}, and CASE \cite{sharma2019dataset}. We use the fusion of `ECG and EDA', `BVP and EDA', and `ECG, EDA, RESP, and skin temperature (ST)' to show the generalizability of our proposed method for enhancing multimodal fusion for wearable-based emotion recognition. Various experiments demonstrate the effectiveness of our method as models equipped with AttX outperform or obtain competitive results to uni-modal or other multimodal solutions.

Our contributions can be summarized as follows:
(\textbf{1}) We propose attentive cross-modal connections for sharing intermediate information between wearable modalities. To enable uni-directional or bi-directional information sharing between the modalities, we introduce three AttX connection types. Our proposed method can be integrated into different stages of deep learning pipelines and can successfully enhance the overall learned representations.
(\textbf{2}) We extensively test our solution on three popular and public multimodal datasets (WESAD, SWELL-KW, and CASE). We exploit leave-one-subject-out cross-validation and observe that simple backbones equipped with AttX outperform the state-of-the-art methods. We show the applicability of our proposed method on different pipelines comprising VGG and ResNet encoder blocks. We perform thorough experiments to show the impact of adding AttX connections when used in a single-AttX configuration (only integrated at one stage) and a multi-AttX configuration (integrated at multiple stages) on different networks.
(\textbf{3}) We perform analyses on the outcome of using different AttX configurations in the network to provide insights on the impact of sharing representations from one modality to another. Additionally, we show that the optimum stage to share information with Type I and Type II connections is generally halfway through the network. In contrast, Type III connections are more effective in learning better representations in later stages of the network. Further, when used in the contexts of learning multimodal ECG-EDA or BVP-EDA, our analysis shows that the sharing information from EDA to ECG or BVP is more beneficial than the opposite or the two-way connection (Type III). Lastly, our analysis shows that while single-AttX configurations across connection types and stages generally perform better than the multi-AttX configurations, the best performance of a network is often observed when a multi-AttX configuration is used simultaneously in the middle and end of the pipeline.

This paper is an extension of our work presented in \cite{bhatti2021attentive}, a \textit{ACII 2021 Workshop} paper, compared to which this work comprises the following additions: (\textit{i}) We add experiments for a 3-class classification task (neutral vs. amusement vs. stress) in WESAD dataset. (\textit{ii}) We also use two additional publicly available datasets in our experiments, SWELL-KW (stress classification) and CASE (arousal classification). (\textit{iii}) We show the application of AttX connections on two pipelines with different encoders: VGG and ResNet. (\textit{iv}) We explore the fusion of more than two modalities using our proposed cross-modal connections. (\textit{v}) Further, we perform an in-depth analysis to gain insights into selecting the direction and the optimum location for sharing intermediate information between modality streams.

\section{Background and Related Work}
\label{sec:relatedwork}
\subsection{Uni-modal Affective Computing}
Recent works in the field of affective computing have studied the use of machine learning and deep learning to determine the emotional state of subjects from physiological data. Here we review some of the works that specifically use physiological signals such as ECG, EDA, BVP, RESP, and ST. Ferdinando et al. \cite{ferdinando2016comparing} proposed the use of statistical distribution of dominant frequencies, calculated through spectrogram analysis. They then used k-nearest neighbours (KNN) for the classification of emotions in arousal and valence space in a multi-class setting. Hwang et al. \cite{hwang2018deep} proposed a deep learning framework to monitor stress using ECG. They introduced a method to design a deep learning architecture based on periodic patterns of raw ECG signals. 
In \cite{sarkar2019classification}, Sarkar et al. used ECG to classify participants' level of cognitive load and expertise using a multi-task neural network in the context of dynamically adaptive simulation. 
Sarkar and Etemad \cite{sarkar2020self, sarkar2020self_icassp} proposed a self-supervised method for classifying arousal and valence based on ECG signals. In this method, generalized representations of unlabelled ECG were learned by training a multi-task convolutional neural network (CNN) to recognize transformations applied to the input signals as pretext tasks. This was followed by transfer learning for downstream supervised classification. Behinaein et al. \cite{behinaein2021transformer} proposed a transformer mechanism to detect stress using ECG signals in two publicly available datasets. In this study, the deep learning network comprises a convolutional subnetwork, a transformer encoder, and a fully connected subnetwork for stress classification.

EDA has also been used for the classification of emotions, notably stress. Hsieh et al. \cite{hsieh2019feature} used EDA and extracted time, frequency, entropy, and wavelet domain features for detecting stress using XGBoost. In a work by Setz et al. \cite{setz2009discriminating}, time-domain features of EDA signals were used to discriminate stress from cognitive load in a laboratory environment using standard machine learning algorithms, e.g., linear discriminant analysis (LDA), KNN, and support vector machines (SVM). Recently, Aqajari et al. \cite{aqajari2020gsr} developed a tool for the analysis of EDA, which used deep learning along with statistical algorithms to extract features for stress detection.

\subsection{Multimodal Affective Computing}
The use of multimodal approaches for classification of affective states using machine learning methods has also been a focus of a number of studies in recent times. Plarre et al. \cite{plarre2011continuous} proposed two models for continuous prediction of stress from physiological signals. Handcrafted features from ECG and RESP were used for detecting stress and other psychologically and physically demanding conditions. Stamos et al. \cite{katsigiannis2017dreamer} used handcrafted features from ECG and EEG such as power spectral density and heart rate variability (HRV) to classify emotions using SVM. Ross et al. \cite{ross2019toward} used ECG and EDA features along with feature-level fusion to perform multimodal classification of two levels of expertise, i.e., expert and novice, using several machine learning models. 
AlZoubi et al. \cite{alzoubi2012detecting} relied upon handcrafted features from ECG, electromyography (EMG), and EDA, and used KNN and linear Bayes classifier to determine boredom, confusion, and curiosity. The work done by Das et al. \cite{das2016emotion} relied on frequency-domain features of ECG and EDA for the classification of happy, sad, and neutral states. Their work showed that the multimodal features with an SVM classifier perform better than the uni-modal features. Schmidt et al. \cite{schmidt2018introducing} experimented on different models to perform binary classification of stress vs. non-stress states and multi-class classification of stress, amusement, and neutral states. For classification, their method relied on handcrafted statistical features, HRV, and frequency domain features from ECG, EDA, EMG, RESP, and ST. Mohammadi et al. \cite{mohammadi2022integrated} use Kruskal-Wallis analysis on 65 features extracted from ECG, EDA, RESP, and ST to identify features that demonstrate significant difference between stress and relaxed states and then used KNN to distinguish these states.

A work by Siddharth et al. \cite{siddharth2019utilizing} used a combination of manually extracted features and deep representations from ECG, EDA, and EEG. An extreme learning machine then used the concatenated features to classify arousal, valence, liking, and emotions. Bota et al. \cite{bota2020emotion} performed a multimodal classification of emotions in the arousal and valence space using feature-level and decision-level fusion of physiological data (ECG, EDA, RESP, and BVP). Lin et al. \cite{lin2019explainable} introduced a multimodal-multisensory sequential fusion model to detect three affect states. The proposed fusion model was trained on different modalities with different sampling rates in the same training batch. Li et al. \cite{li2020stress} proposed a one-dimensional (1D) CNN to automatically extract features from ECG, EDA, EMG, RESP, and ST and perform binary classification of stress versus non-stress state and multi-class classification of stress, amused, and neutral state. Bacciu et al. \cite{bacciu2021benchmarking} benchmarked recurrent neural networks on human state and activity recognition tasks. This work showed that echo-state network models’ performance on human state and activity recognition is comparable to other recurrent models and can serve as an alternative for implementing predictive models on low-powered devices. Samyoun et al. \cite{samyoun2020stress} presented a solution to detect stress using physiological data from wrist sensors that emulate the standard chest sensors. In this work, the data from wrist sensors is translated into the data from chest sensors using a generative adversarial network, recurrent neural network, or multilayer perceptron based translation model. The translated data is then used for stress detection without requiring the users to wear any device on the chest.

Yang et al. \cite{yang2019attribute} proposed a variational autoencoder to learn personality-invariant physiological signal representations for ECG, EDA, and EEG. Arousal levels were then classified using an SVM by exploiting these latent representations. Ross et al. \cite{ross2020unsupervised} proposed a method to use multi-corpus wearable data to learn multimodal representations in an unsupervised framework using auto-encoders followed by supervised classifiers. Yin et al. \cite{yin2017recognition} introduced an ensemble-based multiple-fusion-layer classifier of stacked auto-encoders for emotion classification in arousal and valence space using multiple physiological signals. They used handcrafted features from EEG, electrooculography, EMG, ST, EDA, BVP, and RESP signals to classify emotions.


In the end, we observe that in most of the prior works on multimodal representation learning for affective computing, either feature-level or decision-level fusion strategies have been explored. More specifically, the notion of cross-modal connections for sharing information between the models responsible for learning each modality has not been studied in this area.

\begin{figure*}[h]
    \centering
        \includegraphics[width=1\textwidth]{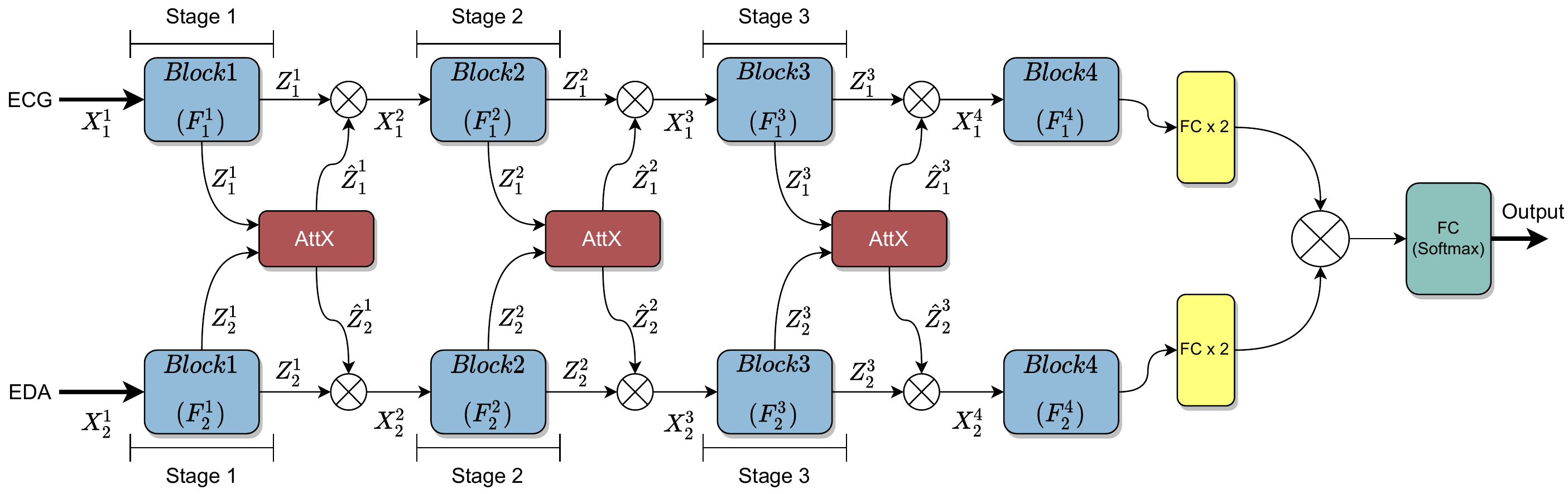}
        \caption{A multimodal pipeline consisting of two CNNs, one for learning ECG and the other for learning EDA, is presented. The CNNs are connected via our AttX connections (Type III) after each convolutional block. $\bigotimes$ denotes concatenation.}
        \label{fig:fig_sim}
\end{figure*}

\begin{figure*}[h]
    \begin{center}
     \begin{subfigure}[b]{.3\textwidth}
         \includegraphics[width=.95\textwidth]{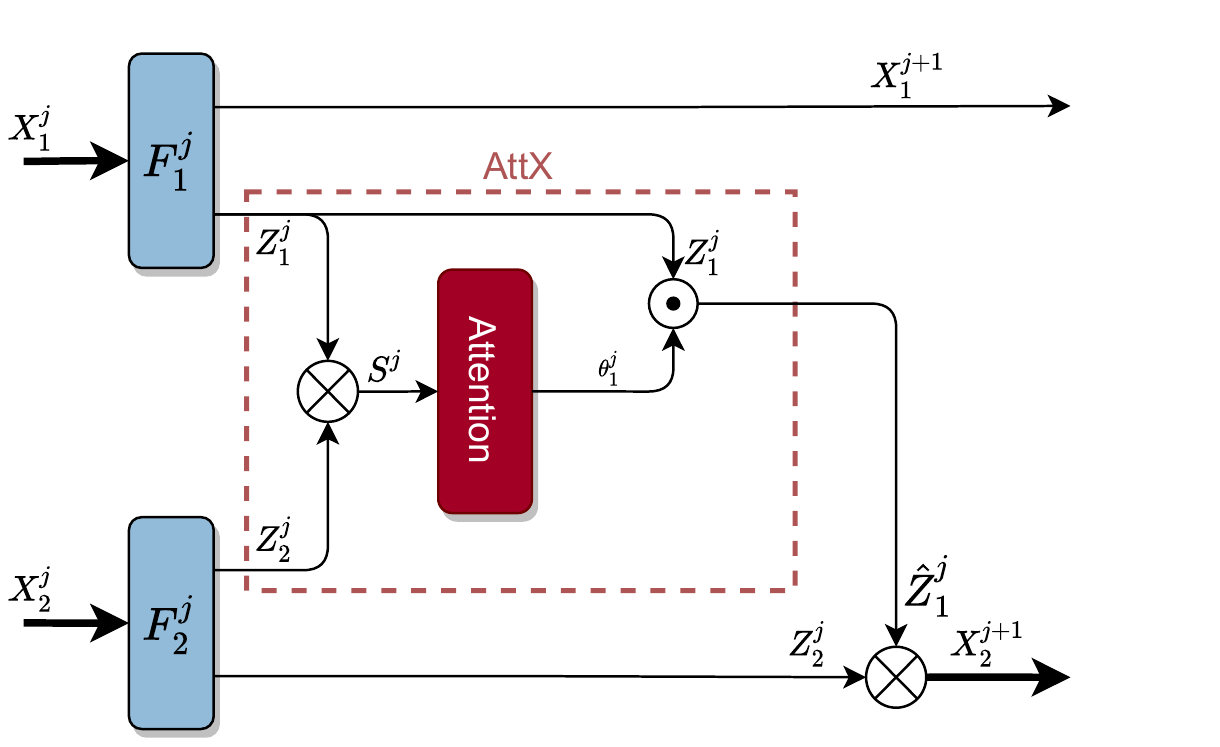}
         \caption{Type I}
         \label{fig:typI}
     \end{subfigure}
     \begin{subfigure}[b]{.3\textwidth}
         \includegraphics[width=.95\textwidth]{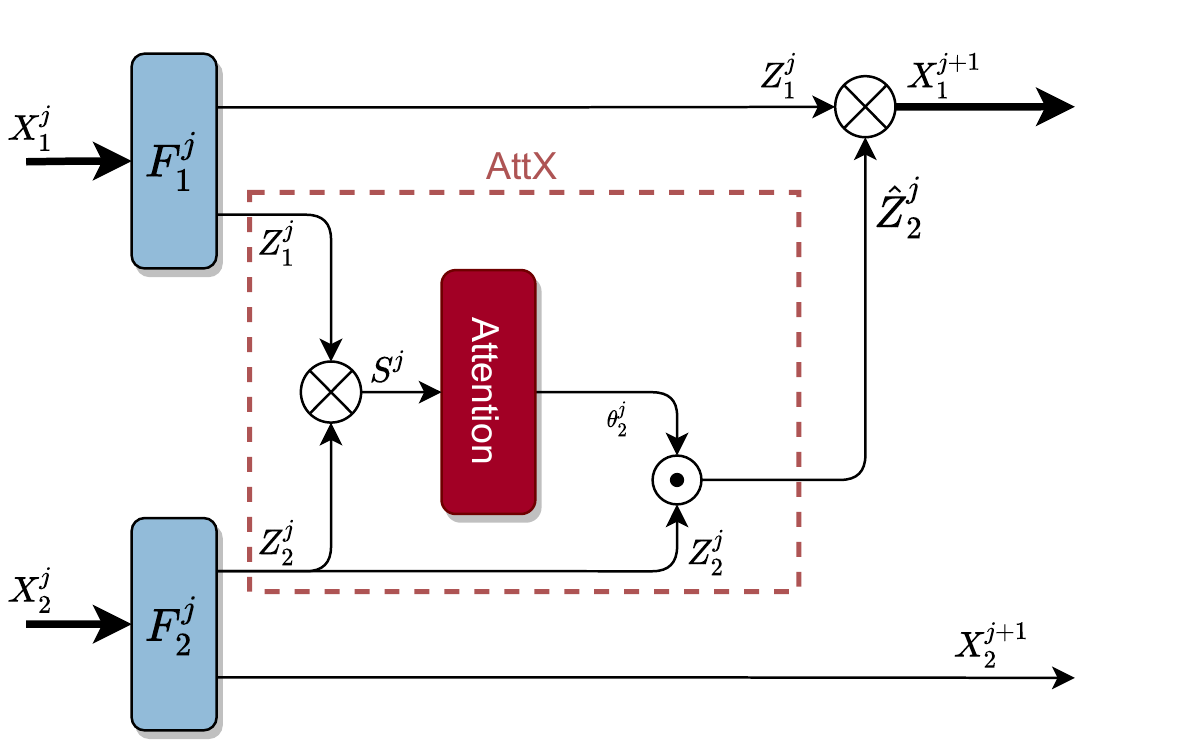}
         \caption{Type II}
         \label{fig:typII}
     \end{subfigure}
     \begin{subfigure}[b]{.3\textwidth}
         \includegraphics[width=.95\textwidth]{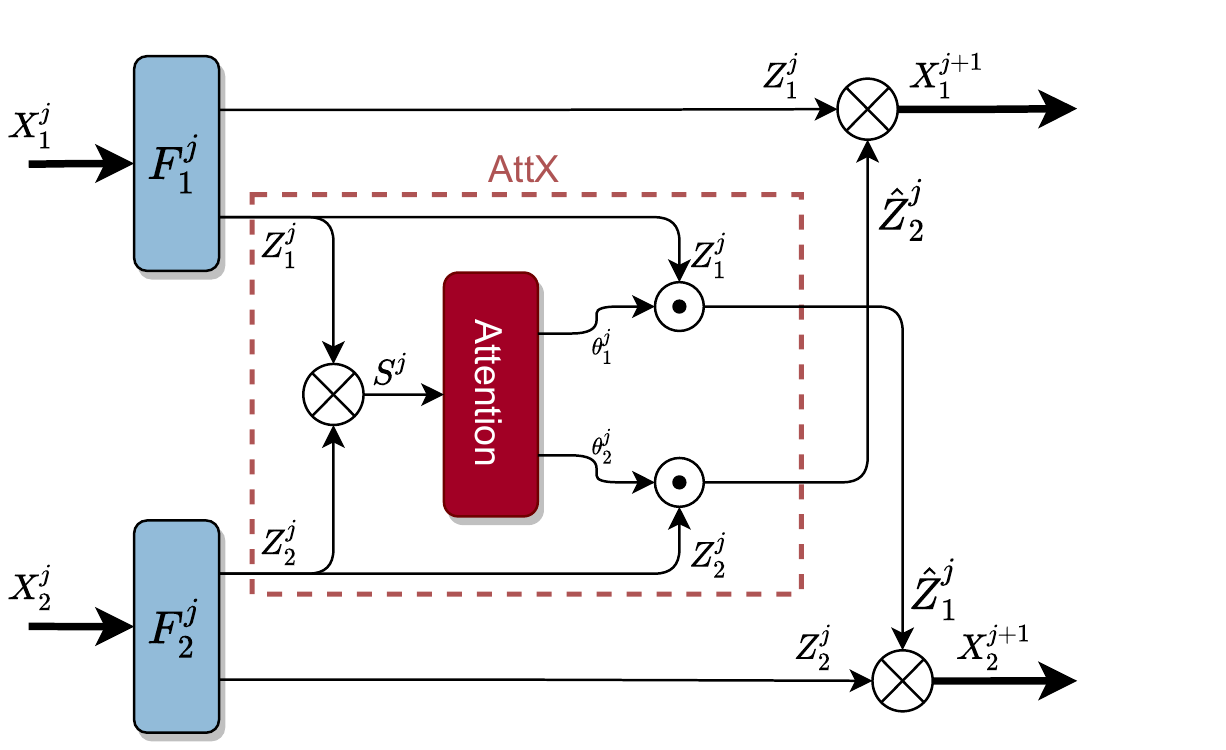}
         \caption{Type III}
         \label{fig:typIII}
     \end{subfigure}     
     \caption{The three configurations for the proposed attentive cross-modal connections are presented: Type I: ECG to EDA, Type II: EDA to ECG, and Type III: ECG to EDA and EDA to ECG. $\bigotimes$ denotes concatenation, and $\bigodot$ represents element-wise multiplication.}
     \label{fig:types}
    \end{center}
\end{figure*}

\section{Method}
\label{sec:methodology}
\subsection{Problem Statement}
Assume we have two signal modalities, for instance recorded by a wearable device, which we denote by $X_1$ and $X_2$. Here $X_1 \in \mathbb{R}^{m_1}$ and $X_2 \in \mathbb{R}^{m_2}$, where $m_1$ and $m_2$ are the dimensionality of the modalities.
Further, let's assume we have two separate encoders, $F_1$ and $F_2$, which we use for learning representations from $X_1$ and $X_2$, respectively. Each encoder $F_i$ consists of a number of individual convolutional blocks (also henceforth referred to as `\textit{stages}'), denoted by $F^j_i$, where $j$ is the index for a given stage. Our goal in this work is to design cross-modal connections $\Phi$ capable of exchanging information between $F_1$ and $F_2$. We assume $\Phi$ can be integrated between $F^j_1$ and $F^k_2$, where for simplicity, $j=k$. Accordingly, the key research questions in our work are as follows:\\
(\textbf{1}) How should $\Phi$ be designed such that the flow of information can be automatically regulated between $F_1$ and $F_2$?\\
(\textbf{2}) How does $\Phi$ impact the overall performance in comparison to uni-modal learning, standard feature-level fusion, and score-level fusion?\\
(\textbf{3}) What are the best configurations of $\Phi$ for integration into the multimodal $F_1 - F_2$ architecture? In other words, should $\Phi$ share information from $F1$ to $F2$ (given a specific set of modalities), vice-versa, or in both directions? Moreover, in which stages of the overall pipeline should $\Phi$ be integrated, i.e., what is the optimum set of $j$ in a $F^j_1, F^j_2$ setup?

In the following subsection, we design $\Phi$ with the capability of automatically learning to regulate the sharing of information. Following that, we design and conduct a set of experiments to address research questions 2 and 3.

\subsection{Proposed Solution}
\label{subsec:attx}
We propose $\Phi$ as attentive cross-modal connections to tackle the above-mentioned problems, henceforth referred to as AttX for simplicity. Let's define the output of each encoder block $F^j_i$ as $Z^j_i$. Accordingly, we define AttX as a feedforward network that takes $Z^j_1$ and $Z^j_2$ as its inputs and learns weighted intermediate representations $\hat{Z}^j_1$ and $\hat{Z}^j_2$, which are then shared between the respective encoder blocks. An overview of AttX integrated into a deep learning pipeline is depicted in Figure \ref{fig:fig_sim}.
First, we define AttX with the aim of sharing information from the first modality ($X_1$) to ($X_2$). Accordingly, the inputs to the AttX block at stage $j$, which is integrated between $Z^j_1$ and $Z^j_2$, are concatenated to obtain an intermediate representation tensor, $S^j$:
\begin{align}
    S^{j} &= \begin{bmatrix}
                Z^j_1 \\  
                Z^j_2     
\end{bmatrix},~\text{where}~~ S^{j} \in \mathbb{R}^{n \times m \times d},
\end{align}
where $d$ represents the number of input modalities, in our case $d$ = 2, and $n$ and $m$ represent the dimensions of $Z^j_1$ and $Z^j_2$. 
In case the dimensions of $Z^j_1$ and $Z^j_2$ are not equal, pre-processing steps can be made to modify the dimensionality such that equal dimensions are achieved.

Subsequently, the intermediate representation tensor $S^{j}$ is fed to an attention block comprising a feedforward network with a hidden layer and an activation layer to obtain a projection of the tensor $S^{j}$ \cite{vaswani2017attention, poria2017multi}. Thus,
\begin{align}
  &U^j = ReLU(S^j W^j) ,
  \label{equ:projection}
\end{align}
where the learned weight matrix $W^{j} \in \mathbb{R}^{d\times d}$, and $U^{j}$ is the projection of the intermediate representation tensor $S^{j}$.

To obtain the attention weights $\theta^j$, a softmax function is applied to $U^{j}$. The attention weights are computed according to 
\begin{align}
  &\theta^j = softmax({(U^j)}^{T(2,3)} w^j_u),
  \label{equ:attnwgt}
\end{align}
where $w^{j}_{u} \in \mathbb{R}^{m}$ represents the learned weight vector, and softmax() computes the softmax of the dot product of the transposed projection of $U^{j}$ and the learned weight vector $w^{j}_{u}$ along the second axis. Here, the learnt attention weight tensor $\theta^j \in \mathbb{R}^{n \times d \times m}$. 
The transpose of the second and third dimensions of the projection $U^j$ is indicated by ${U^j}^{T(2,3)}$.

To obtain the weighted intermediate representations from $F^j_1$, we extract the attention weights $\theta^j_1$, corresponding to $F^j_1$, from the learned attention weight tensor $\theta^j$. Given $d = 2$, as we assume only two modalities (e.g., ECG and EDA), the $\theta^j_1$ would be:
\begin{align}
    \theta^j_1 &= \theta^{j}_{*, 1, *} \in \mathbb{R}^{n\times m}. 
    \label{equ:theta_1}
\end{align}
Further, the attention weight tensor for $\theta^j_1$ is multiplied by $Z^j_1$, i.e., the outputs of encoder blocks $F^j_1$, to obtain weighted intermediate representations $\hat{Z}^j_1$:
\begin{align}
 \label{equ:Zhat_1}
  \hat{Z}^j_1 &= \theta^j_1 \odot Z^j_1, 
\end{align}
where $\odot$ denotes the element-wise multiplication. In Eq. \ref{equ:Zhat_1}, $\hat{Z}^j_1$ indicates the output representations from the attention block weighted by its respective attention weight tensor $\theta^j_1$.

The weighted intermediate representations, $\hat{Z}^j_1$, obtained by Eq. \ref{equ:Zhat_1}, is combined with the output representations $Z^j_2$ to generate input, $X^{j + 1}_2$, for the next encoder blocks in the pipeline, i.e., $F^{j + 1}_2$. This input $X^{j + 1}_2$ is obtained according to:
\begin{align}
    \label{equ:ecg_to_eda}
  X^{j + 1}_2 &= Z^j_2 \otimes \hat{Z}^j_1 ,
\end{align}
where $\otimes$ denotes the concatenation operation. Figure \ref{fig:types} (left) presents the final architecture of AttX which shares information from $X_1$ to $X_2$, referred to as Type I.

Similar to this approach, we also define another type of AttX in which information is shared from the second modality to the first modality ($X_2$ to $X_1$). To do so, the entire process remains the same until Eq. \ref{equ:theta_1}, which will be modified as: 
\begin{align}
    \theta^j_2 &= \theta^{j}_{*, 2, *} \in \mathbb{R}^{n\times m} .
    \label{equ:theta_2}
\end{align}
Similarly, in order to provide the intermediate weighted representations for $X^j_2$, Eq. \ref{equ:Zhat_1} is replaced with:
\begin{align}
  \hat{Z}^j_2 &= \theta^j_2 \odot Z^j_2.
 \label{equ:Zhat_2}
\end{align}
Finally, instead of Eq. \ref{equ:ecg_to_eda}, the weighted intermediate representations, $\hat{Z}^j_2$, are combined with the output representations $Z^j_1$ to generate the input $X^{j + 1}_1$, as follows:
\begin{align}
  X^{j + 1}_1 &= Z^j_1 \otimes \hat{Z}^j_2.
  \label{equ:eda_to_ecg}
\end{align}
The architecture of this type of AttX, which is capable of sharing information from $X_2$ to $X_1$ is illustrated in Figure \ref{fig:types} (middle), which we refer to as Type II.

In addition to Type I and Type II AttX connections, we also define a third type of attentive cross-modal connections (Type III), for sharing information simultaneously between both modalities (from $X_1$ to $X_2$ and from $X_2$ to $X_1$). In this type of AttX, Eqs. \ref{equ:theta_1} and \ref{equ:theta_2} are used together to extract the attention weights for both modalities, $\theta^j_1$ and $\theta^j_2$ . Next, Eqs. \ref{equ:Zhat_1} and \ref{equ:Zhat_2} are used simultaneously to compute the weighted intermediate representations, $\hat{Z}^j_1$ and $\hat{Z}^j_2$. Lastly, Eqs. \ref{equ:ecg_to_eda} and \ref{equ:eda_to_ecg} are used to combine $\hat{Z}^j_2$ and $\hat{Z}^j_1$ with $Z^j_1$ and $Z^j_2$ to generate $X^{j + 1}_1$ and $X^{j + 1}_2$. The architecture of Type III connections is presented in Figure \ref{fig:types} (right).

\vspace{5pt}

\noindent \textbf{Beyond Two Modalities.}
We expand our AttX connections for integration into pipelines with more than two modalities, i.e., $d > 2$. Accordingly, the intermediate representation tensor $S^j$ can be denoted as
\begin{align}
    S^{j} &= \begin{bmatrix}
                    Z^j_1 \\
                    Z^j_2 \\
                    \vdots \\
                    Z^j_d
            \end{bmatrix}.
\end{align}
Thus the attention weight tensor, ${\theta^j}^{T(2,3)}$ (where the transpose of the second and third dimensions is represented by $T(2,3)$) in Eq. \ref{equ:attnwgt}, includes attention weights for $d$ modalities and is represented as:
\begin{align}
    {\theta^j}^{T(2,3)} &= \begin{bmatrix}
                    \theta^j_1 \\
                    \theta^j_2 \\
                    \vdots \\
                    \theta^j_d
            \end{bmatrix},~\text{where}~~ \theta^j_1, \theta^j_2, ..., \theta^j_d \in \mathbb{R}^{n \times m}.
\end{align}
Further, intermediate weighted representations $\hat{Z}^j_1$, $\hat{Z}^j_2$, …, $\hat{Z}^j_d$ can be computed from their respective attention weights $\theta^j_1$, $\theta^j_2$, …, $\theta^j_d$ using the Eq. \ref{equ:Zhat_1} or \ref{equ:Zhat_2}. 
These computed intermediate weighted representations can be combined with the output representation of the other modalities as follows:
\begin{align}
  X^{j + 1}_i &= Z^j_i \otimes \hat{Z}^j_k ,
  \label{equ:d_mods}
\end{align}
where $1 \le i \le d$ and $k=\{1,...,d\} - \{i\}$.

Here, there can be $2^d-1$ types of AttX connections for sharing information between $d$ modalities. For instance, as shown already for $d$ = 2, we can have three types of connections, i.e., Type I (from $X_1$ to $X_2$), Type II (from $X_2$ to $X_1$), and Type III (from $X_1$ to $X_2$ and from $X_2$ to $X_1$). For three or more modalities, i.e., $d > 2$, we follow a greedy approach to reduce the search space to obtain an optimum AttX type. In this approach, we first attain the best AttX type for two modalities and only explore the subset of that type in combination with other modalities. For example, for $d = 3$, there can be seven different combinations in which we can share information between the three modalities. However, assuming that the best connection is Type II for two modalities, then using the greedy approach, we reduce the search space to just two combinations, i.e., from $X_2$ to $X_1$ \& $X_3$, and $X_2$ \& $X_3$ to $X_1$. 

{\renewcommand{\arraystretch}{1.1}
\begin{table} 
\begin{center}
\caption{The architectural details (filter size, no. of filters, and stride) of the network for both VGG and ResNet variants are presented.}
\label{tbl:two_class_archt}
\scriptsize
\begin{tabular}{|c|c|c|} 
\hline
\multirow{2}{*}{\textbf{Block No.}} & \multicolumn{2}{|c|}{\textbf{Backbone Architecture}} \\ \cline{2-3}
 & \multirow{1}{*}{VGG Blocks} & \multirow{1}{*}{ResNet Blocks} \\ \hline
\multirow{3}{*}{Block 1} & \multirow{3}{*}{$\begin{array}{cc}
    \text{Conv1D, 64, 32, 1}\\
    \text{Conv1D, 64, 32, 3}\\
\end{array}$} & 

\multirow{3}{*}{$\left.\begin{array}{cc}
	 \text{Conv1D, 1, 32} \\
	 \text{Conv1D, 64, 32}\\
	 \text{Conv1D, 1, 64} \\
	 \end{array}\right\rbrace\times{2}, \text{s = [7, 1]} $} \\
& & \\
& & \\ \hline
\multirow{3}{*}{Block 2} & \multirow{3}{*}{$\begin{array}{cc}
    \text{Conv1D, 32, 64, 1}\\
    \text{Conv1D, 32, 64, 3}\\
\end{array}$} & 

\multirow{3}{*}{$\left.\begin{array}{cc}
	 \text{Conv1D, 1, 64} \\
	 \text{Conv1D, 32, 64}\\
	 \text{Conv1D, 1, 128} \\
	 \end{array}\right\rbrace\times{2}, \text{s = [3, 1]} $} \\
& & \\
& & \\ \hline

\multirow{3}{*}{Block 3} & \multirow{3}{*}{$\begin{array}{cc}
    \text{Conv1D, 17, 128, 1}\\
    \text{Conv1D, 17, 128, 3}\\
\end{array}$} & 

\multirow{3}{*}{$\left.\begin{array}{cc}
	 \text{Conv1D, 1, 128}\\
	 \text{Conv1D, 17, 128}\\
	 \text{Conv1D, 1, 256}\\
	 \end{array}\right\rbrace\times{2}, \text{s = [3, 1]} $} \\
& & \\
& & \\ \hline

\multirow{3}{*}{Block 4} & \multirow{3}{*}{$\begin{array}{cc}
    \text{Conv1D, 7, 256, 1}  \\
    \text{Conv1D, 7, 256, 3}\\
\end{array}$} & 

\multirow{3}{*}{$\left.\begin{array}{cc}
	 \text{Conv1D, 1, 256} \\
	 \text{Conv1D, 7, 256}  \\
	 \text{Conv1D, 1, 512} \\
	 \end{array}\right\rbrace\times{2}, \text{s = [3, 1]} $} \\
& & \\
& & \\ \hline
\end{tabular}
\end{center}
\end{table}}

\subsection{Integration}
Our proposed AttX connections can be integrated into pipelines with different backbone architectures responsible for processing each individual modality (see Figure \ref{fig:fig_sim}). In this paper, we explore two commonly used CNN architecture types as the backbones for multimodal learning. For bi-modal representation learning, we design a pipeline with two branches, one for each modality. We integrate an encoder in each branch followed by FC layers, where the encoders consist of a number of individual encoder blocks. We keep the number of blocks equal between the two branches to reduce the complexity when studying all the possible integration strategies of AttX in the pipeline. Empirically, we find that four encoder blocks in each branch yield strong results. We use VGG- and ResNet-like encoder blocks as two different architectures to thoroughly understand the effect of AttX connections on the performance and the learnt embeddings. We integrate AttX connections in various combinations, i.e., Type I, II, and III at stages 1, 2, and 3 (please see Figure \ref{fig:fig_sim}). 

For the VGG-like encoder blocks, we use two 1D CNNs with ReLU activation and a MaxPool layer of filter size 2 and stride 2. For ResNet-like encoders, each encoder block comprises a pair of residual blocks. These residual blocks have three consecutive 1D CNNs followed by a Batch Normalization layer and ReLU activation. Encoder blocks ($F^j_1$ and $F^j_2$) at different stages (i.e., $j = $ 1, 2, 3, and 4) have a different number of filters, filter sizes, and strides; however, $F^j_1$ and $F^j_2$ are kept the same for simplicity as well as a fair comparison between the different modalities under consideration. The details of each encoders are presented in Table \ref{tbl:two_class_archt}.

The outputs of the fourth encoder block ($F^4_1$ and $F^4_2$) are fed to two fully-connected layers before merging. As we will describe later, three popular public datasets (WESAD, SWELL-KW, and CASE) are used to evaluate our work (the details are described in Section \ref{sec:datasets}). For WESAD and SWELL-KW, two fully connected layers after the fourth encoder have 512 and 256 units, while for CASE, these two layers have 256 and 64 units. Finally, we add a classifier with SoftMax activation to generate the output class.

\section{Experiments}
\label{sec:experiments}
\subsection{Datasets}
\label{sec:datasets}
To evaluate our proposed AttX connections on different emotion recognition tasks under different data collection settings, we use three publicly available multimodal datasets, WEarable Stress and Affect Detection (WESAD) \cite{schmidt2018introducing}, SWELL Knowledge Work (SWELL-KW) \cite{koldijk2014swell}, and Continuously Annotated Signals of Emotion (CASE) \cite{sharma2019dataset}. The details of these datasets are provided in the sections below.

\subsubsection{WESAD}
WESAD is a multimodal dataset for stress and affect detection using wearable devices \cite{schmidt2018introducing}. The dataset comprises physiological and motion data collected from two sensors worn on the wrist and chest from 15 participants. A RespiBAN Professional\footnote{https://www.pluxbiosignals.com/collections/biosignalsplux} sensor, worn on the chest, is used to collect ECG, EDA, EMG, ST, 3-axis accelerometer (ACC), and RESP, at a sampling rate of 700 Hz. Three-electrode ECG is recorded from the chest, while the EDA is recorded from the rectus abdominis. An Empatica E4\footnote{https://www.empatica.com/en-int/research/e4/}, worn on the non-dominant hand of the participants, is used to collect EDA (4 Hz), BVP (64 Hz), ST (4 Hz), and ACC (32 HZ). For our work, we only consider ECG, EDA, RESP, and ST from the chest-based sensors. The dataset contains three different affect states, namely, neutral, stress, and amusement. In the study, a baseline of 20 minutes was recorded at the beginning to induce a neutral affective state, where participants read neutral reading material, e.g., magazines. To induce the amusement condition among the participants, the participants watched eleven funny videos followed by a short neutral video of five seconds. The amusement condition lasted for approximately 6 minutes and 30 seconds. The participants performed public speaking and a mental arithmetic task for 10 minutes to simulate stress conditions. A guided meditation session to bring participants back to a neutral affective state was introduced to transition from amusement to stress conditions or vice versa. Positive and Negative Affect Schedule \cite{muaremi2013towards} scheme was used to collect the ground truths labels for the affect states at the end of each trial.

\subsubsection{SWELL-KW}
The SWELL Knowledge Work dataset allows the study of stress and user modelling in a typical office environment \cite{koldijk2014swell}. The dataset is collected from 25 participants that performed typical knowledge work such as writing reports, making presentations, reading emails, and searching for information. During the experiment, the working condition of the participants is manipulated with stressors: email interruptions and time-pressure. Participants are allowed to work on multiple tasks under normal conditions for 45 minutes. For the time-pressure session, the duration to perform similar tasks is reduced to 30 minutes. In the interruption session, the participants are asked to respond to incoming emails to distract them from their usual tasks. At the start of each experiment session, an eight-minutes relaxation period is recorded in which the subject watches nature videos. Physiological modalities such as ECG, EDA, and others are recorded. ECG signals are collected using a TMSI Mobi \footnote{https://www.tmsi.com/products/} device with self-adhesive electrodes, while EDA signals are collected using a Mobi device with finger electrodes. Both ECG and EDA are recorded at a sampling frequency of 2048 Hz. Self-reported affect scores are recorded on a scale of 1 to 9 at the end of each scenario.

\subsubsection{CASE}
The CASE dataset focuses on the real-time continuous annotation of emotions experienced by participants while watching videos \cite{sharma2019dataset}. The dataset contains eight physiological signals and annotation data from 30 participants (15 male and 15 female). These eight physiological signals, ECG, EDA, BVP, EMG, RESP, and ST, were collected at 1000 Hz. The authors developed a joystick-based annotation interface to facilitate real-time annotations for simultaneous reporting of arousal and valence. The videos were selected to elicit amused, bored, relaxed, and scared emotional states. During the experiment, participants first watched a calming video at the session's start and at the end, i.e., the cool-down phase. A blue-screen video was shown when switching from one emotional video to another. The dataset provides arousal/valence ratings on nine levels.

\subsection{Data Pre-processing}
For this work, we consider the following physiological signals for classifying emotion states: ECG, EDA, BVP, RESP, and ST. We apply basic pre-processing steps to filter the raw data. Usually, ECG signals contain EMG noise, powerline noise, baseline wander, T-wave interference, and other artifacts. To remove these artifacts, we apply a Butterworth bandpass filter with a passband frequency of 5-15 Hz \cite{pan32real}. The filtered ECG signals are normalized using user-specific z-score normalization. We remove high-frequency noise from raw EDA signals by applying a lowpass filter with a cut-off frequency of 3 Hz. We use a Butterworth filter with a passband frequency of 0.5 Hz and a stopband frequency of 8 Hz for filtering raw BVP. For the raw RESP, we use a bandpass filter with a passband frequency of 0.1-0.35 Hz \cite{schmidt2018introducing}. ST is filtered using a Butterworth bandpass filter with a passband frequency of 0.0001-10 Hz. Like ECG, filtered EDA, RESP, and ST signals are normalized using user-specific z-score normalization. Normalized EDA signals are decomposed into skin conductance level, also known as tonic level, and skin conductance response, also known as the phasic response, using a high pass filter with a cut-off frequency of 0.05 Hz.

For \textit{WESAD}, the filtered signals (ECG, EDA, RESP, and ST) are re-sampled at 256 Hz. For binary stress vs. non-stress classification, neutral and amusement classes are combined to create the non-stress class as done in prior literature \cite{schmidt2018introducing, samyoun2020stress}, while for 3-class classification, stress vs. amusement vs. neutral is used for evaluation purposes. For \textit{SWELL-KW}, the initial relaxation period (8 minutes) is removed from each experiment session. Also, the noisy 1-minute segment from the end of the signals is dropped. The filtered signals (ECG and EDA) are re-sampled from 2048 Hz to 256 Hz. For binary stress classification, time-pressure and interruption sessions are combined to create the stress class and neutral as non-stress class \cite{koldijk2016detecting}. For \textit{CASE}, filtered signals (ECG, EDA, and BVP) are re-sampled from 1000 Hz to 256 Hz. The CASE dataset recorded arousal ratings on a scale of 1-9. For binary arousal classification, low class is considered if the reported arousal value by the participant is less than 5, and high class is considered for values equal to or greater than 5. 

The pre-processed physiological signals are then segmented using 10-second window with 60 percent overlap and stacked into an array to form individual data samples. It is to be noted that the window size and the overlap are selected empirically.

\begin{figure*}[!t]
\begin{center}
    \centering
        \begin{tabular}{cccccc}
               \parbox[c]{.005\textwidth}{\multirow{2}{*}{(A)}} & \parbox[c]{.17\textwidth}{\includegraphics[width=.18\textwidth]{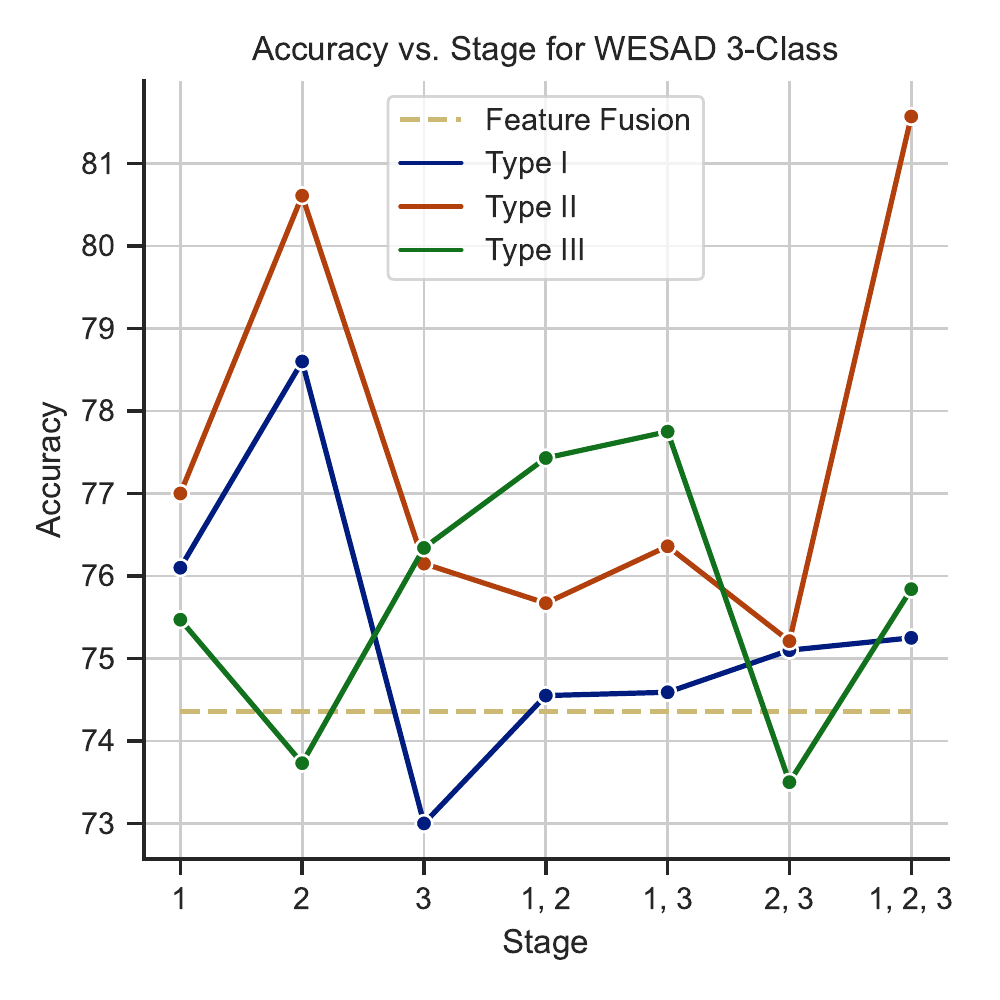}}
               & \parbox[c]{.17\textwidth}{\includegraphics[width=.18\textwidth]{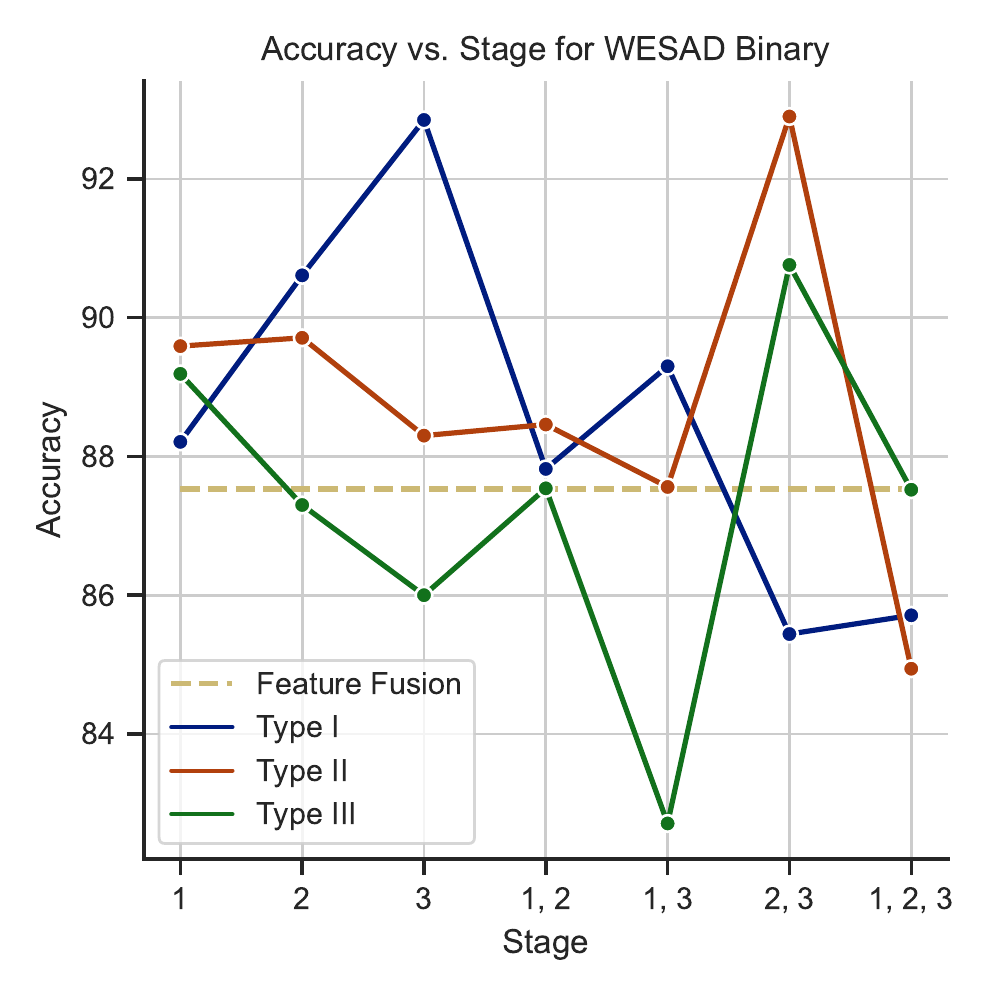}}
               & \parbox[c]{.17\textwidth}{\includegraphics[width=.18\textwidth]{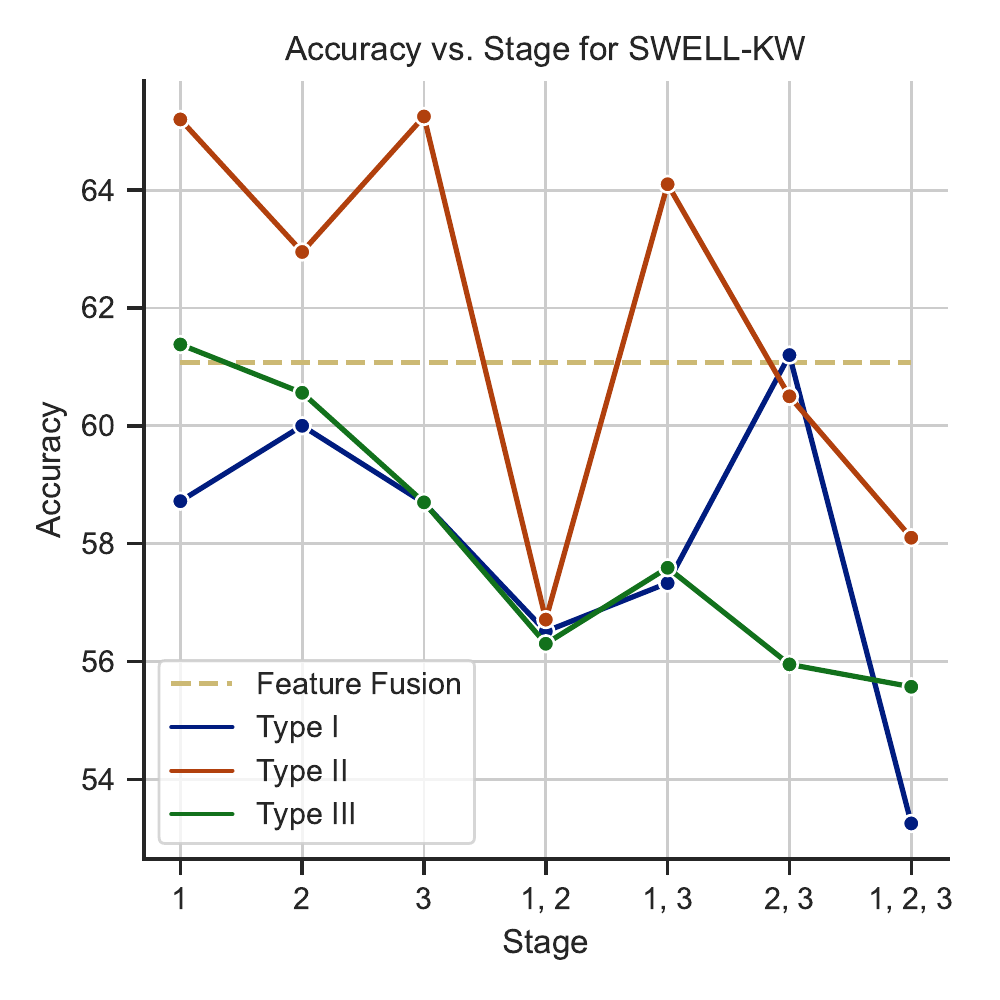}}
               & \parbox[c]{.17\textwidth}{\includegraphics[width=.18\textwidth]{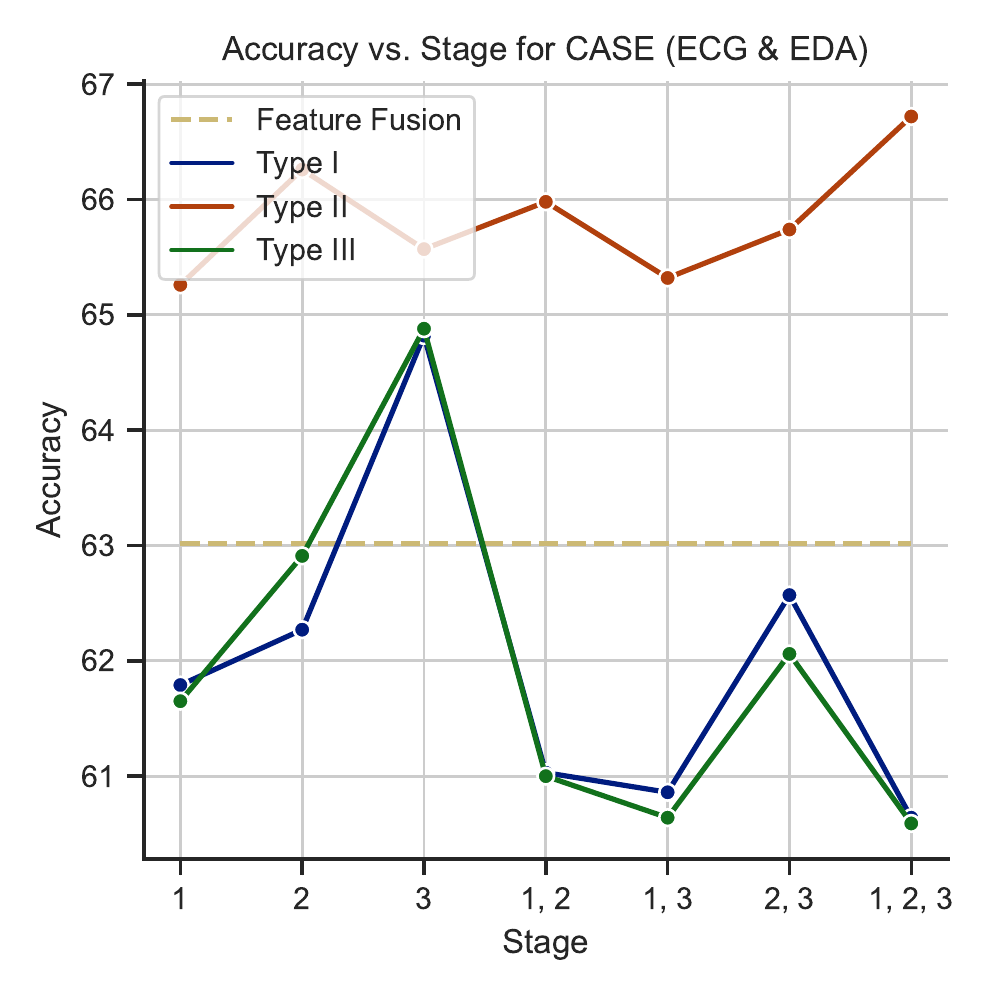}}
               & \parbox[c]{.17\textwidth}{\includegraphics[width=.18\textwidth]{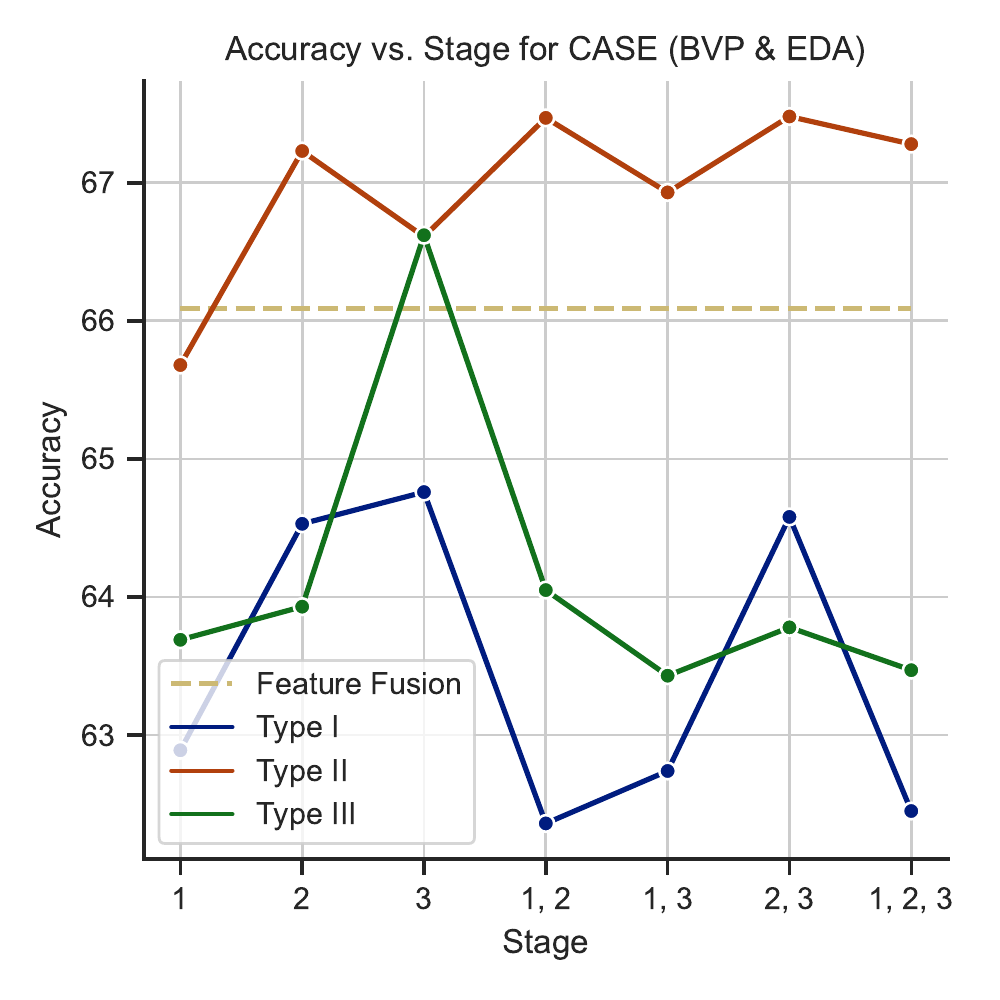}} \\
                & \parbox[c]{.17\textwidth}{\includegraphics[width=.18\textwidth]{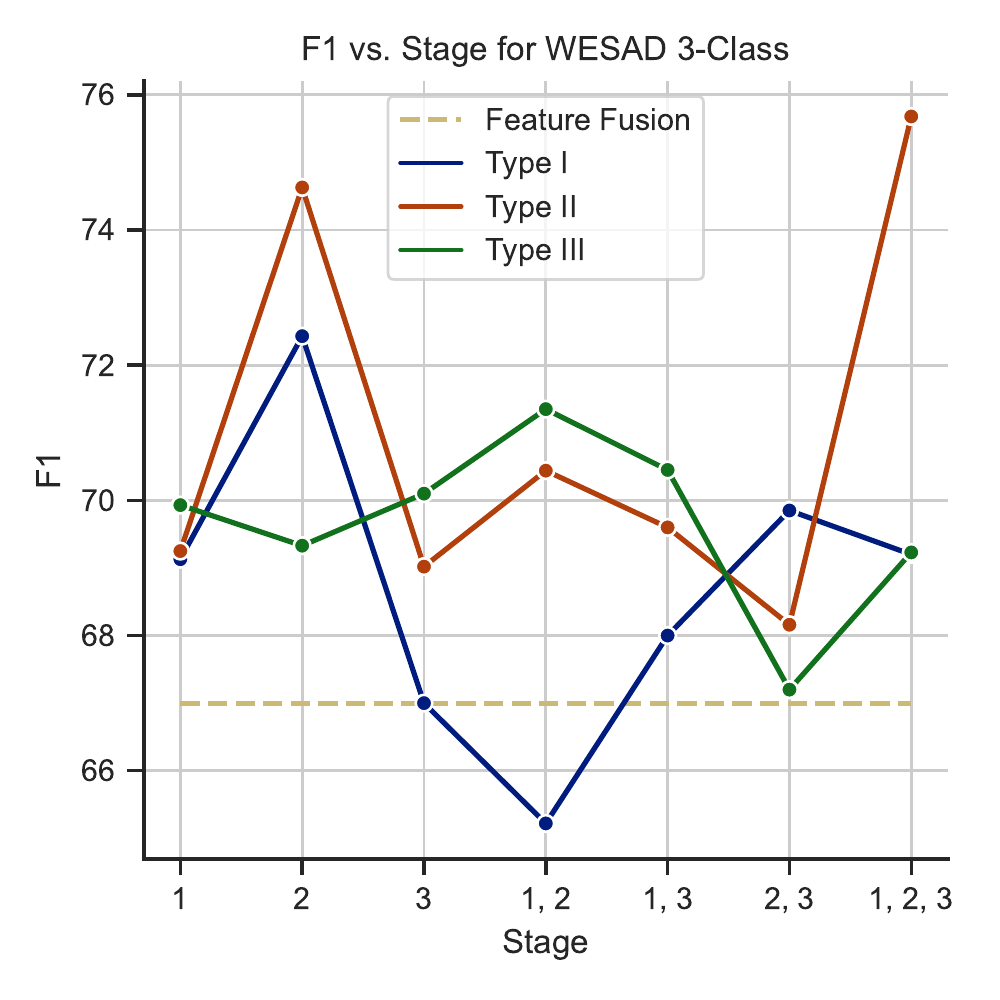}}
               & \parbox[c]{.17\textwidth}{\includegraphics[width=.18\textwidth]{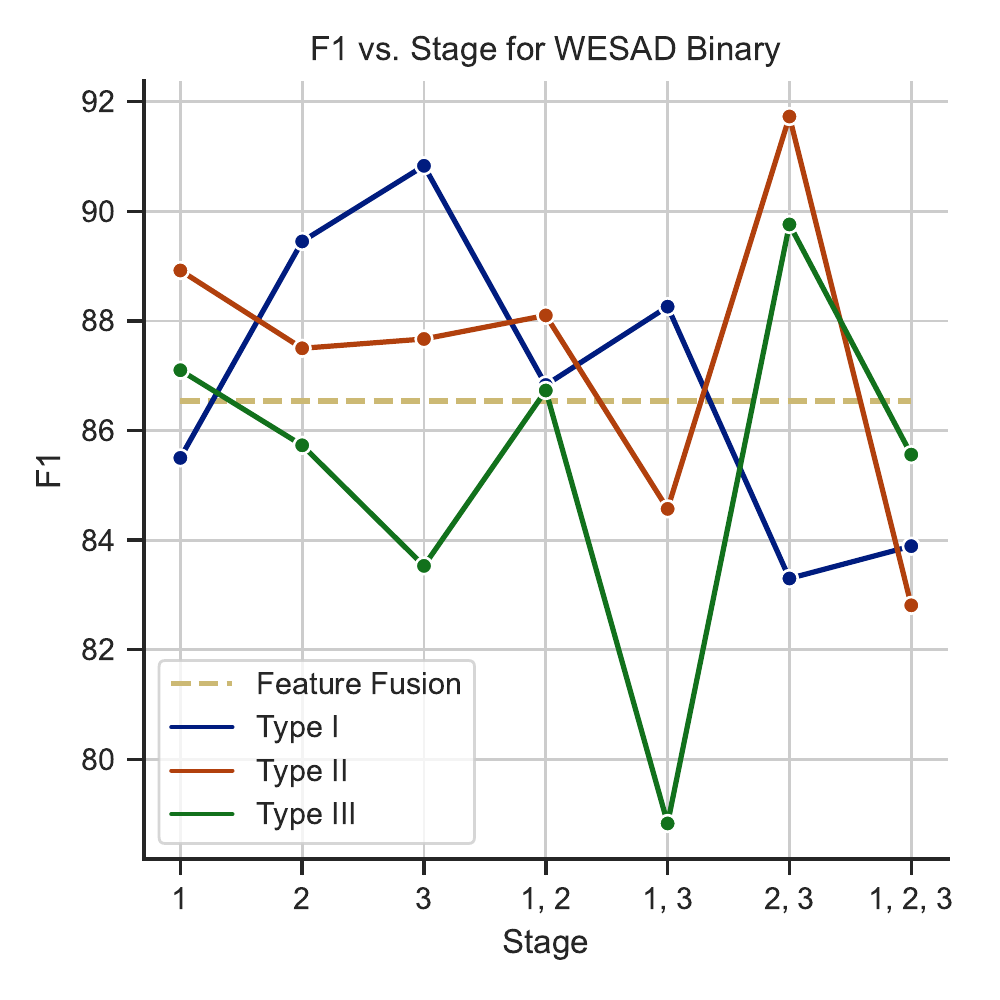}}
               & \parbox[c]{.17\textwidth}{\includegraphics[width=.18\textwidth]{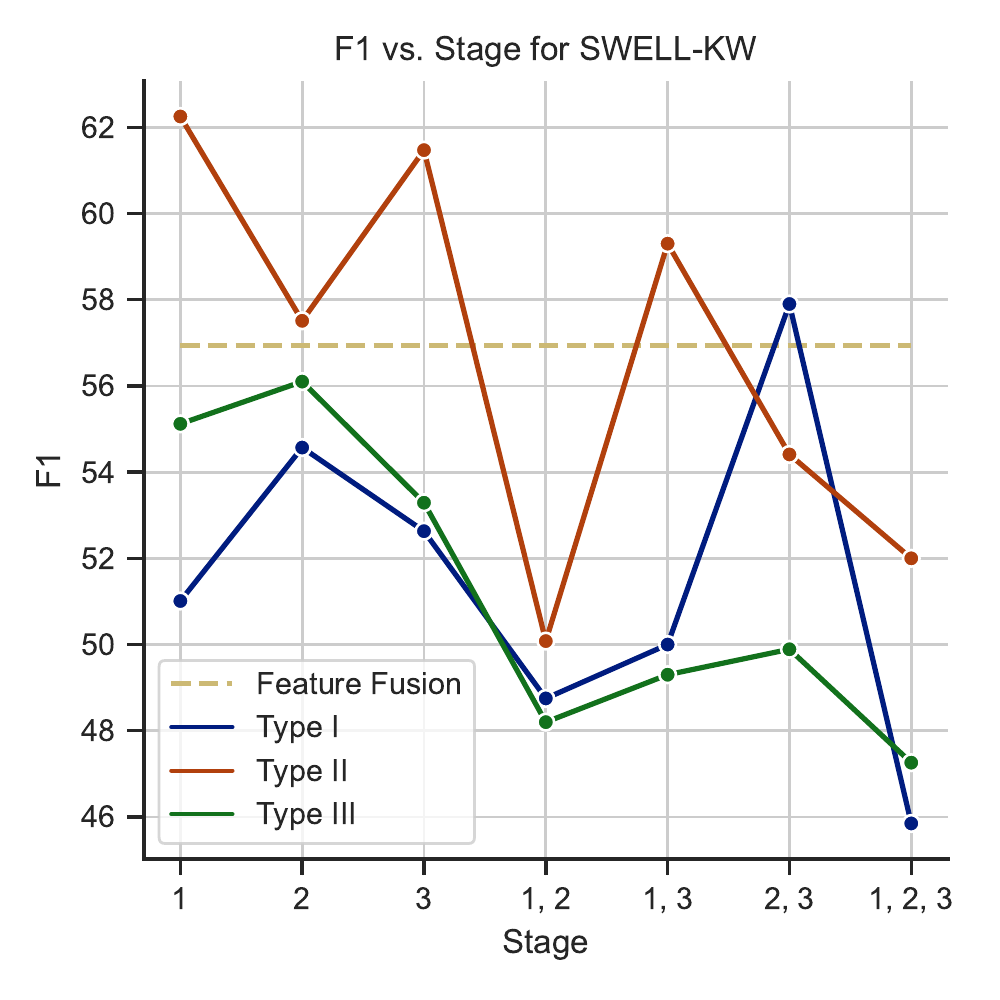}}
               & \parbox[c]{.17\textwidth}{\includegraphics[width=.18\textwidth]{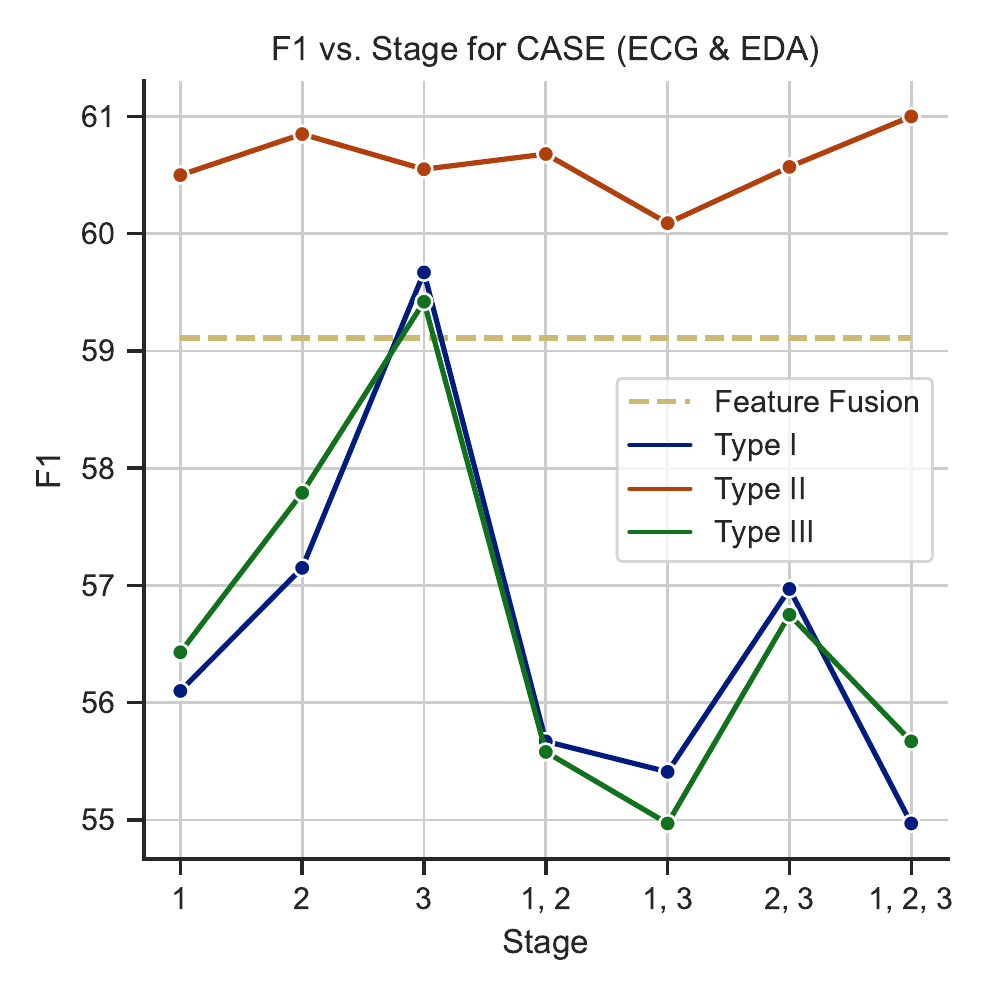}}
               & \parbox[c]{.17\textwidth}{\includegraphics[width=.18\textwidth]{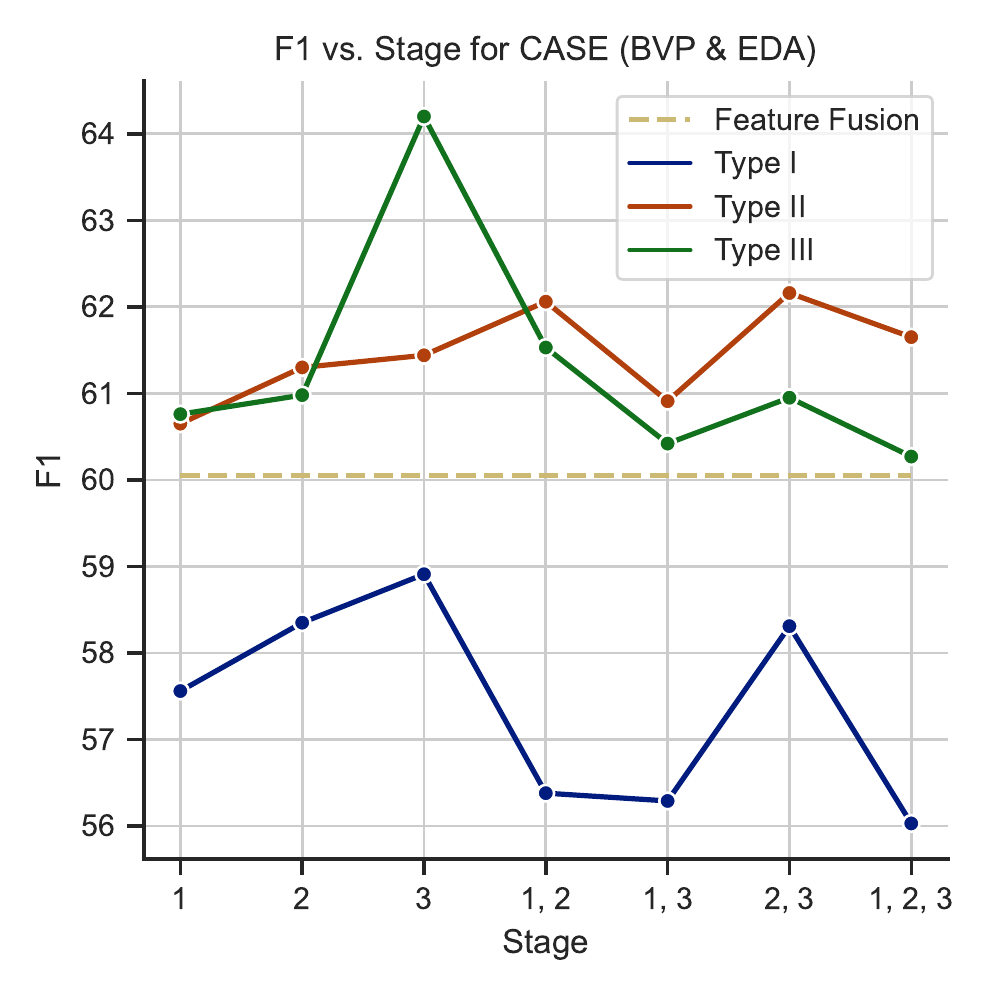}} \\ \hline
               \multirow{2}{*}{(B)} &\parbox[c]{.17\textwidth}{\includegraphics[width=.18\textwidth]{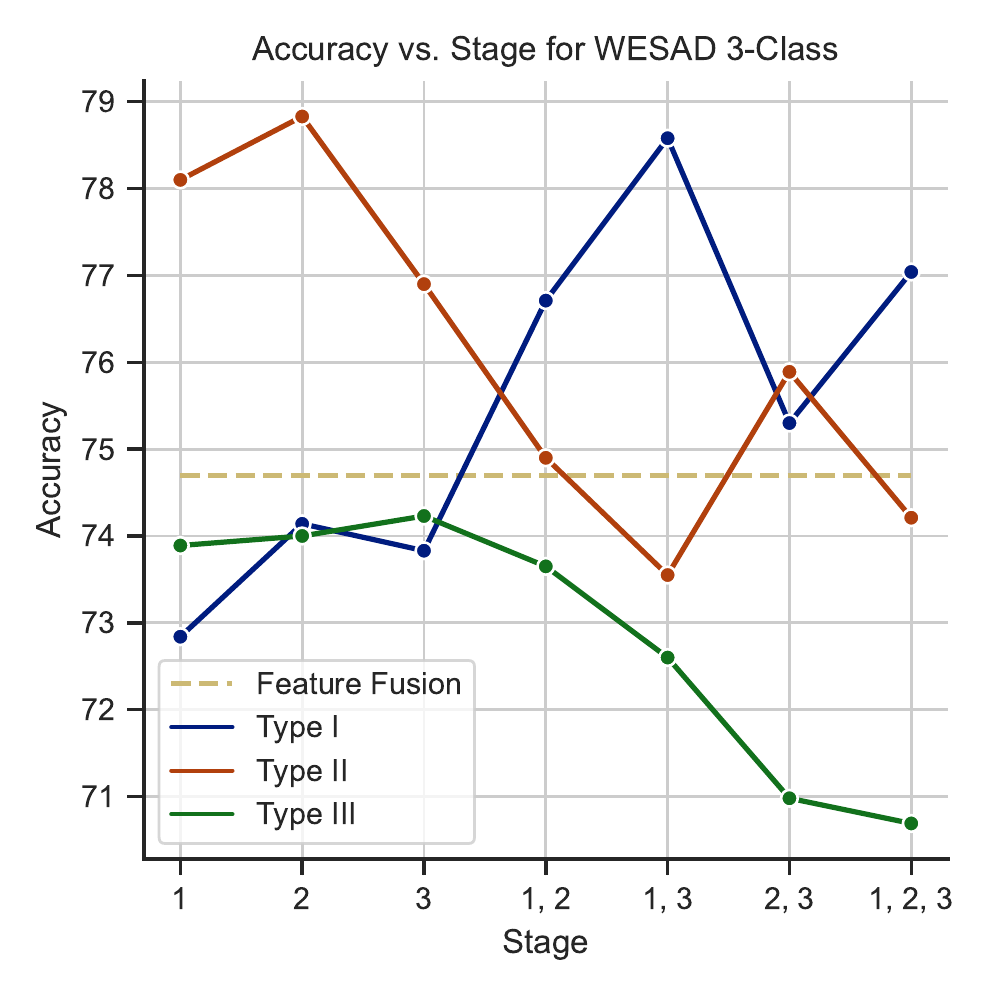}}
               & \parbox[c]{.17\textwidth}{\includegraphics[width=.18\textwidth]{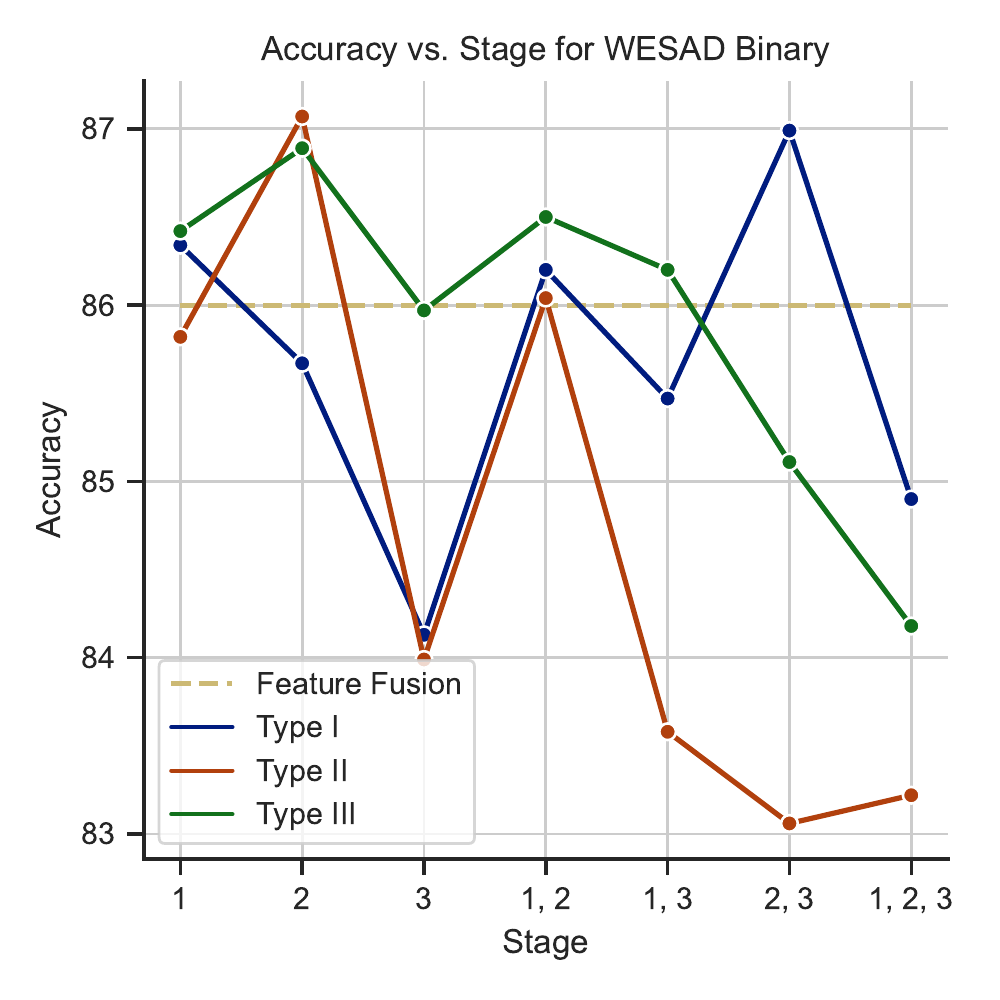}}
               & \parbox[c]{.17\textwidth}{\includegraphics[width=.18\textwidth]{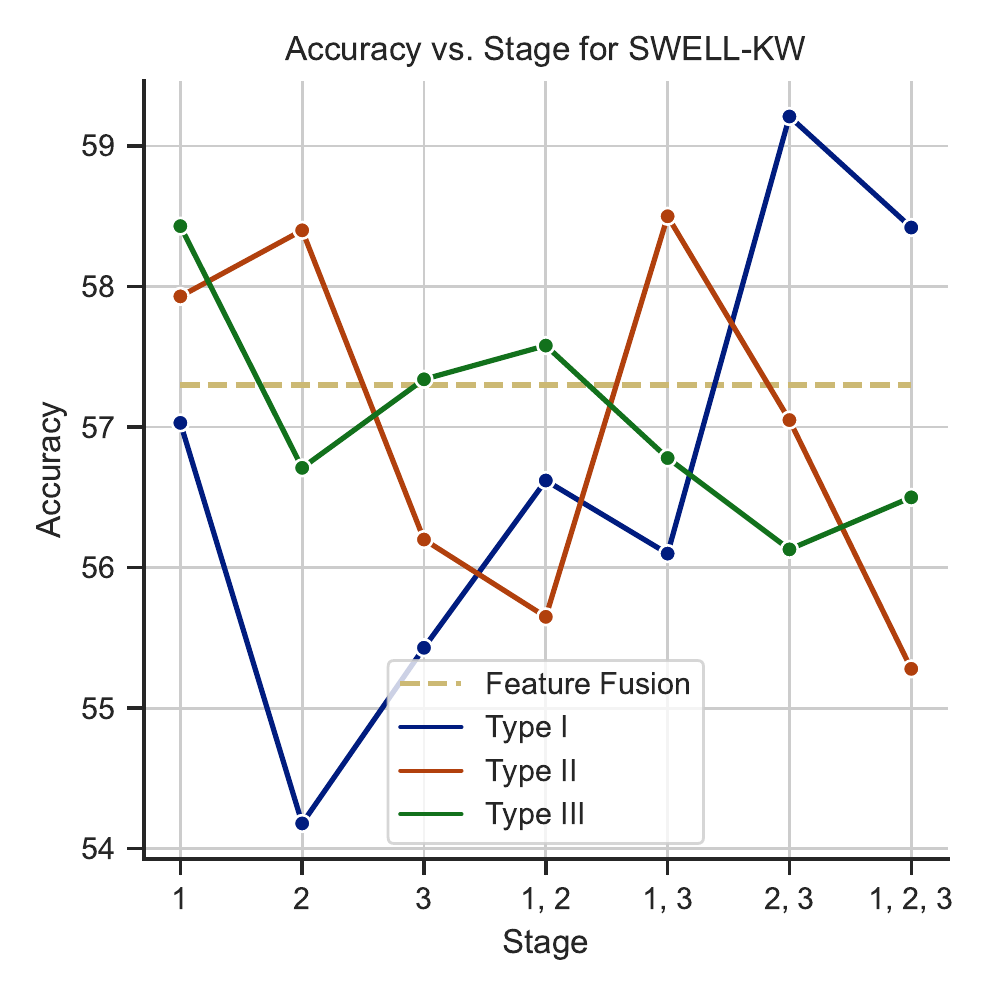}}
               & \parbox[c]{.17\textwidth}{\includegraphics[width=.18\textwidth]{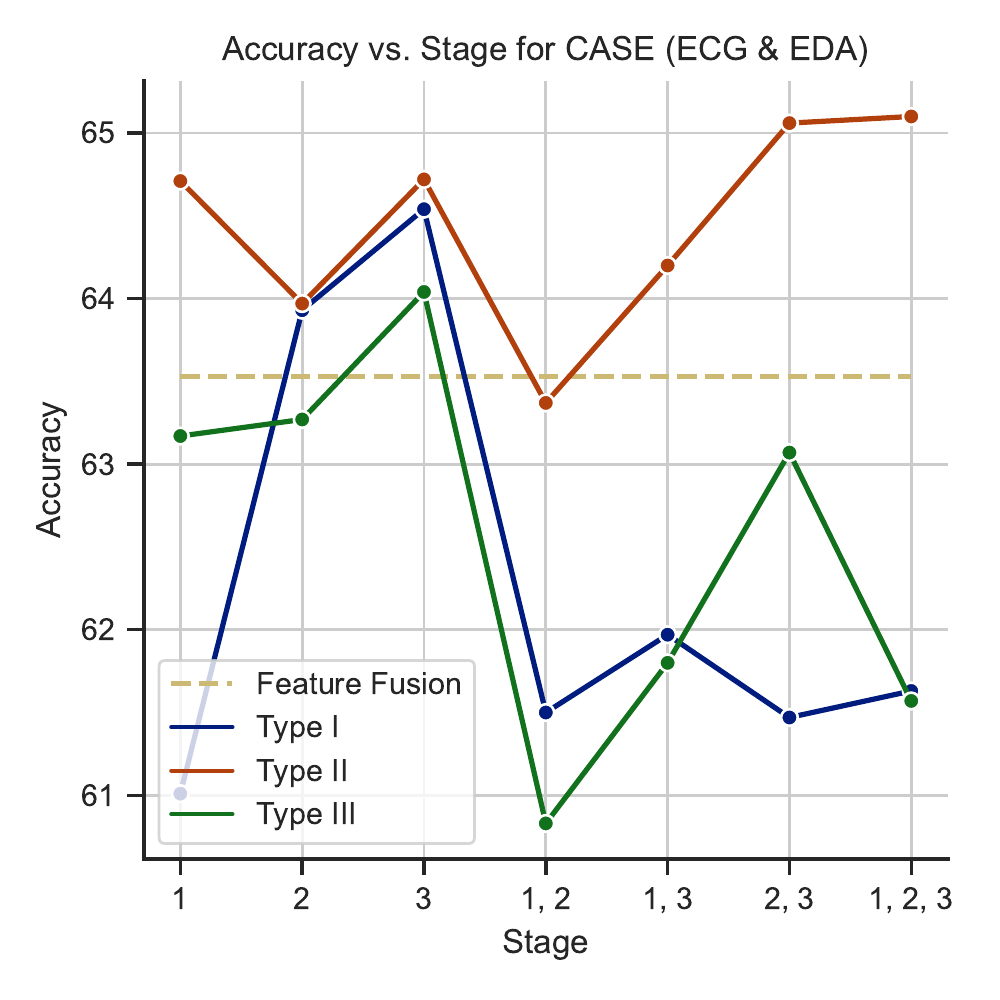}}
               & \parbox[c]{.17\textwidth}{\includegraphics[width=.18\textwidth]{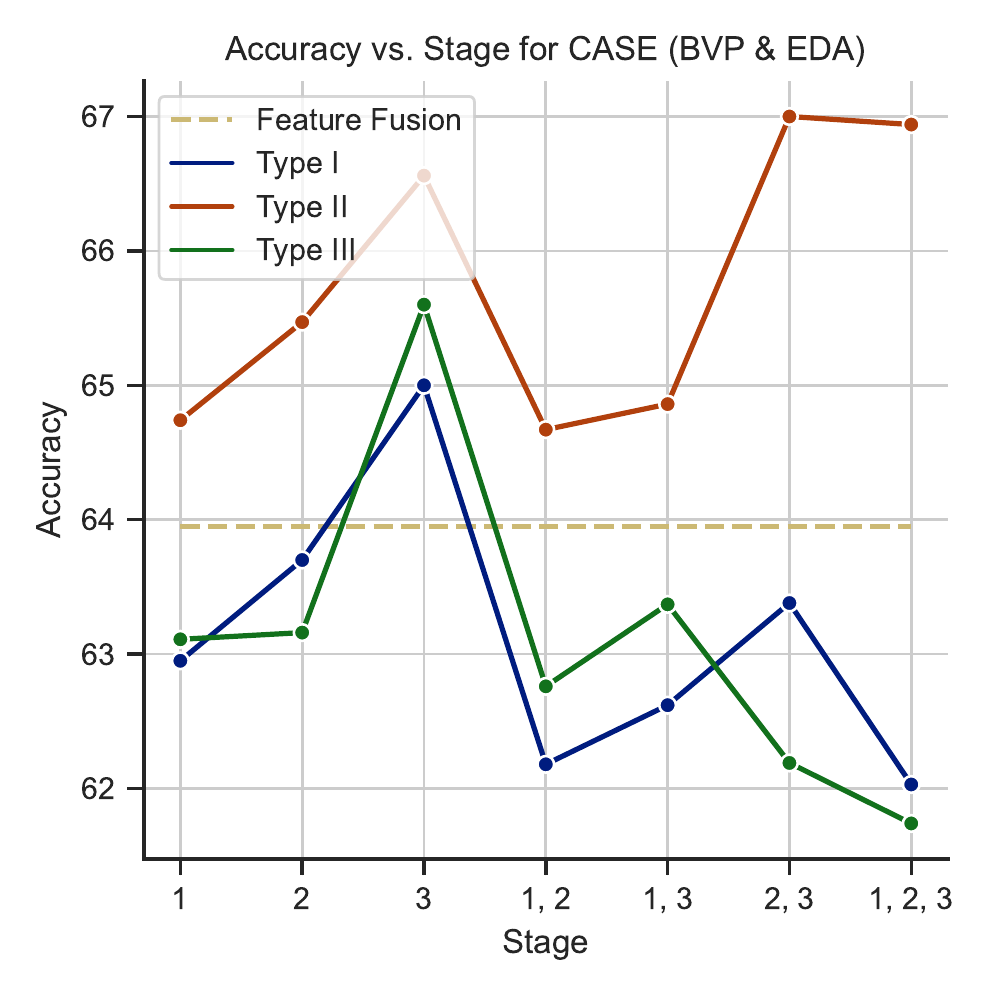}} \\
               &\parbox[c]{.17\textwidth}{\includegraphics[width=.18\textwidth]{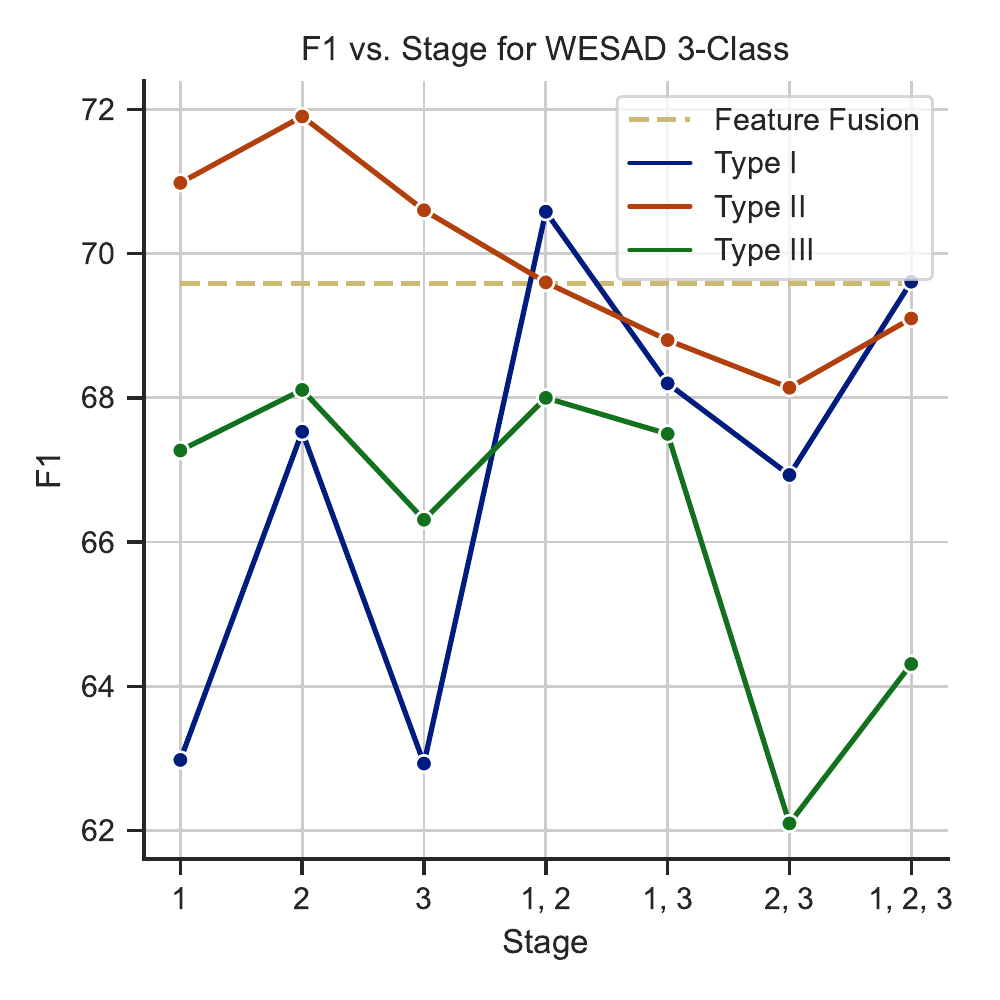}}
               & \parbox[c]{.17\textwidth}{\includegraphics[width=.18\textwidth]{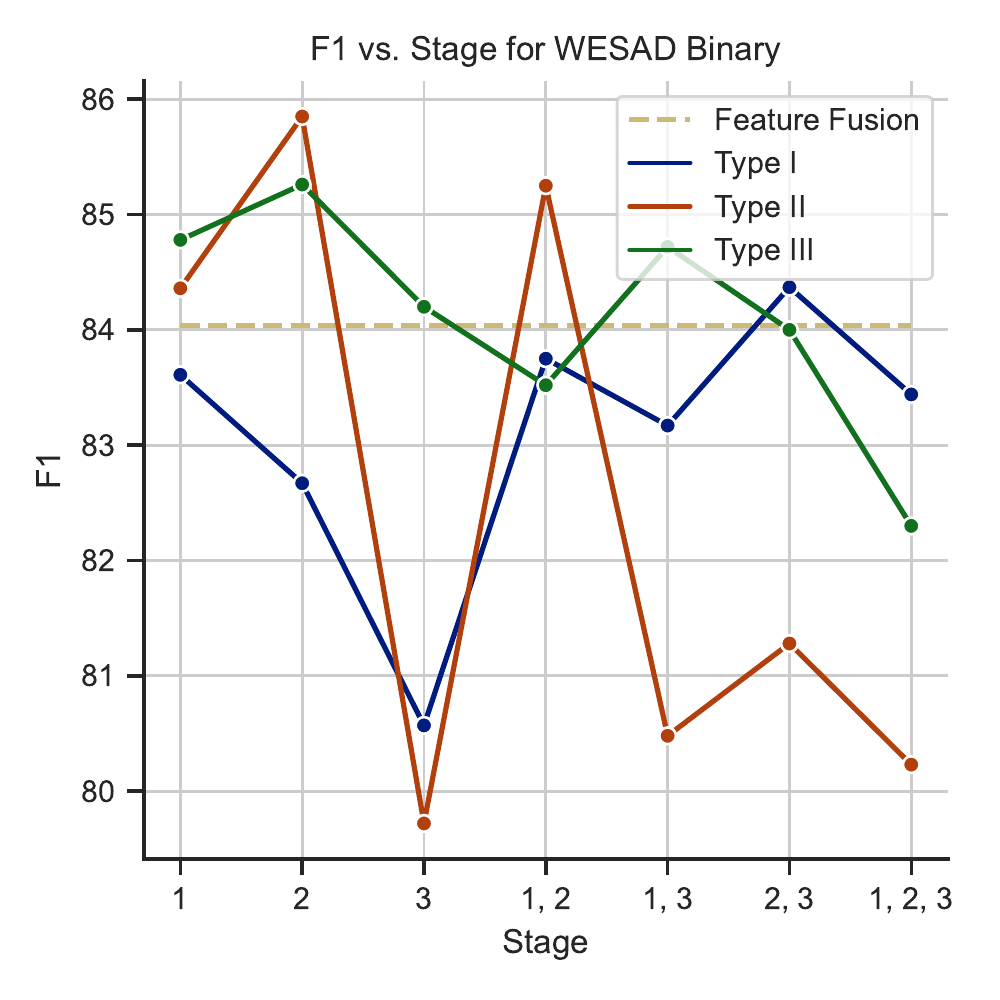}}
               & \parbox[c]{.17\textwidth}{\includegraphics[width=.18\textwidth]{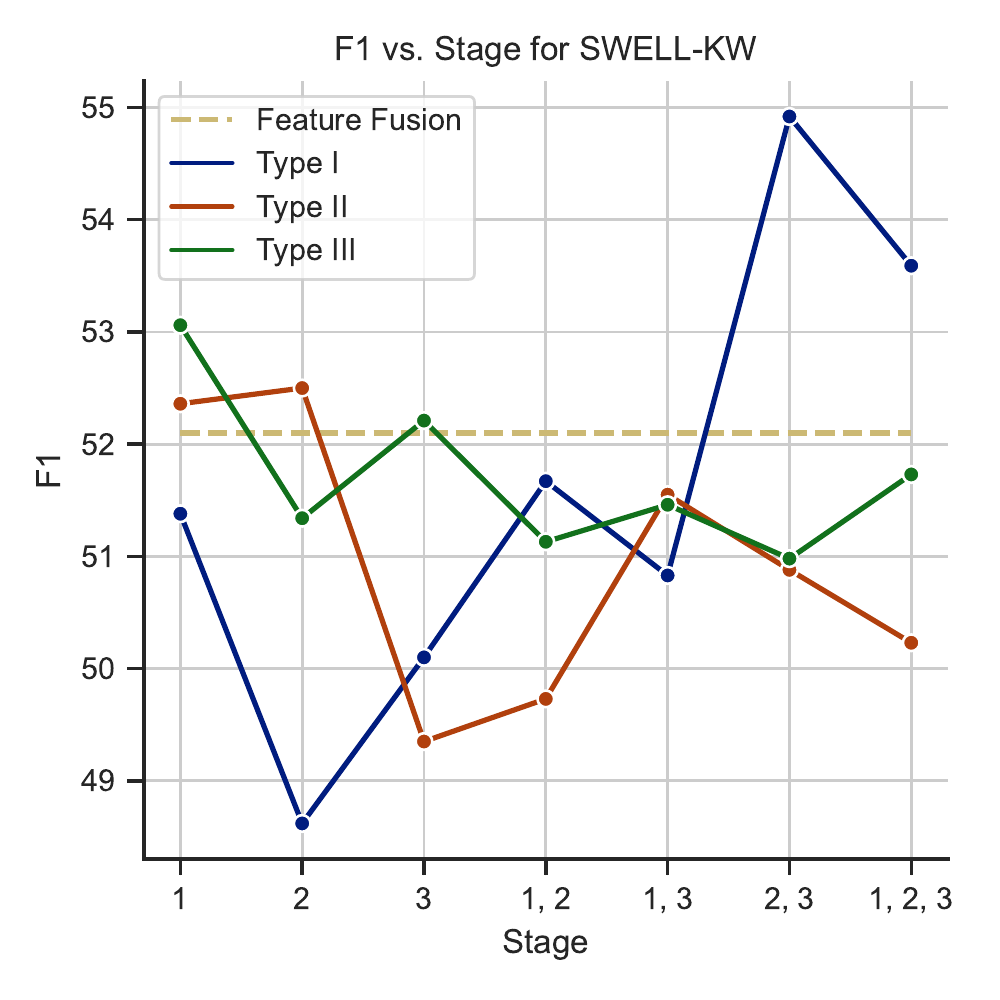}}
               & \parbox[c]{.17\textwidth}{\includegraphics[width=.18\textwidth]{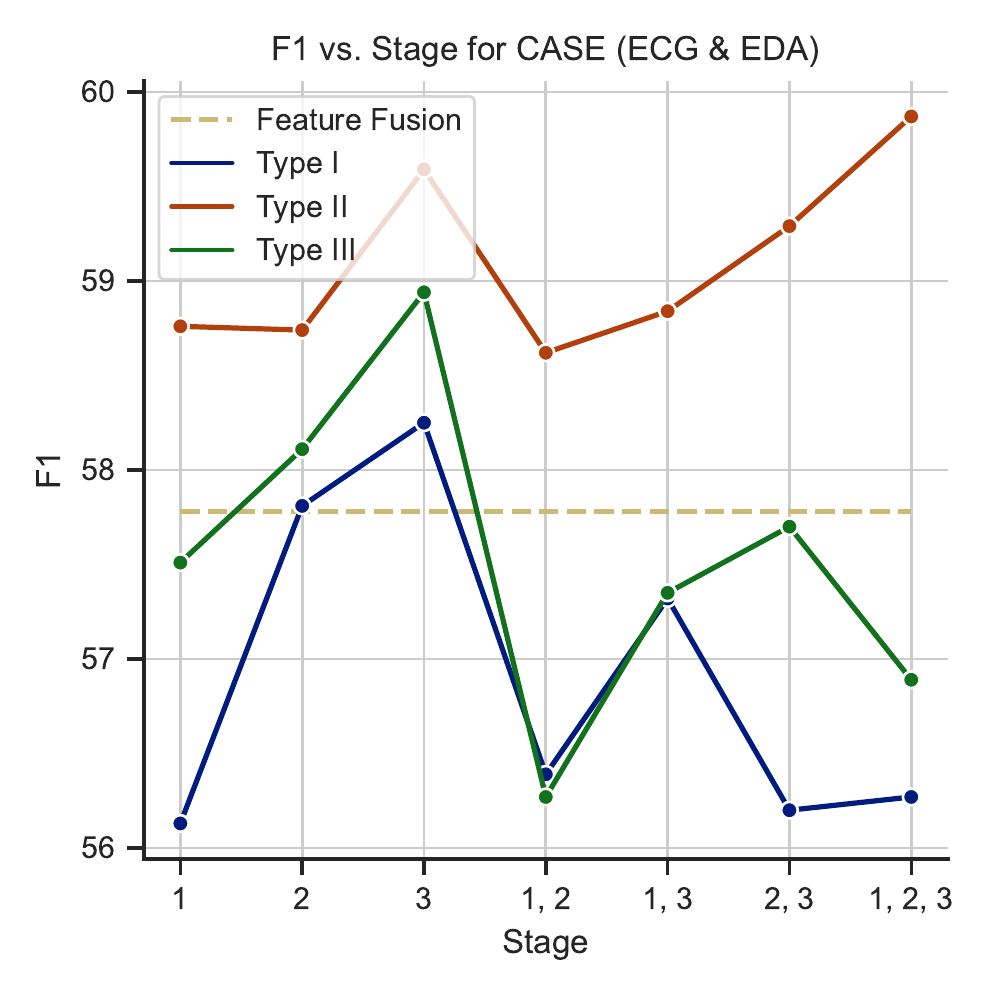}}
               & \parbox[c]{.17\textwidth}{\includegraphics[width=.18\textwidth]{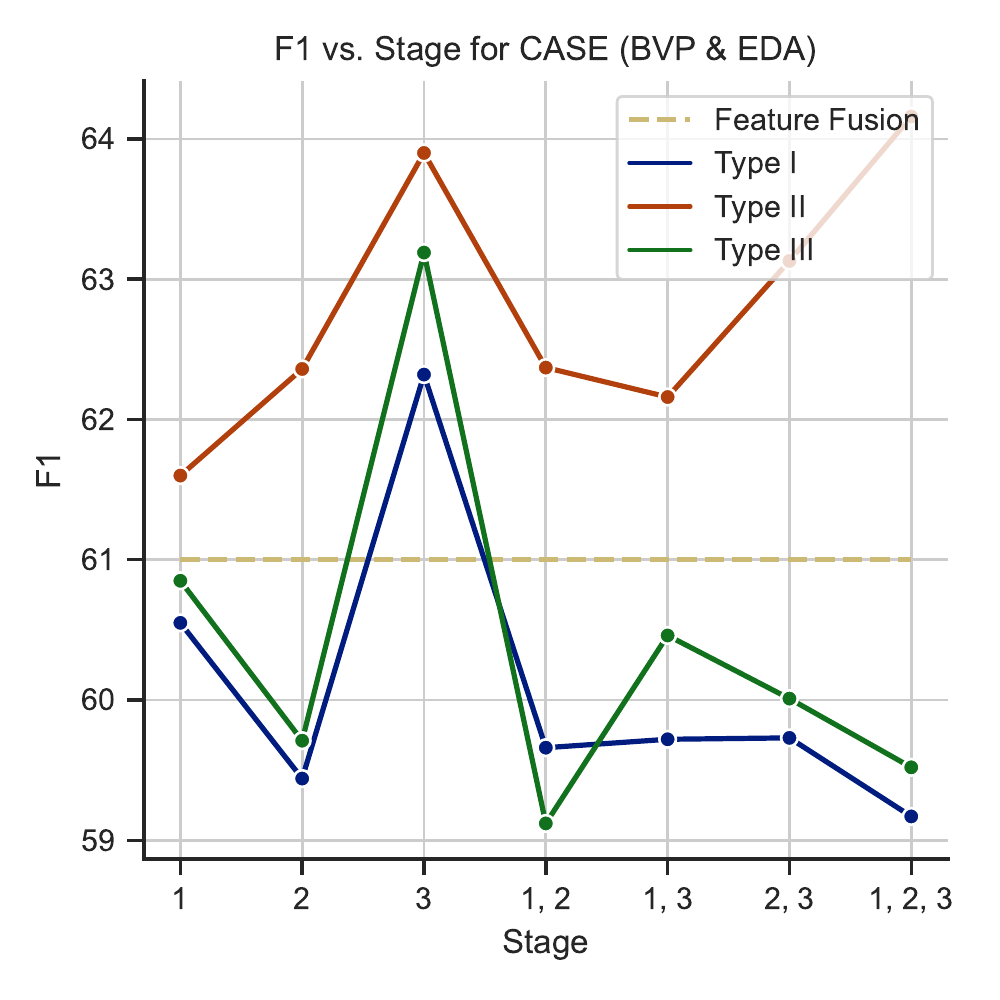}} \\
        \end{tabular}
\end{center}
\caption{Each plot presents the Accuracy or F1 values versus different stages where AttX has been integrated, for different AttX types, across different datasets. 
The horizontal dashed line depicts the Accuracy or F1 when only standard feature fusion is used. Here, (A) represents the performance trend for the VGG pipeline, while (B) represents the performance for the ResNet pipeline.}
\label{fig:attx_config}
\end{figure*}

\subsection{Implementation and Training Details}
We implement our pipeline using Tensorflow on an NVIDIA GeForce RTX 2080 Ti GPU. We use a standard Adam algorithm with a learning rate of 1e-3 for optimization for WESAD and SWELL-KW, while for CASE, AdaDelta with a decay rate of 0.95 and learning rate of 5e-3 was used as it resulted in better and more stable training. For WESAD, we use focal loss with values of alpha and gamma as $4.0$ and $2.0$ \cite{lin2017focal}, respectively, while for SWELL-KW and CASE, we use standard cross-entropy loss. The networks are trained with a batch size of 256 for 100 epochs for all the experiments.

For evaluating our method on different architectures (both VGG and ResNet encoder pipelines), we use accuracy and F1-score with macro-averaging. For testing our model, we use Leave-One-Subject-Out (\textbf{LOSO}) evaluation scheme. For tuning the hyperparameters, we use twenty percent of the training set as a validation set.

\section{Results and Analysis}
\label{sec:RAW}
In this section, we explore the performance of our solution across various configurations. We then evaluate the optimum type and location for our proposed cross-modal connections across different datasets and modality combinations. We then further analyze our method by visualizing the learned representations. We wrap up our results section by comparing our method with state-of-the-art solutions in the field.

\subsection{Performance}
We present the performance of AttX for all the configurations (different types and stages) on the three datasets. Figure \ref{fig:attx_config} presents the results where WESAD 3-class is depicted in column 1, WESAD binary class in column 2, SWELL-KW in column 3, CASE with ECG and EDA in column 4, and CASE with BVP and EDA in column 5. The figure shows the accuracy and the f1-score plots for Type I, II, and III versus different stages for the VGG (first and second rows) and ResNet (third and fourth rows) pipelines. The results are discussed below.

Figure \ref{fig:attx_config} column 1 shows the performance of different AttX configurations for WESAD, 3-class classification, with VGG and ResNet backbones. For Type I connection, the best performance from the VGG pipeline (accuracy of 78.60 and f1 of 72.43, hereafter mentioned in the same order) is obtained at stage 2. In contrast, the ResNet pipeline achieves the best performance at stages 1 \& 3 (78.58, 68.20). For Type II, VGG obtains the best performance at stages 1, 2, \& 3 (81.57, 75.68), and ResNet performs best at stage 2 (78.83, 71.90). Type III connections with VGG and ResNet obtain the best performance at stages 1 \& 2 (77.43, 71.35) and stage 2 (74.00, 68.11), respectively. 
In column 2 of the same figure, we present the performance of AttX configurations for WESAD, binary classification task. Type I connections perform best at stage 3 for VGG (92.85, 90.83) and stages 2 \& 3 for ResNet (86.99, 84.37). The best performance with Type II connections is achieved at stages 2 \& 3 for VGG (92.90, 91.73) and stage 2 for ResNet (87.07, 85.85). For Type III, VGG performs best at stages 2 \& 3 (90.76, 89.76), while ResNet performs best at stage 2 (86.89, 85.26). 

Column 3 of Figure \ref{fig:attx_config} presents the performance of AttX for SWELL-KW dataset. For Type I, Attx at stages 2 \& 3 performs best for VGG (61.20, 57.90) and ResNet (59.21, 54.92). For Type II, the best performance for VGG is achieved at stage 1 (65.20, 62.25) and at stage 2 for ResNet (58.40, 52.50). Type III connection perform best for VGG (61.38, 55.12) and ResNet (58.43, 53.06) at stage 1. 

Performance of AttX configurations for CASE (ECG and EDA) is shown in Figure \ref{fig:attx_config} column 4. The Type I connection performs best at stage 3 for VGG (64.82, 59.67) and ResNet (64.54, 58.25). Type II connection performs best at stages 1, 2, \& 3 for VGG (66.72, 61.00) and ResNet (65.10, 59.87). For Type III, for both VGG (64.88, 59.42) and ResNet (64.04, 58.94), the best score is achieved at stage 3.
Column 5 of the figure presents the performance of AttX configurations for CASE (BVP and EDA). For Type I connection, we observe that the best stage to add AttX is at stage 3 for VGG (64.76, 58.91) and ResNet (65.00, 62.32). In contrast, the Type II connection at stage 2 \& 3 and 1, 2, \& 3 performs the best for VGG (67.48, 62.16) and ResNet (66.94, 64.16), respectively. For Type III, stage 3 performs the best for VGG (66.62, 64.20) and ResNet (65.60, 63.19). 

\begin{figure}
    \centering
    \begin{subfigure}[t]{\linewidth}
    \centering
    \includegraphics[width=0.49\linewidth]{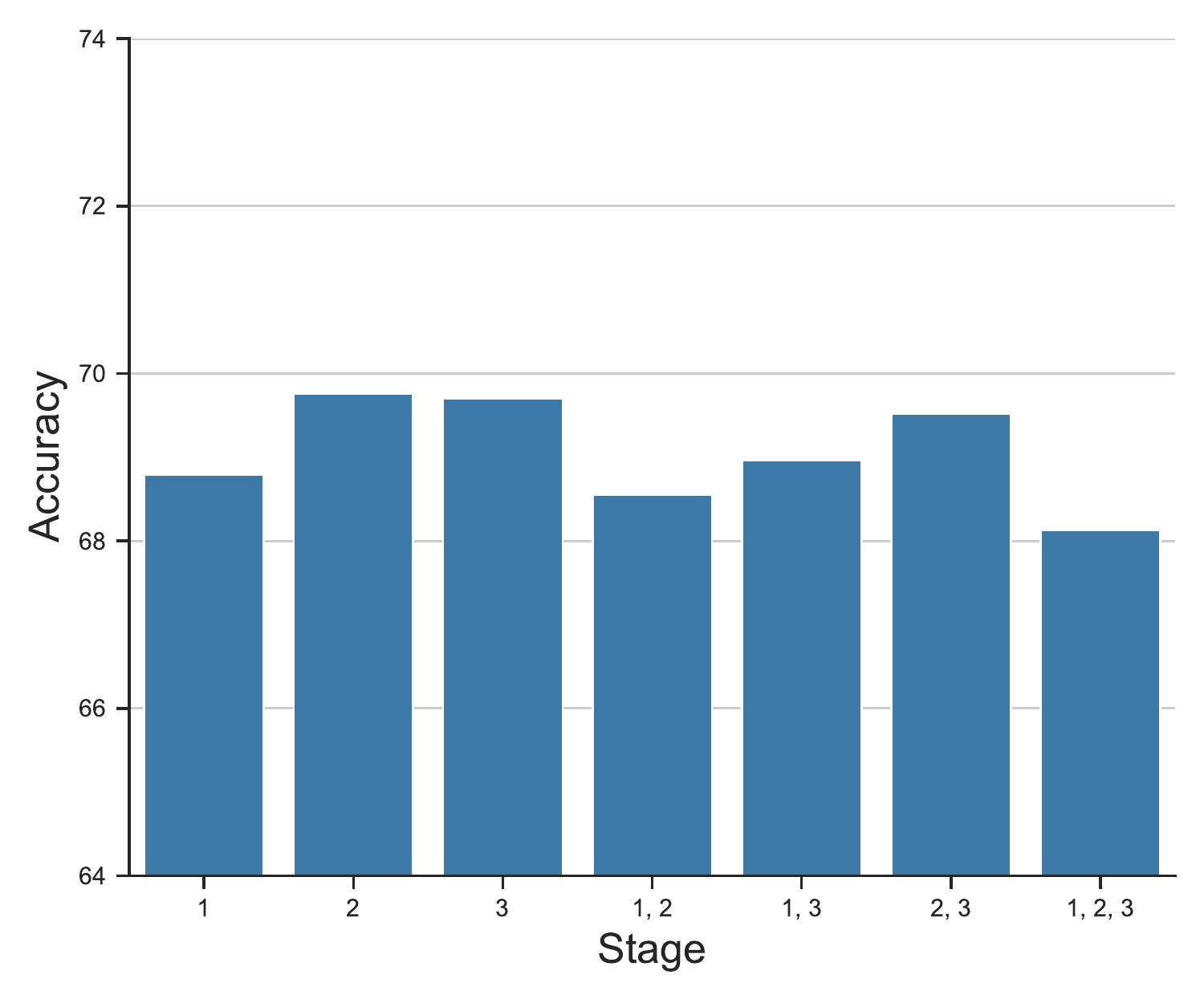}
    \includegraphics[width=0.49\linewidth]{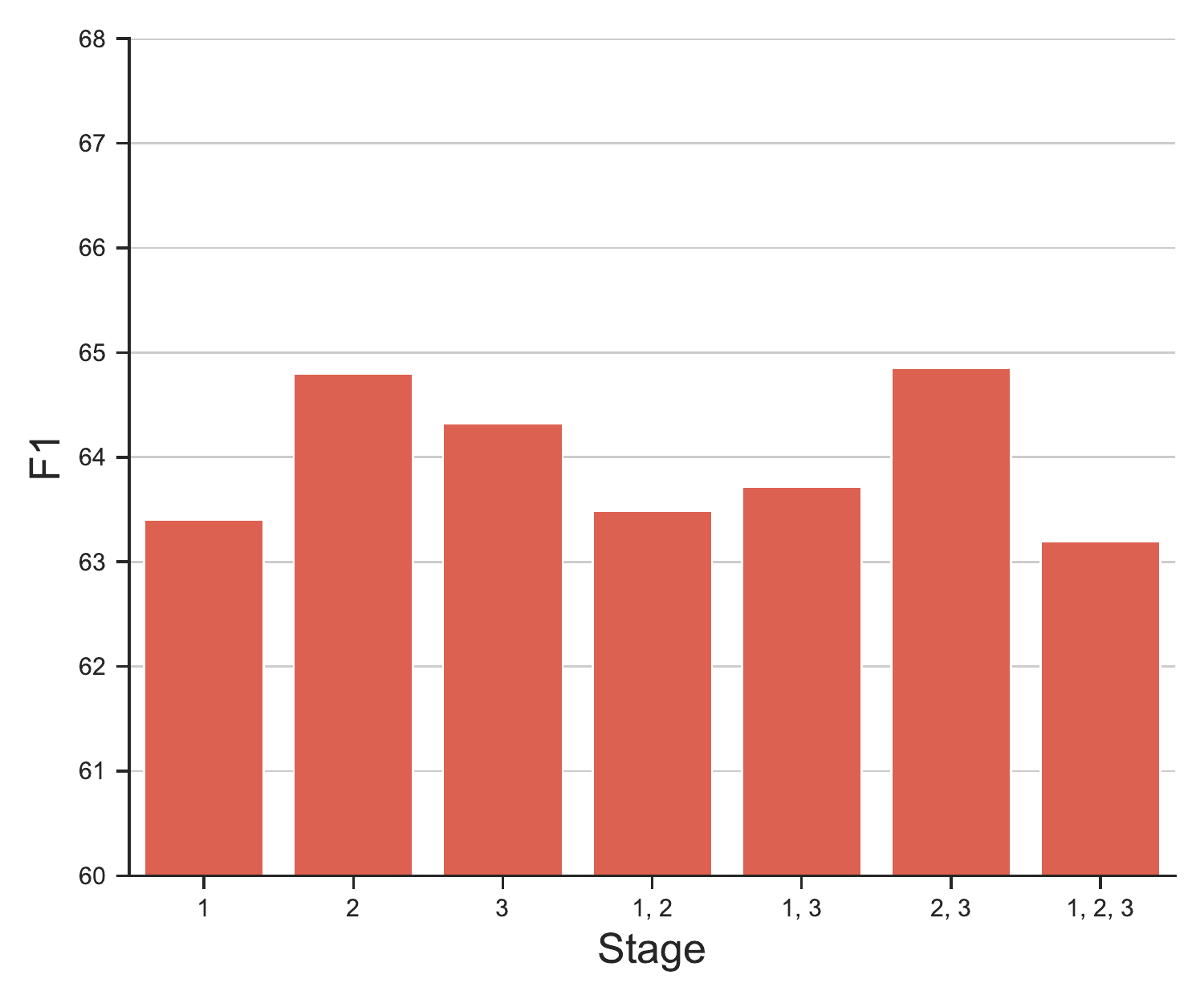}
    \caption{Type I}\label{fig:type_1}
    \end{subfigure}
    \begin{subfigure}[t]{\linewidth}
    \centering
    \includegraphics[width=0.49\linewidth]{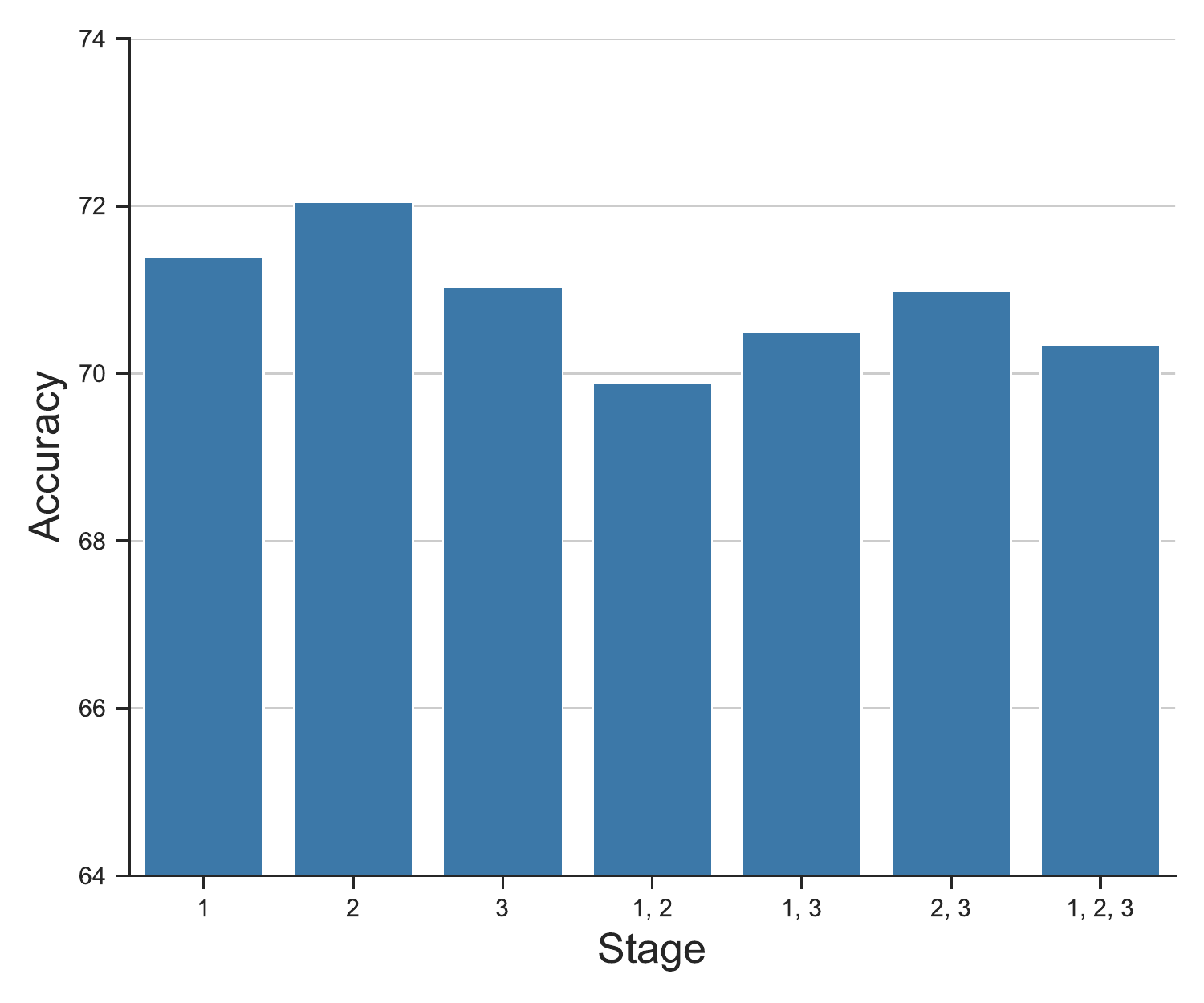}
    \includegraphics[width=0.49\linewidth]{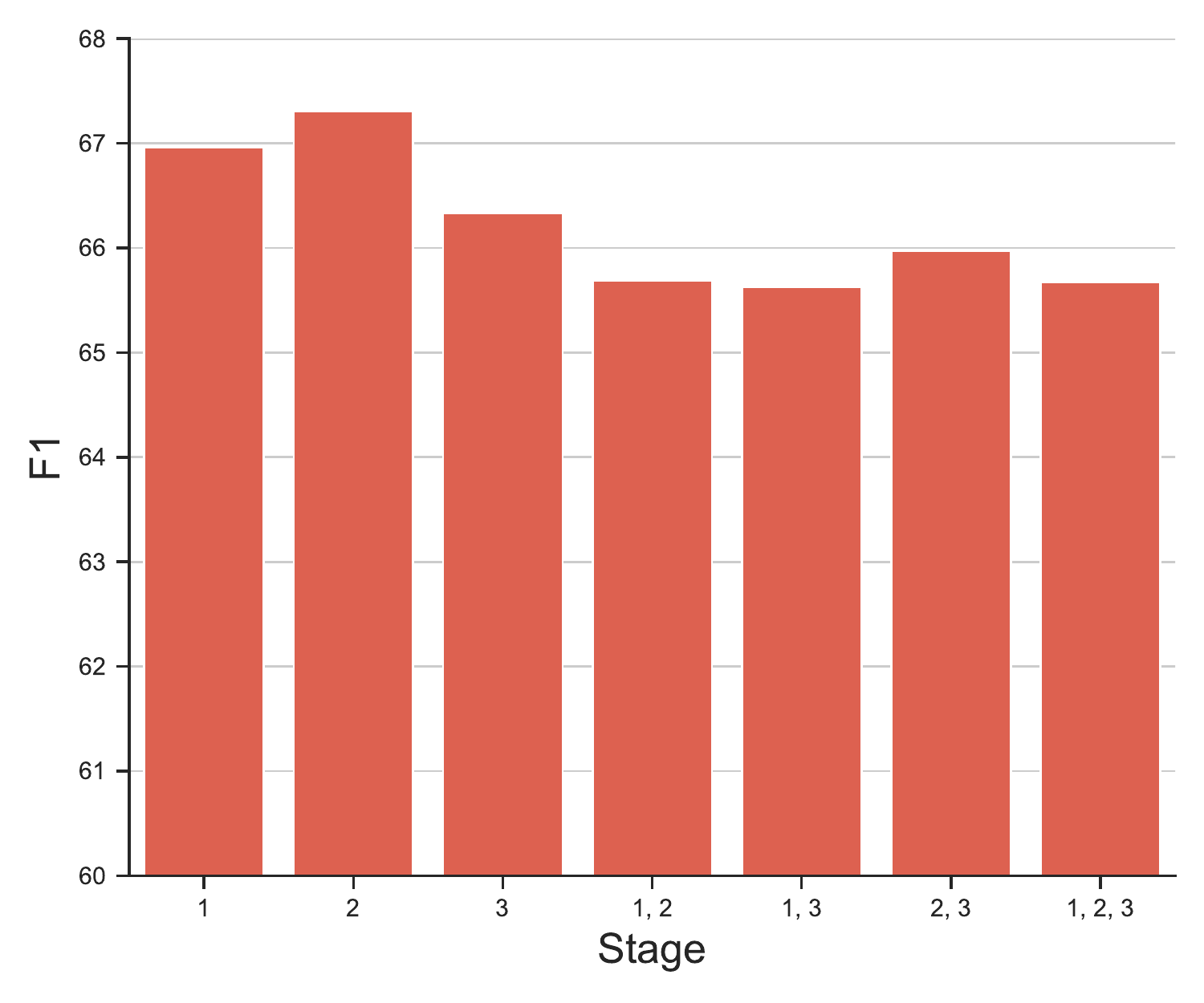}
    \caption{Type II}\label{fig:type_2}
    \end{subfigure}
    \begin{subfigure}[t]{\linewidth}
    \centering
    \includegraphics[width=0.49\linewidth]{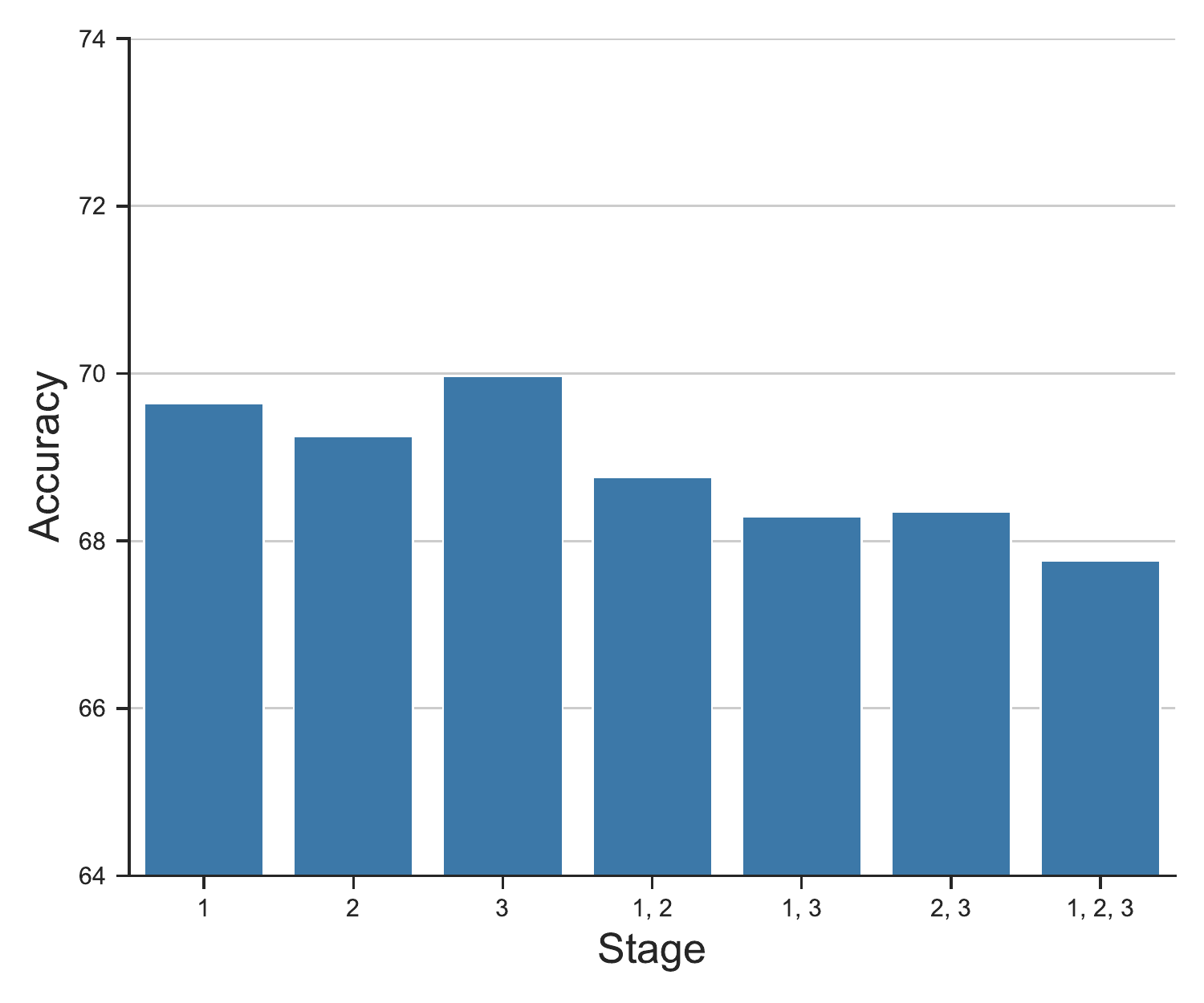}
    \includegraphics[width=0.49\linewidth]{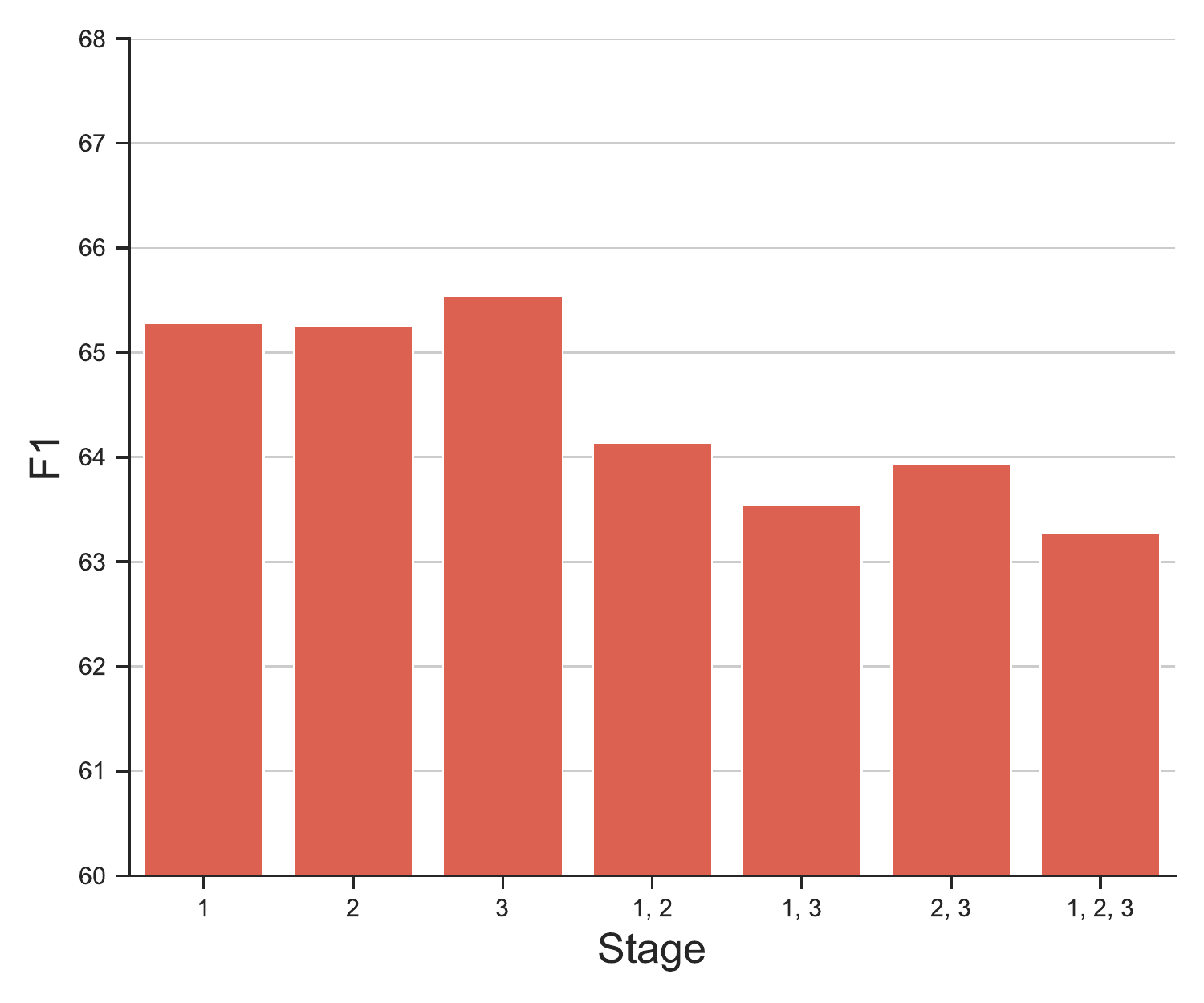}
    \caption{Type III}\label{fig:type_3}
    \end{subfigure}
    \caption{Plots for average Accuracy (left) and F1 (right) for Type I, II, and III versus different stages where AttX has been integrated.}
    \label{fig:Types_bar}
\end{figure}

\begin{figure}
    \centering
    \begin{subfigure}[t]{\linewidth}
    \centering
    \includegraphics[width=0.49\linewidth]{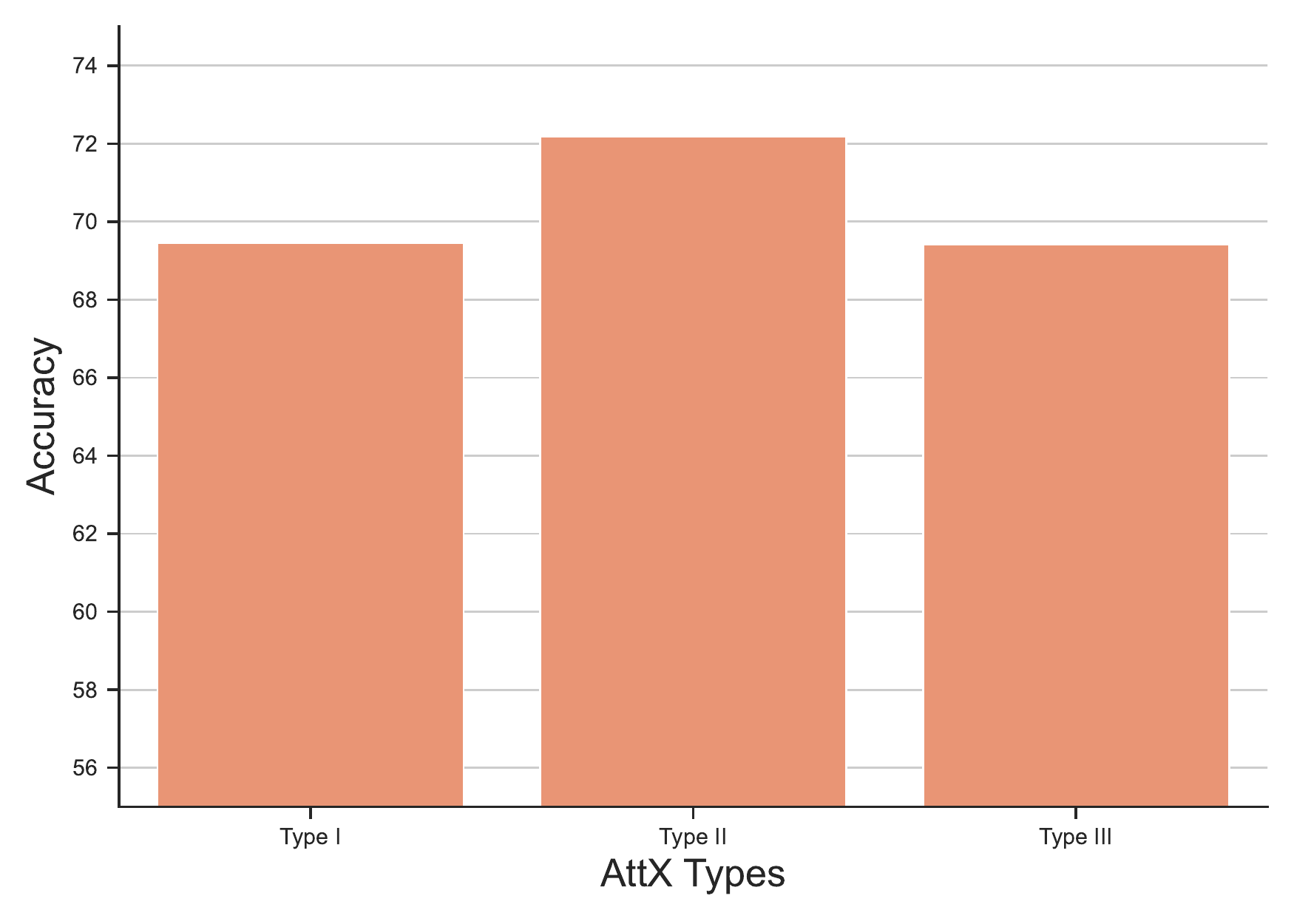}
    \includegraphics[width=0.49\linewidth]{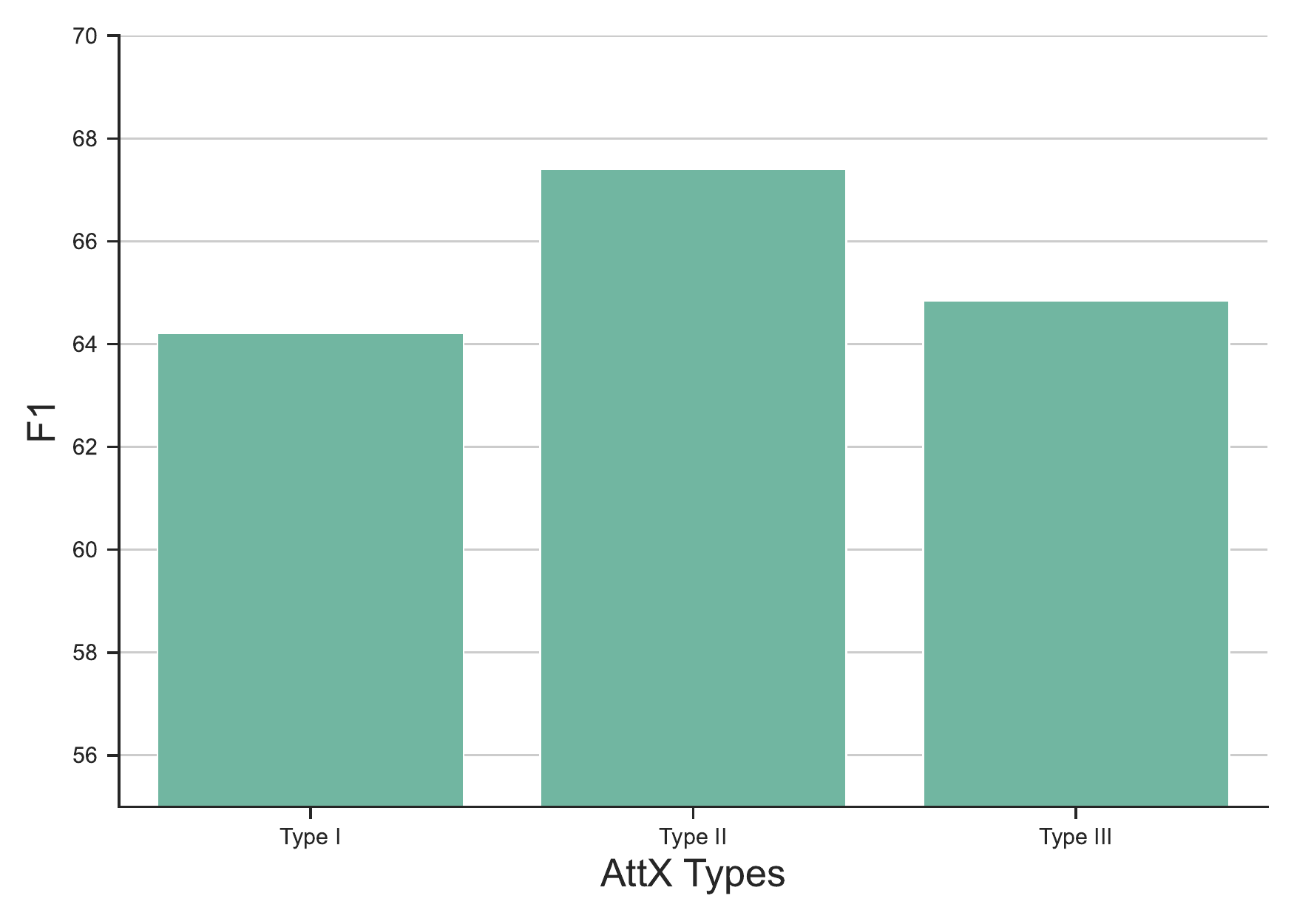}
    \caption{VGG Pipeline}\label{fig:type_vgg1}
    \end{subfigure}
    \begin{subfigure}[t]{\linewidth}
    \centering
    \includegraphics[width=0.49\linewidth]{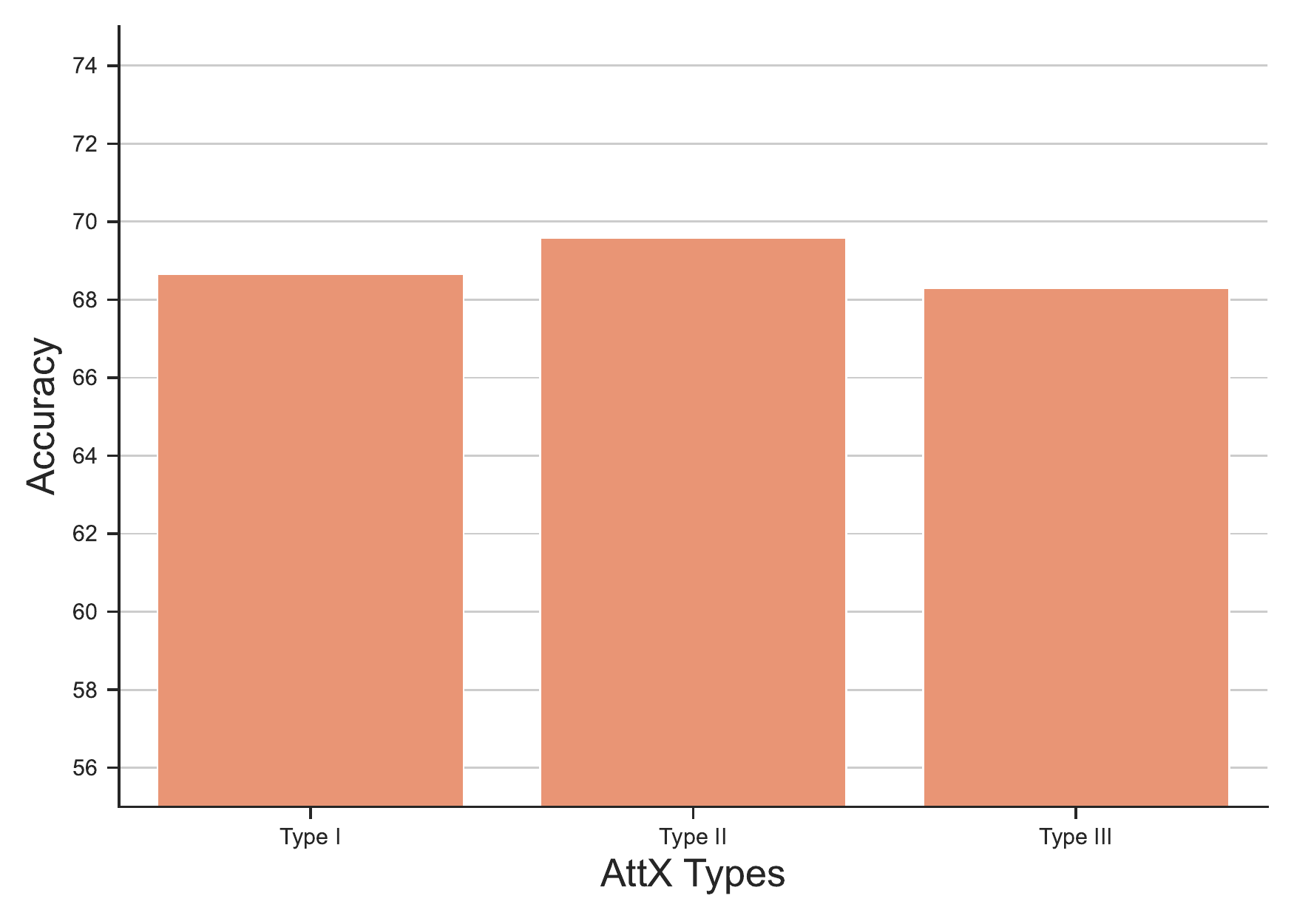}
    \includegraphics[width=0.49\linewidth]{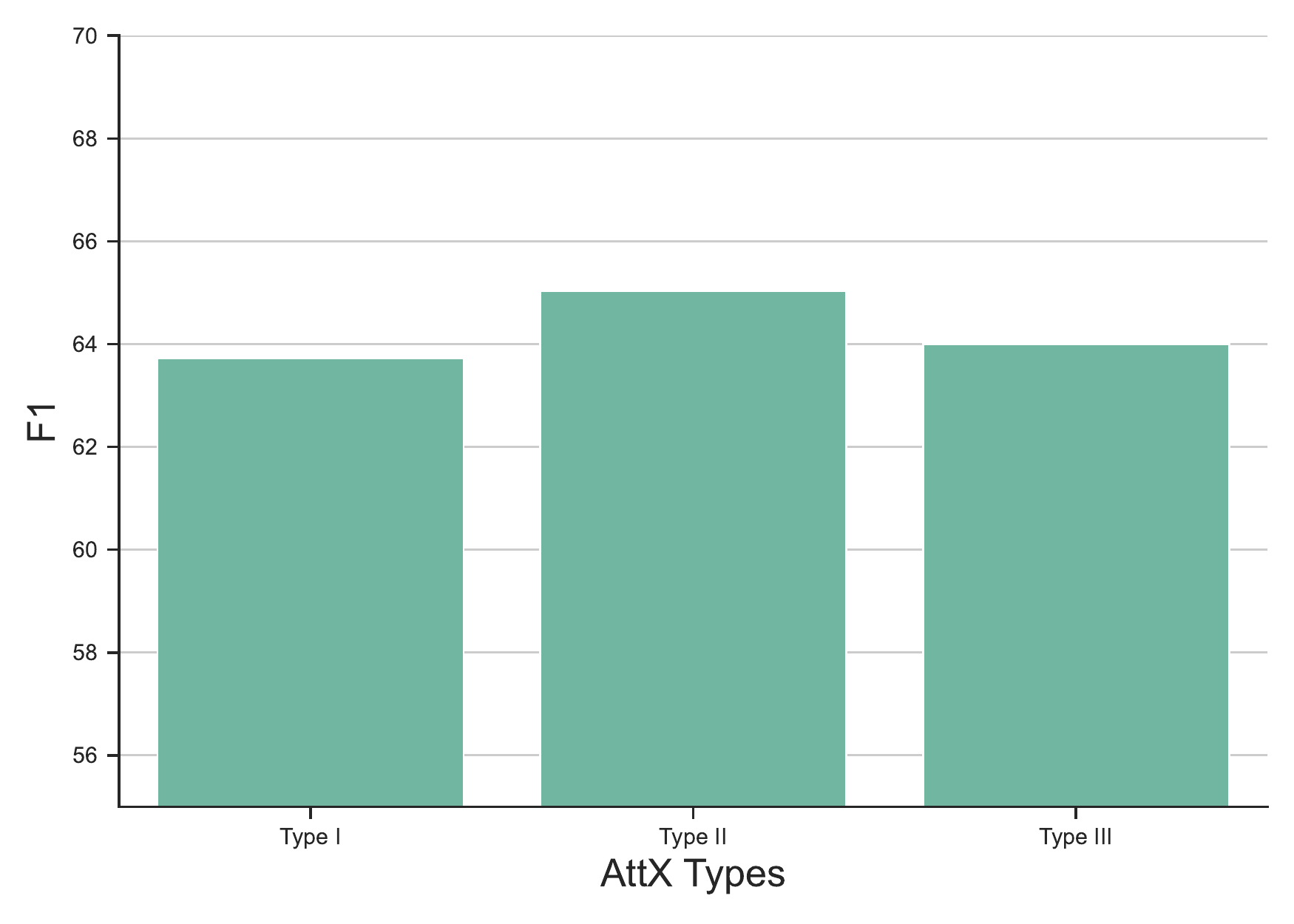}
    \caption{ResNet Pipeline}\label{fig:type_resnet1}
    \end{subfigure}    
    \caption{Plots for average Accuracy (left) and F1 (right) for different AttX types.}
\label{fig:best_type}
\end{figure}

\begin{figure}
    \centering
    \begin{subfigure}[t]{\linewidth}
    \centering
    \includegraphics[width=0.49\linewidth]{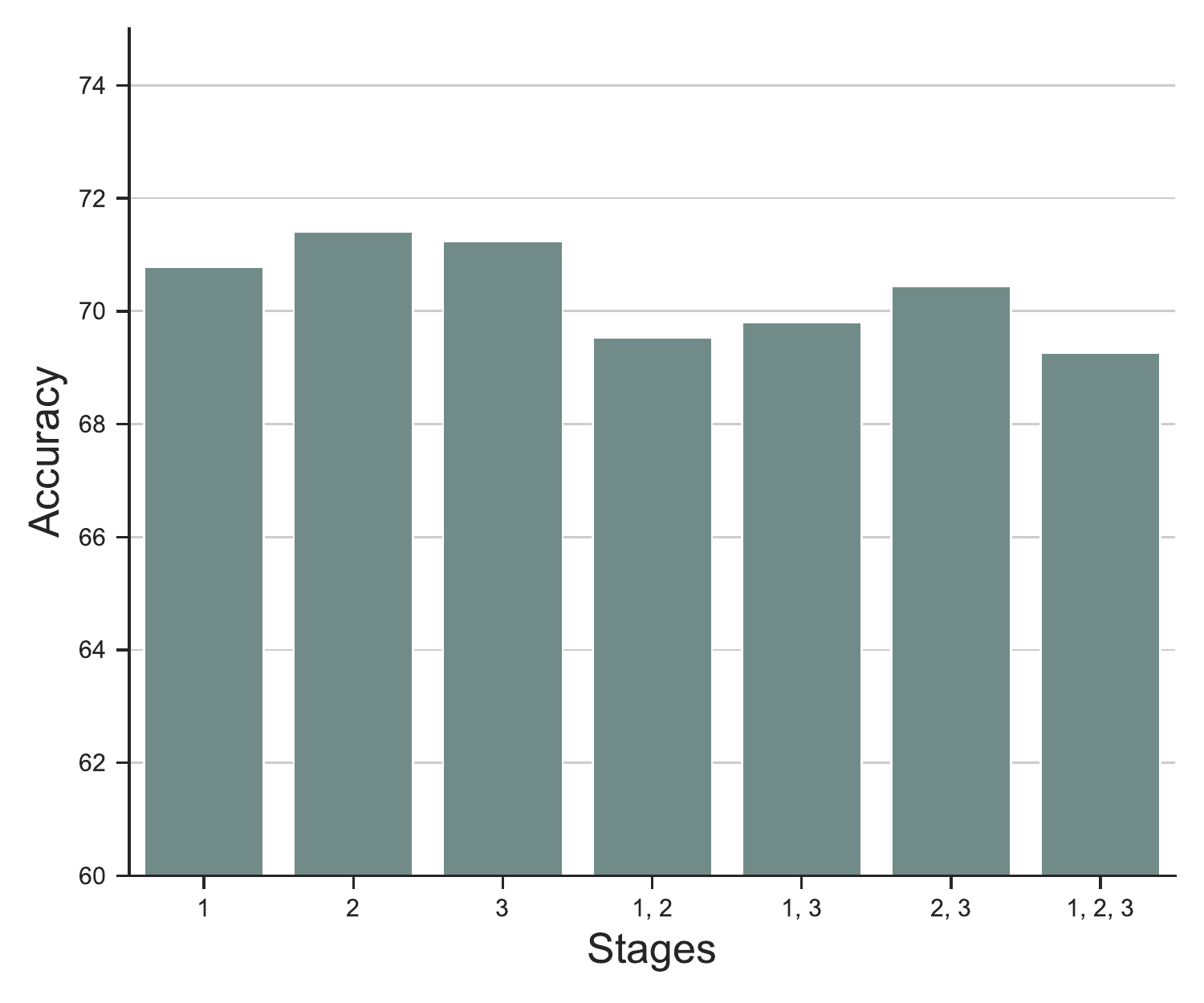}
    \includegraphics[width=0.49\linewidth]{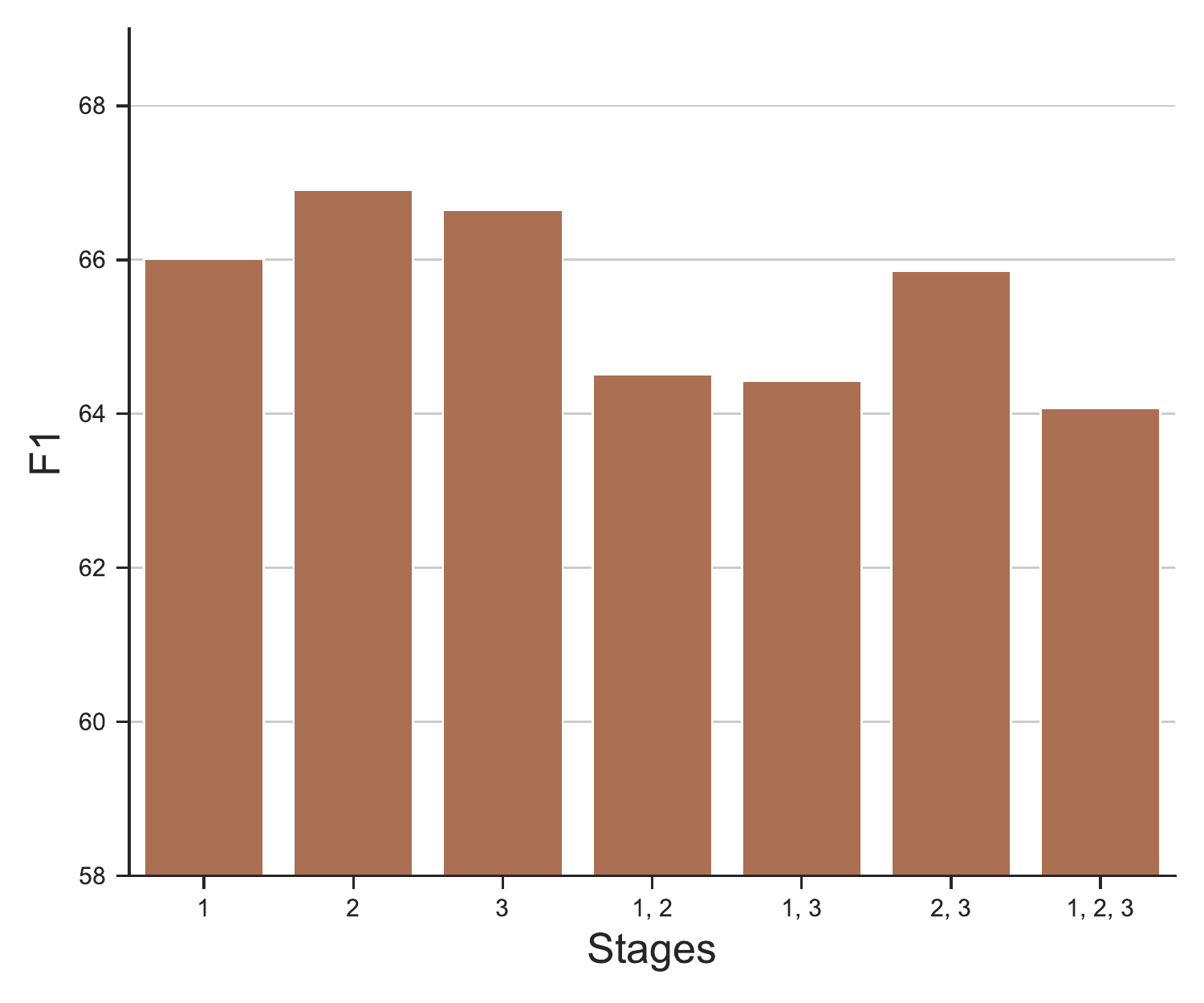}
    \caption{VGG Pipeline}\label{fig:type_vgg}
    \end{subfigure}
    \centering
    \begin{subfigure}[t]{\linewidth}
    \centering
    \includegraphics[width=0.49\linewidth]{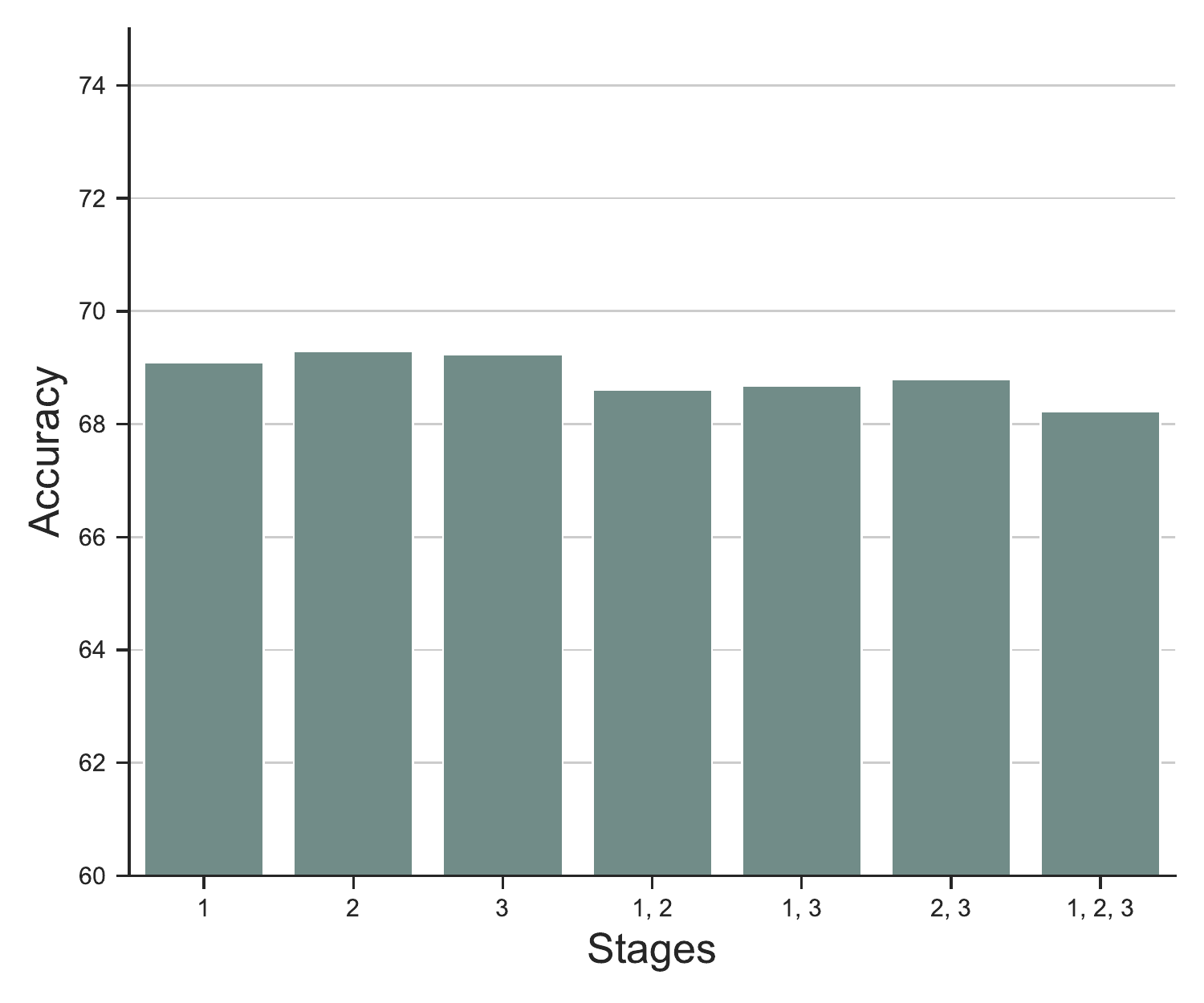}
    \includegraphics[width=0.49\linewidth]{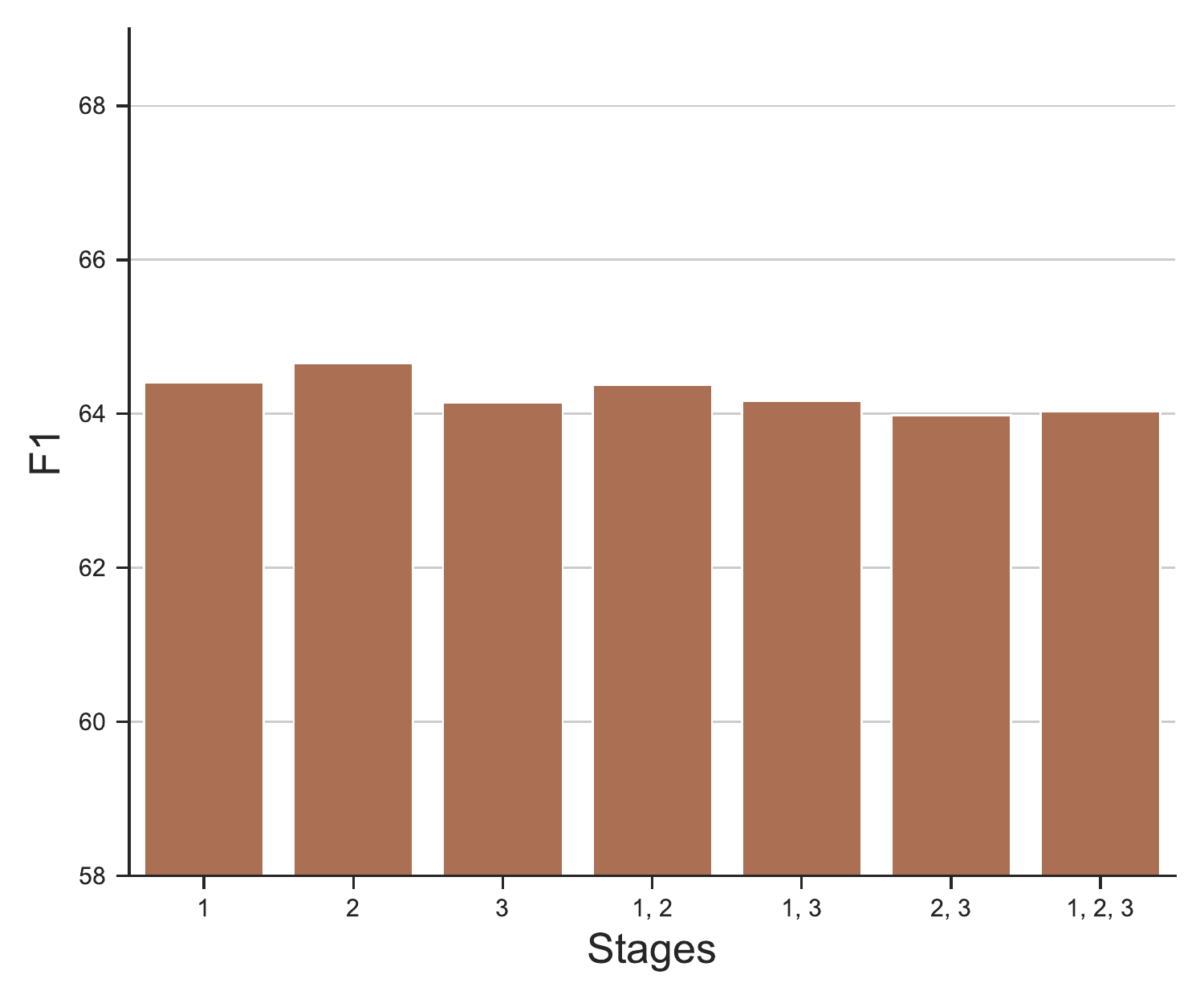}
    \caption{ResNet Pipeline}\label{fig:type_resnet}
    \end{subfigure}    
     \caption{Plots for average Accuracy (left) and F1 (right) for different stages where AttX has been integrated.}
     \label{fig:beststage}
\end{figure}

\subsection{Analysis}
To further explore the effectiveness of different AttX configurations and get a holistic sense of which Type of AttX connection works best at which stage(s), we merge the performance of all the models by taking the arithmetic mean of accuracy as well as f1 across all the datasets. To help us understand which stage works best with what type of AttX connection, we plot accuracy and f1-scores at each stage for Types I, II, and II in Figure \ref{fig:Types_bar}. Figure \ref{fig:type_1} shows that for Type I, fusing the modalities at stage 2 benefits the model the most; interestingly, fusion at stage 3 also yields comparable performance. Following the same trend, shown in Figure \ref{fig:type_2}, the ideal stage to fuse the modalities for Type II is stage 2. When embeddings from both the modalities are shared, i.e., Type III connection, it seems better to perform the fusion later, i.e., stage 3 (Figure \ref{fig:type_3}). Among the multi-instance configurations, we observe that for Types I and II, AttX connections at stages 2 \& 3 perform better than the other multi-AttX configurations, while for Type I, AttX connections at stages 1 \& 2 perform better. It is interesting to note that while the average performance of single-connections (AttX applied only in one stage of the network) is better than that of multi-stage, the best performances are generally achieved with multi-stage connections.

Figure \ref{fig:best_type} shows the AttX connection types and their accumulated performance for VGG and ResNet models across all the datasets. We observe that for the VGG encoder network, the Type II AttX connections outperform the other two types, i.e., I and III, by 2.72 \% and 2.76 \% in accuracy, and 3.2\% and 2.56\% in f1, respectively. For the ResNet encoder network, Type II AttX connections perform better than Type I and III by 0.9\% and 1.3\% in accuracy and 1.31\% and 1\% in f1, respectively. Accordingly, we can conclude that sharing information from EDA to ECG or BVP is more beneficial for the model, as opposed to the other direction or two-way connections.

Figure \ref{fig:beststage} shows the performance of all AttX connections at different stages accumulated over all datasets. The plot helps us understand the impact of adding AttX connections at different stages of the network and gives us more sense in identifying the optimum stage(s) to fuse the physiological modalities. The figure shows that our networks (VGG and ResNet) perform better when the fusion of the modalities occurs at stage 2, closely followed by stage 3. This indicates that \textit{early} attentive fusion of the modalities (stage) generally results in sub-optimal learning.

\begin{figure*}
    \centering
     \begin{subfigure}{0.19\textwidth}
         \includegraphics[width=1\textwidth]{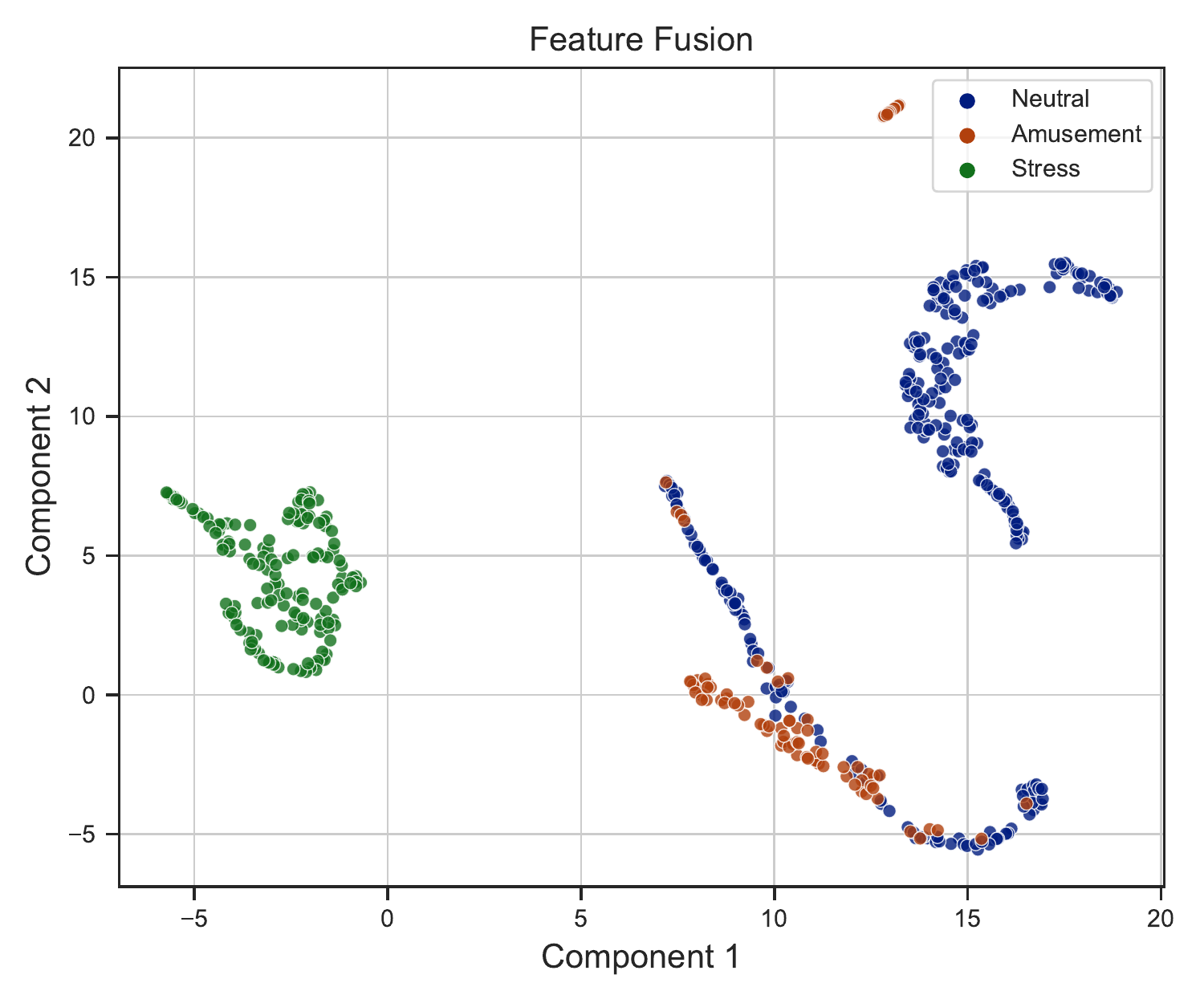}
         \label{fig:umapwesad3cFF}
     \end{subfigure}
     \begin{subfigure}{0.19\textwidth}
         \includegraphics[width=1\textwidth]{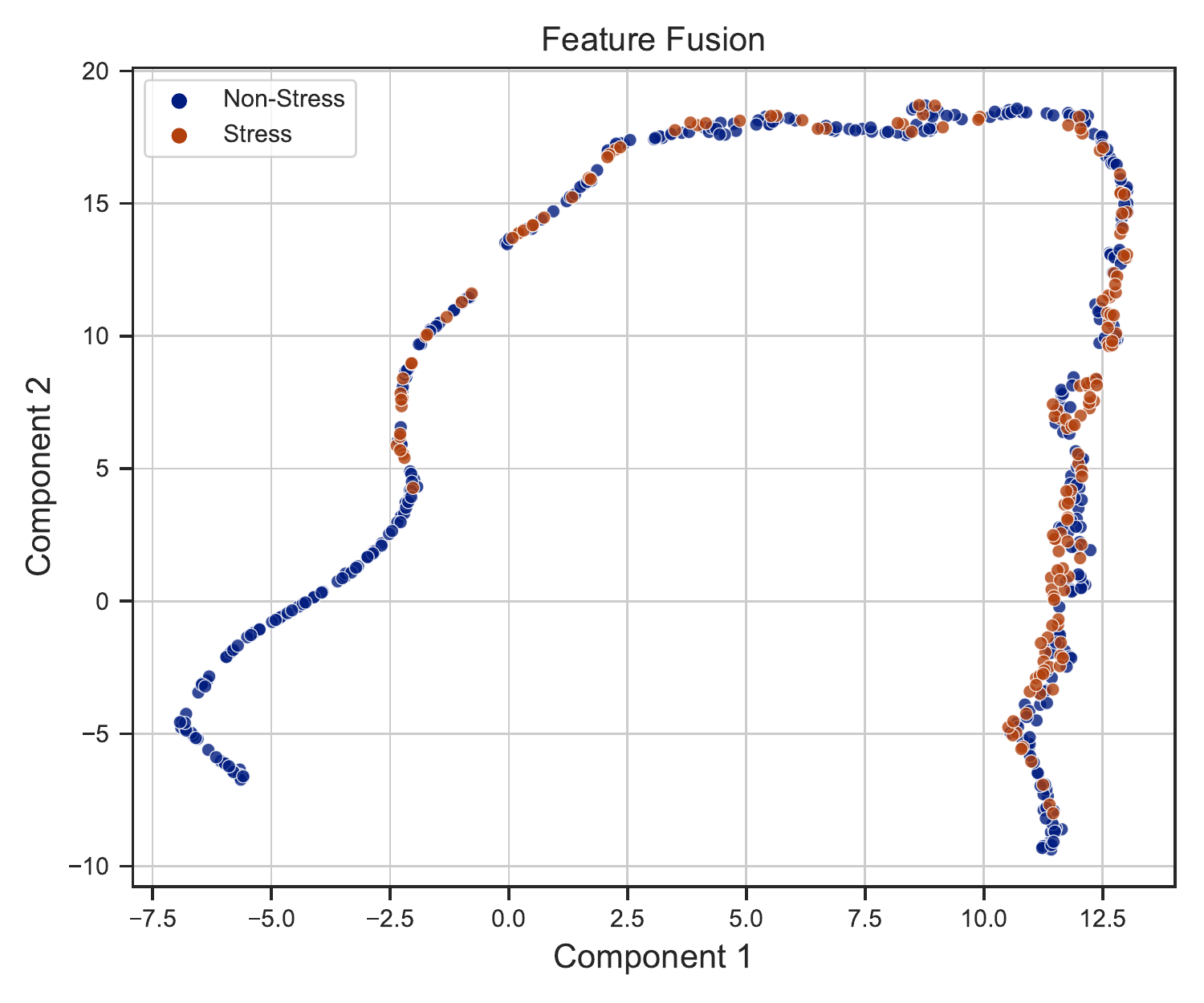}
         \label{fig:umapwesad2cFF}
     \end{subfigure}
     \begin{subfigure}{0.19\textwidth}
         \includegraphics[width=1\textwidth]{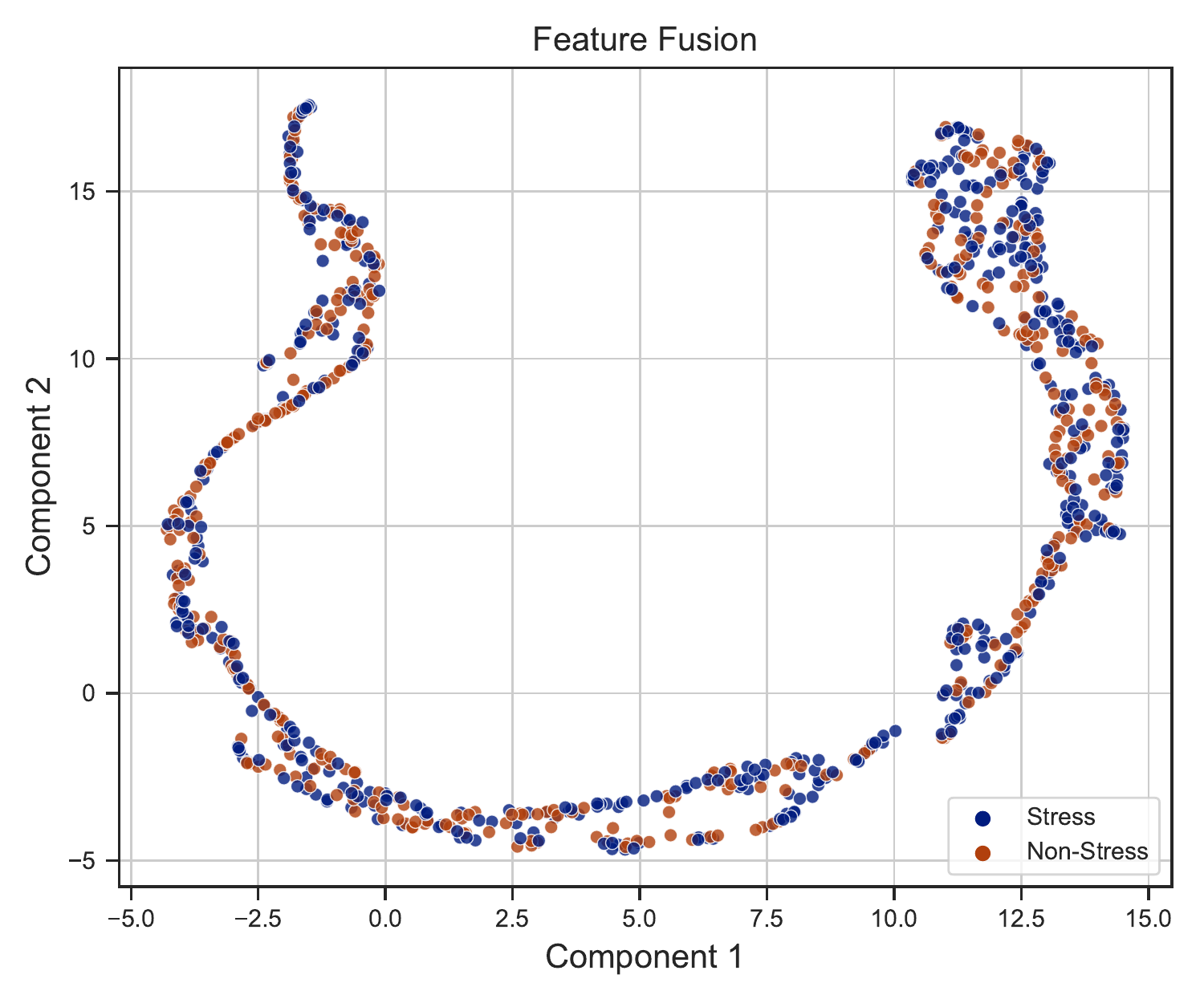}
         \label{fig:umapswell2cFF}
     \end{subfigure}
     \begin{subfigure}{0.19\textwidth}
         \includegraphics[width=1\textwidth]{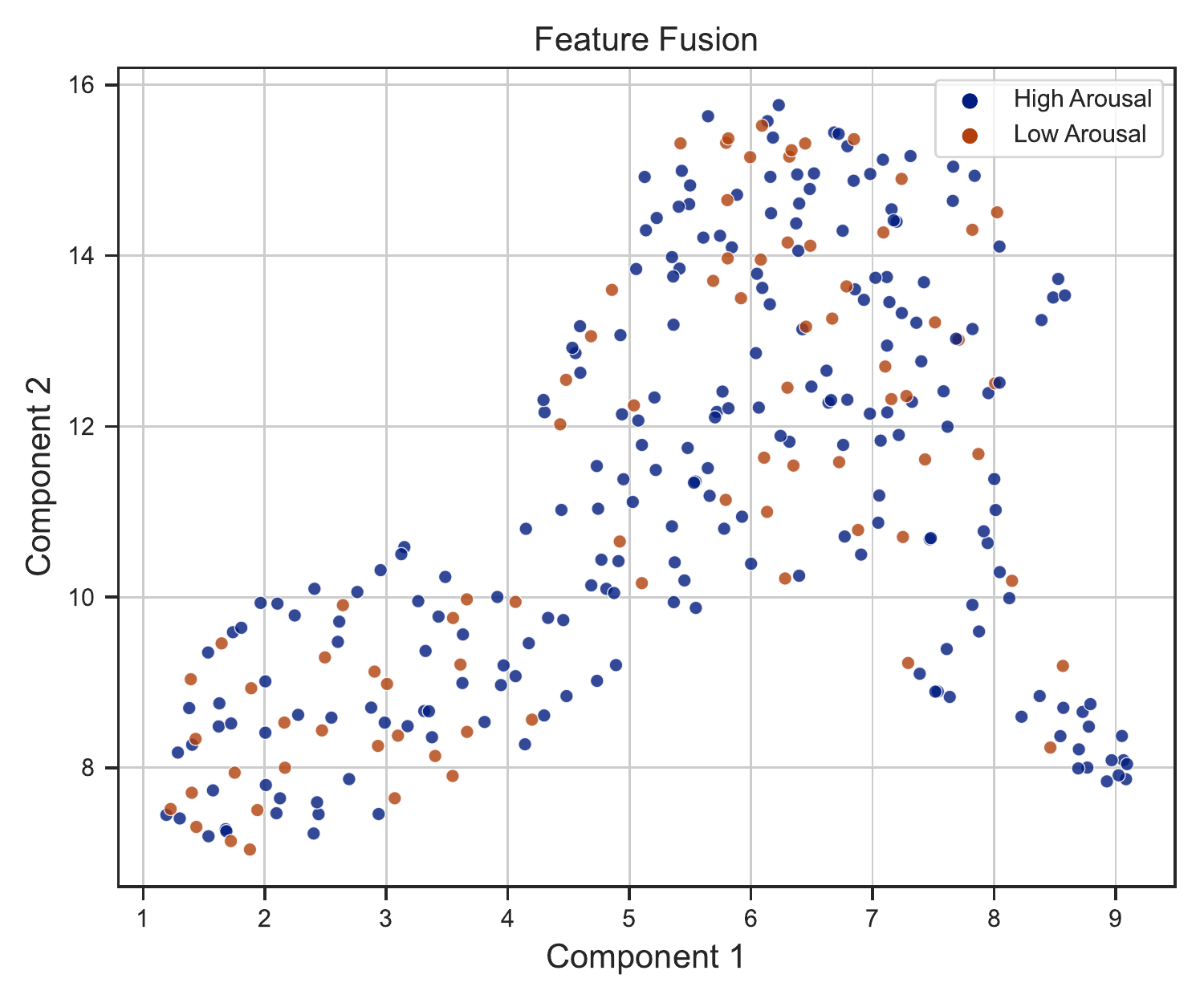}
         \label{fig:umapcase2cFF}
     \end{subfigure}
     \begin{subfigure}{0.19\textwidth}
         \includegraphics[width=1\textwidth]{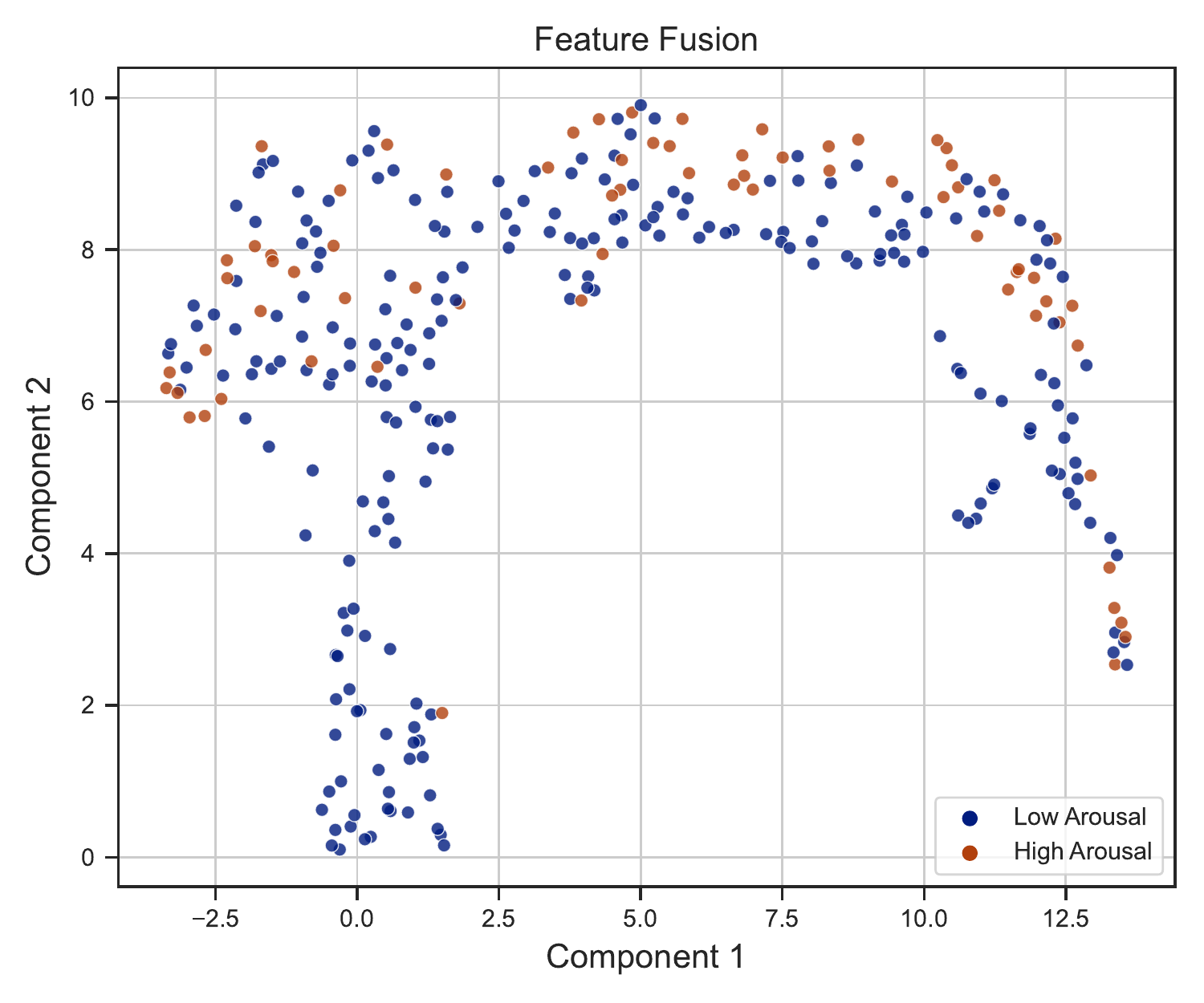}
         \label{fig:umapcase3cFF}
     \end{subfigure}
     \begin{subfigure}{0.19\textwidth}
         \includegraphics[width=1\textwidth]{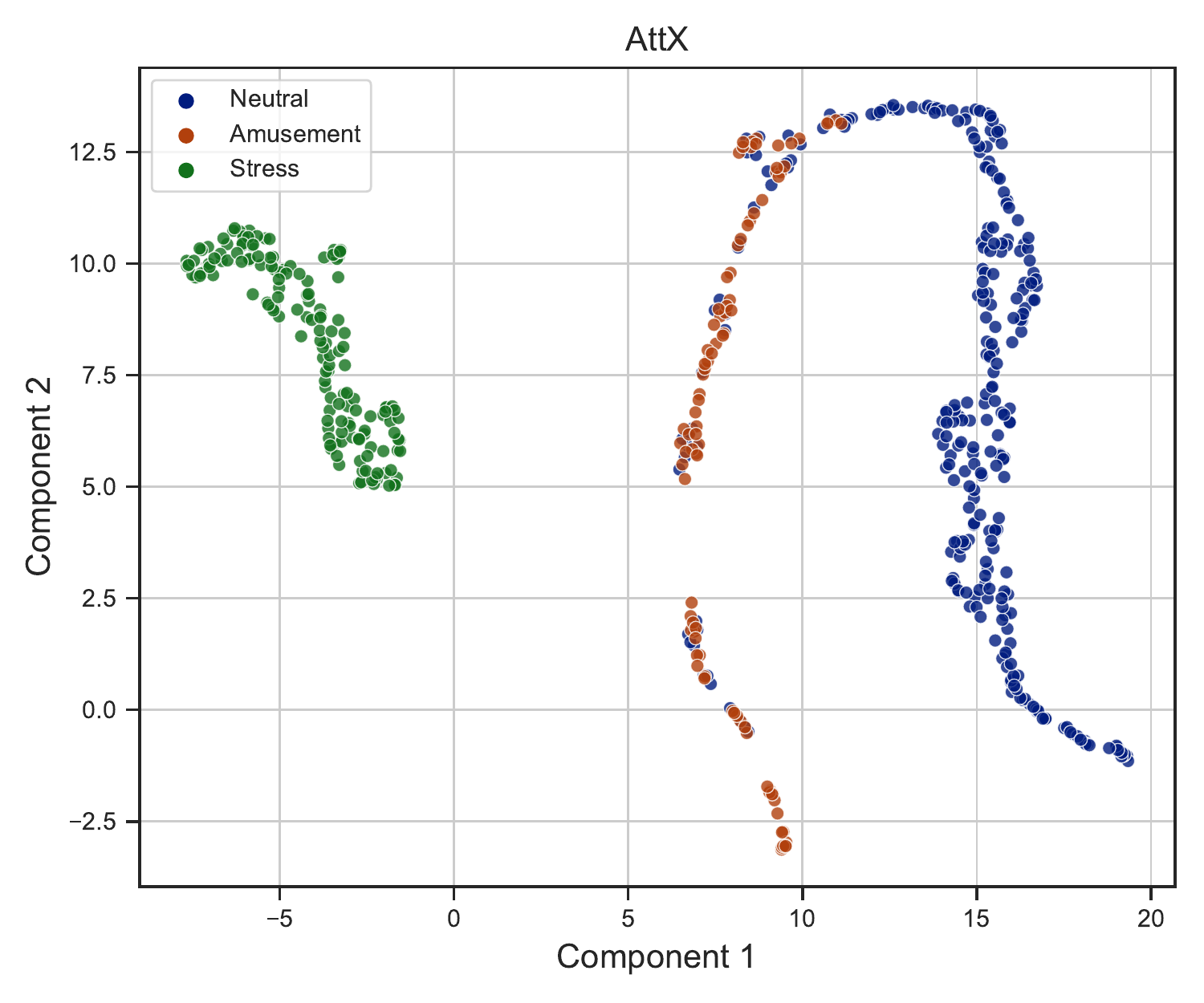}
         \caption{WESAD 3-Class}
         \label{fig:umapwesad3cAttX}
     \end{subfigure}
     \begin{subfigure}{0.19\textwidth}
         \includegraphics[width=1\textwidth]{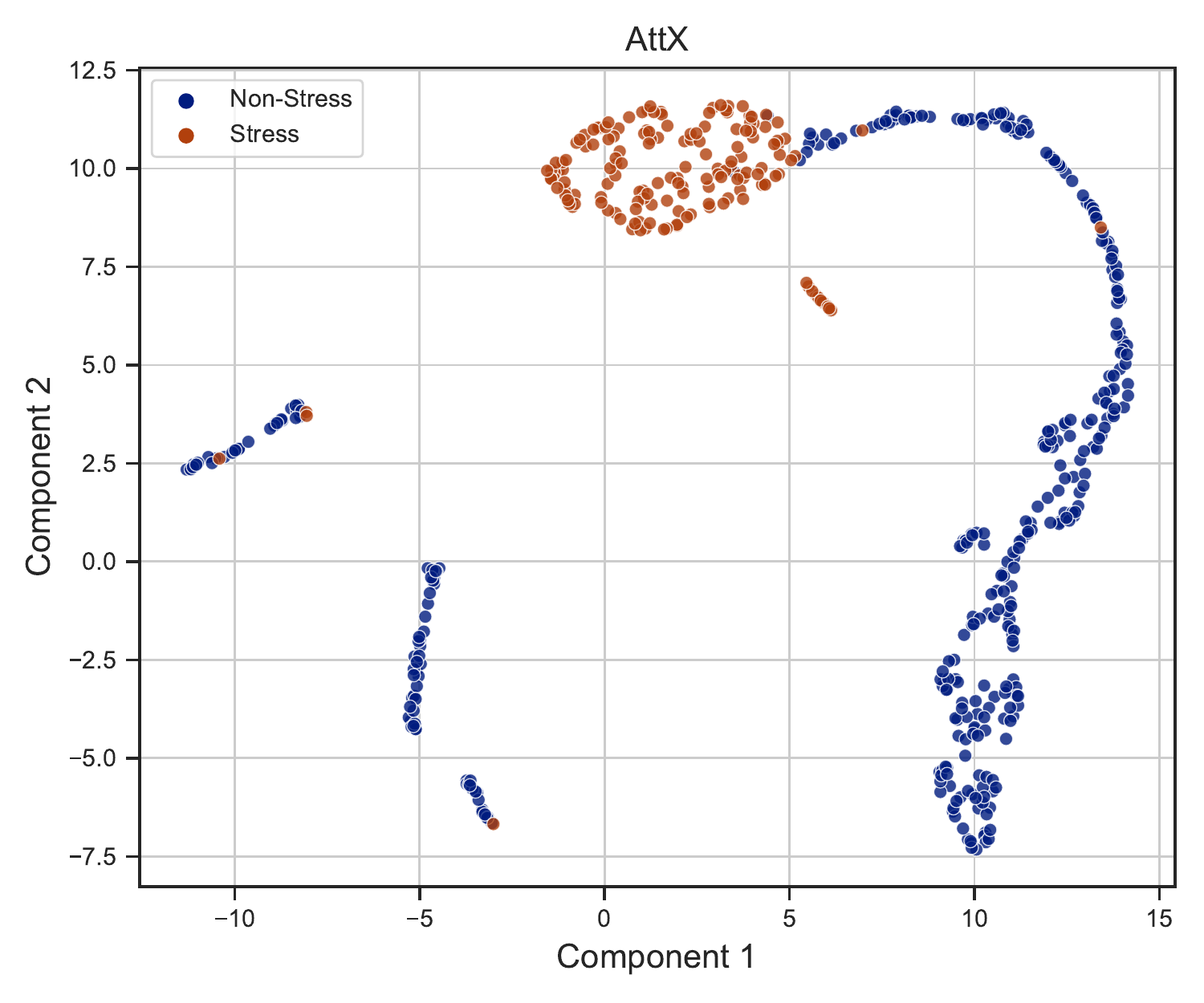}
         \caption{WESAD Binary}
         \label{fig:umapwesad2cAttX}
     \end{subfigure}
     \begin{subfigure}{0.19\textwidth}
         \includegraphics[width=1\textwidth]{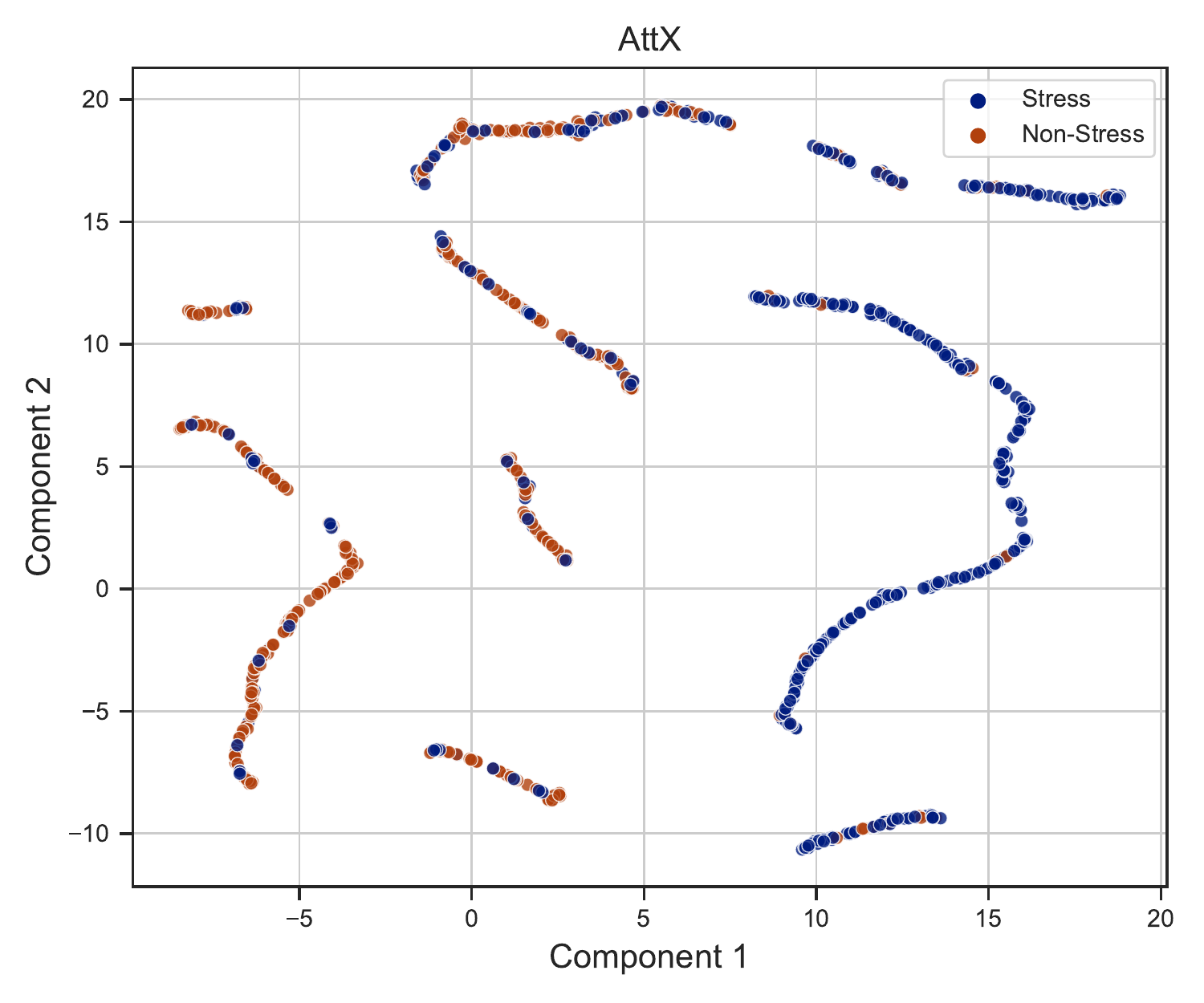}
         \caption{SWELL-KW}
         \label{fig:umapswell2cAttX}
     \end{subfigure}
     \begin{subfigure}{0.19\textwidth}
         \includegraphics[width=1\textwidth]{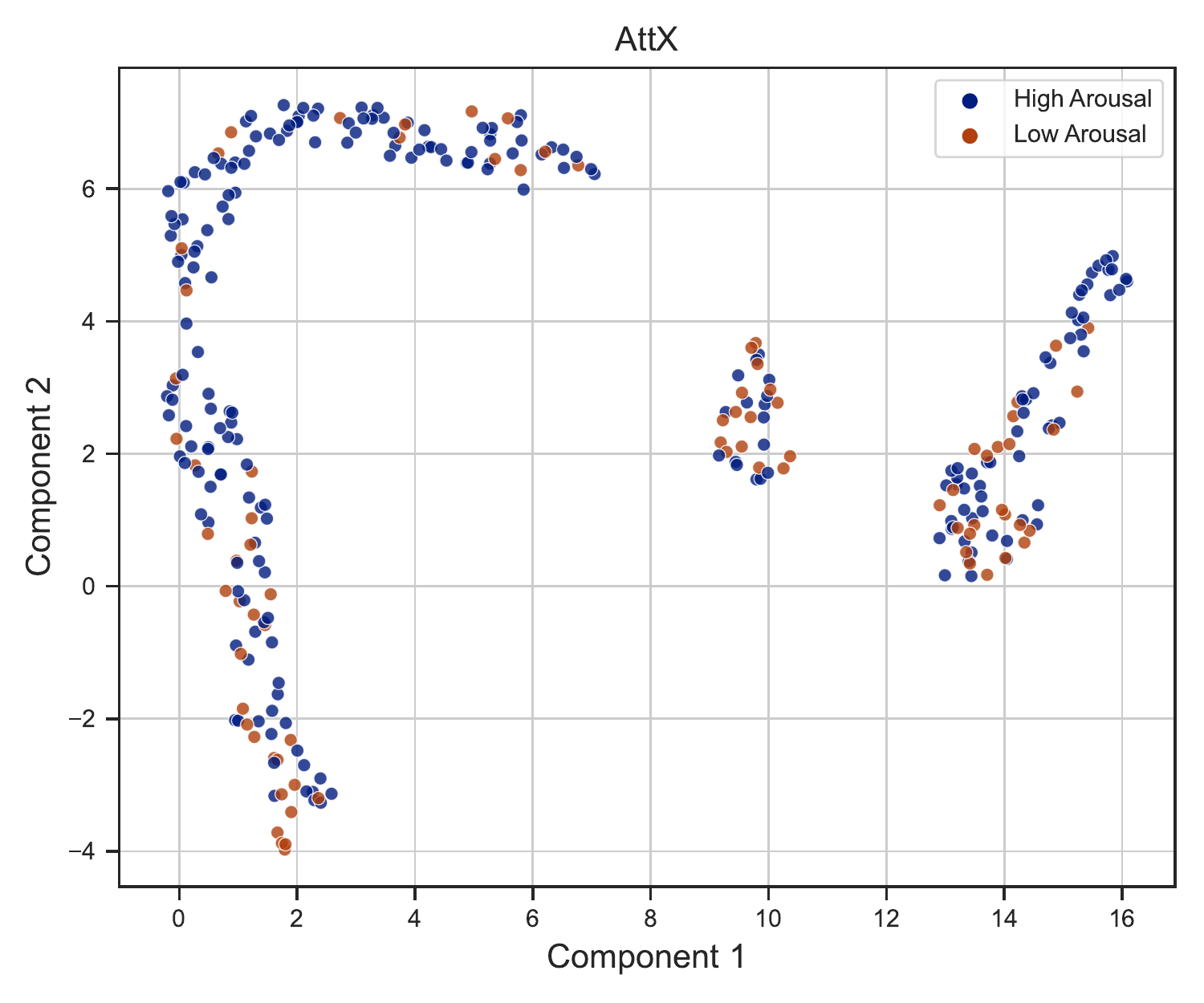}
         \caption{CASE (ECG and EDA)}
         \label{fig:umapcase2cAttX}
     \end{subfigure}
     \begin{subfigure}{0.19\textwidth}
         \includegraphics[width=1\textwidth]{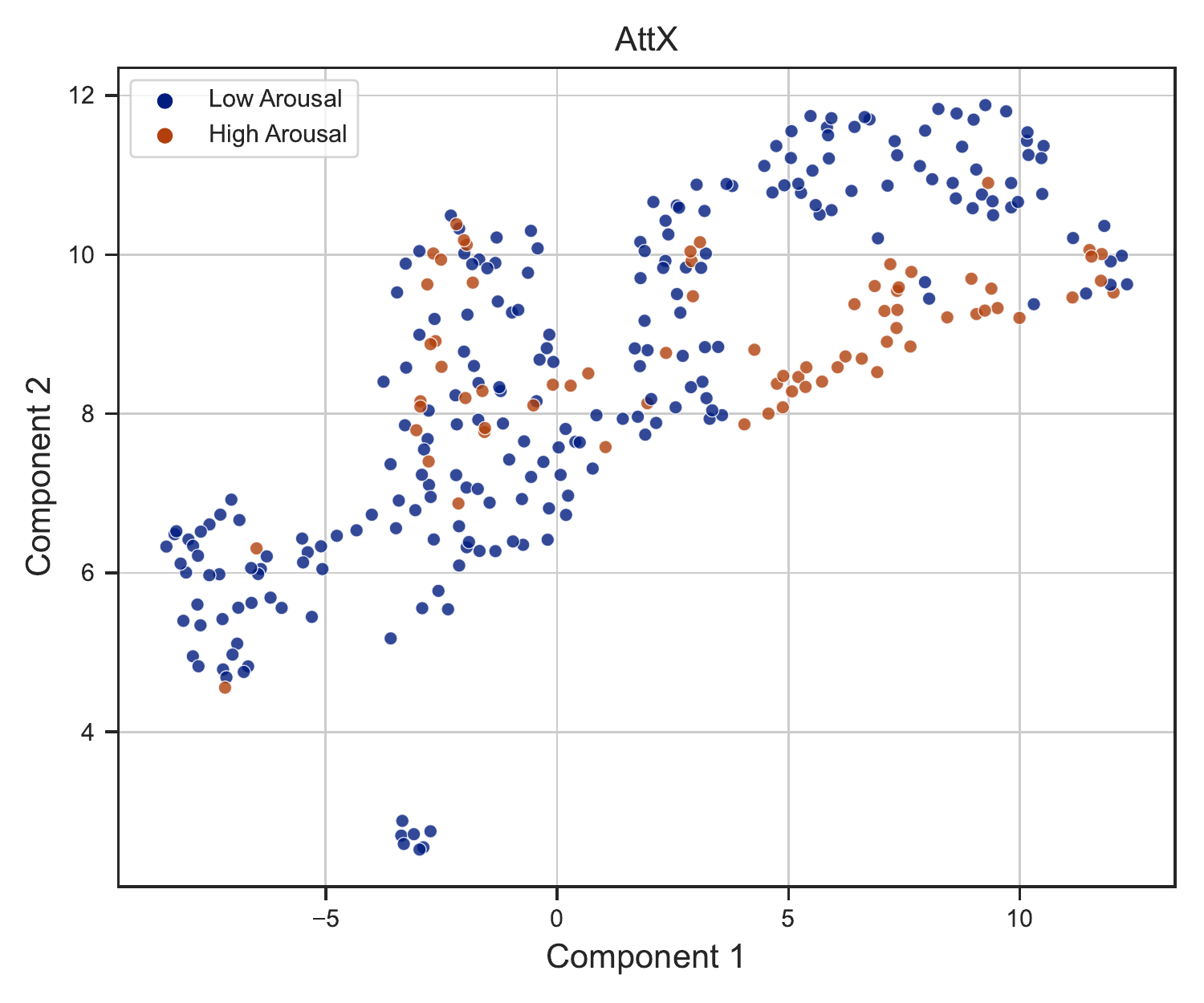}
         \caption{CASE (BVP and EDA)}
         \label{fig:umapcase3cAttX}
     \end{subfigure}
     \caption{Visualization of learned multimodal embeddings, using UMAP, when only feature fusion is used (first row) and when AttX connections are added to the network to share intermediate information (second row).}
     \label{fig:umap}
\end{figure*}

\subsection{Representation Visualization}
We explore the learned multimodal embeddings from our models with and without AttX connections to better understand the impact of sharing the intermediate information between pipelines. To visualize the high dimensional embedding space in 2D space, we utilize Uniform Manifold Approximation and Projection (UMAP) \cite{mcinnes2018umap} to perform the dimensionality reduction. Figure \ref{fig:umap} shows a comparison between learned multimodal embeddings when only feature fusion is used (first row) and when Attx connections are introduced in the network (second row). We observe that when feature fusion is performed, the learned embeddings from each class lie close to each other. However, by adding the AttX connections in the same setting, the embeddings from each class become more separable, which is a desired outcome as it helps the classifier classify each class better. 

\subsection{Comparison with the State-of-the-Art}
In this section, we compare the performance of our best-performing configurations on three datasets, i.e., WESAD (binary and 3-class problems), SWELL-KW, and CASE, to existing state-of-the-art methods and a number of baseline models. 

\subsubsection{WESAD}
Several recent studies have focused on using WESAD dataset to evaluate their methods. The works presented in \cite{schmidt2018introducing, bota2020emotion, lin2019explainable, sarkar2020self, li2020stress, behinaein2021transformer, mohammadi2022integrated, bacciu2021benchmarking, samyoun2020stress}, and \cite{gil2022human} use WESAD for emotion classification. However, the evaluation criteria used in \cite{sarkar2020self, li2020stress, mohammadi2022integrated, bacciu2021benchmarking} is not LOSO, which prevents a fair comparison.

{\renewcommand{\arraystretch}{1.4}
\begin{table}[!t]
\begin{center}
\caption{Classification results on WESAD Neutral vs. Amusement vs. Stress, in comparison to prior work using LOSO validation. The term `all chest modalities' refers to ECG, EDA, EMG, RESP, ACC, and ST.}
\setlength
\tabcolsep{4.5pt}
\label{tbl:wesad3c_results}
\scriptsize
\begin{tabular}{c|c|c|c|c}
\hline
\textbf{Ref.} & \textbf{Method}  & \textbf{Modality} & \textbf{Acc.} & \textbf{F1} \\
\hline \hline
\cite{schmidt2018introducing} & AdaBoost & All chest mod. & 80.34 & 72.51 \\\hline
\cite{hssayeni2021multi} & Gradient Boost & All chest mod. & 80.00 & 79.00 \\\hline
\cite{samyoun2020stress} & Random Forest & All chest mod. & 80.50 & 67.10 \\\hline
\cite{gil2022human} & CNN & All chest mod. & 81.87 & 81.21 \\ \hline
\cite{lin2019explainable} & CNN & All chest mod. & 83.00 & 81.00\textsuperscript{$\ddagger$} \\ \hline

\hline
\multirow{8}{*}{Baselines} & VGG & ECG & 56.62 & 46.00\\ \cline{2-5}
 & VGG & EDA & 79.23 & 73.16 \\ \cline{2-5}
 & ResNet & ECG & 45.60 & 36.63 \\ \cline{2-5}
 & ResNet & EDA  & 71.01 & 65.15\\ \cline{2-5}
 & VGG (feat. fus.) & ECG, EDA & 74.36 & 67.00\\ \cline{2-5}
 & VGG (score fus.) & ECG, EDA & 74.27 & 65.70 \\ \cline{2-5}
 & ResNet (feat. fus.) & ECG, EDA & 74.70 & 69.58\\ \cline{2-5}
 & ResNet (score fus.) & ECG, EDA & 67.68 & 52.56 \\ \hline
 
\hline
\multirow{3}{*}{Ours} & VGG Type II [1,2,3] & ECG, EDA & 81.57 & 75.68 \\ \cline{2-5}
 & VGG Type II$^\dagger$ [1,2,3]* & ECG, EDA, RESP, ST & 89.57 & 82.57 \\ \cline{2-5}
 & VGG Type II$^\dagger$ [1,2,3] & ECG, EDA, RESP, ST & 83.30 & 76.12 \\ \cline{2-5}
\hline 
\end{tabular}
\end{center}
\footnotesize{*Evaluation criteria according to \cite{lin2019explainable}.\\ \textsuperscript{$\ddagger$}Reported Weighted F1-score instead of Macro F1-score. \\Type II$^\dagger$ represents cross-connections from EDA, RESP, and ST to ECG.}
\end{table}}

For 3-class problem, several prior works \cite{ schmidt2018introducing, gil2022human, lin2019explainable} have used all the following modalities, ECG, EDA, EMG, RESP, ACC, and ST, which they collectively refer to as `chest modalities'. When standard LOSO evaluation is used, Table \ref{tbl:wesad3c_results} shows that our model with two modalities, i.e., ECG and EDA, can outperform \cite{schmidt2018introducing, hssayeni2021multi, samyoun2020stress} by achieving an accuracy of 81.57 and an f1 of 75.68. To compare our method with \cite{lin2019explainable}, we follow the same evaluation criteria used in this work and achieve an accuracy of 89.57 and an f1-score of 82.57. Further, when RESP and ST are fused as the third and the fourth modalities in our network, we achieve accuracy and f1 of 83.30 and 76.12, which is competitive to \cite{gil2022human}. We also compare our results with uni-modal baselines using ECG and EDA, and traditional fusion techniques such as feature fusion and score fusion. We show that when AttX connections are added, there is a performance boost of approximately 7\% with respect to feature fusion and score fusion. Table \ref{tbl:wesad2c_results} presents the performance of our model for stress vs. non-stress classification task in comparison to state-of-the-art methods, uni-modal baselines, feature fusion, and score fusion. The table shows that when only two modalities are used, our method achieves accuracy and f1 of 92.90 and 91.73, respectively, outperforming \cite{behinaein2021transformer, holder2022comparing, bota2020emotion}. Including RESP with ECG and EDA improves the model performance by giving a boost of approximately 1\% in accuracy and 1.5\% in f1, outperforming \cite{schmidt2018introducing, gil2022human}. When compared to feature fusion and score fusion, we observe that by adding AttX connections, the performance of the model is increased by approximately 5\% and 0.8\%, respectively.

{\renewcommand{\arraystretch}{1.4}
\begin{table}[!t]
\begin{center}
\caption{Classification results on WESAD Stress vs. Non-Stress, in comparison to prior work using LOSO validation. Th term `all chest modalities' denotes ECG, EDA, EMG, RESP, ACC, and ST, while the term `all modalities' includes additional modalities from the wrist.}
\setlength
\tabcolsep{4.5pt}
\scriptsize
\label{tbl:wesad2c_results}
\scriptsize
\begin{tabular}{c|c|c|c|c}
\hline
\textbf{Ref.} & \textbf{Method}  & \textbf{Modality} & \textbf{Acc.} & \textbf{F1}\\
\hline \hline
\cite{behinaein2021transformer} & CNN & ECG & 80.40 & 69.70 \\\hline
\multirow{2}{*}{\cite{holder2022comparing}} & CNN & EDA & 91.67 & 82.29 \\ \cline{2-5}
& CNN & All modalities & 83.28 & 67.53 \\ \hline
\multirow{2}{*}{\cite{bota2020emotion}} & QDA (DF) & EDA, ECG, BVP, RSP & 85.80  & -- \\ \cline{2-5}
& QDA (FF) & EDA, ECG, BVP, RSP & 87.60 & 19.40 \\ \hline
\cite{samyoun2020stress} & Extra Tree & All chest mod. & 91.10 & 90.20 \\ \hline
\cite{schmidt2018introducing} & LDA & All chest mod. & 93.12 & 91.47 \\ \hline
\cite{gil2022human} & CNN & All chest mod. & 93.20 & 92.70 \\ \hline 

\hline 
\multirow{8}{*}{Baselines} & VGG & ECG & 86.54 & 84.93\\\cline{2-5} 
 & VGG & EDA & 87.32 & 86.76 \\ \cline{2-5} 
 & ResNet & ECG & 85.54 & 83.93 \\ \cline{2-5}
 & ResNet & EDA & 84.32 & 80.76\\ \cline{2-5}
 & VGG (feat. fus.) & ECG, EDA & 87.52 & 86.54\\ \cline{2-5}
 & VGG (score fus.) & ECG, EDA & 92.02 & 90.02\\ \cline{2-5}
 & ResNet (feat. fus.) & ECG, EDA & 86.00 & 84.04\\ \cline{2-5}
 & ResNet (score fus.) & ECG, EDA & 85.19 & 81.56\\\hline
 
\hline 

\multirow{2}{*}{Ours} & VGG Type II [2,3] &	ECG, EDA & 92.90 & 91.73 \\ \cline{2-5}
 & VGG Type II$^\dagger$ [2,3] & ECG, EDA, RESP & 93.70 & 93.28 \\
\hline 
\end{tabular}
\end{center}
\footnotesize{Type II$^\dagger$ represents cross-connections from EDA to ECG, RESP.}
\end{table}}

\subsubsection{SWELL-KW}
The works presented in \cite{koldijk2016detecting, behinaein2021transformer, masood2019modeling, sarkar2020self, liakopoulos2021cnn, motogna2021strategy, walambe2021employing}, and \cite{iqbal2022exploring} use SWELL-KW for classification of workplace stress. However, studies \cite{masood2019modeling, sarkar2020self, liakopoulos2021cnn, motogna2021strategy, walambe2021employing} use a different cross-validation scheme that prevents them from evaluating their method on unseen data. So, to have a fair comparison, we compare our results with studies \cite{ koldijk2016detecting, behinaein2021transformer} as they use the LOSO validation scheme. Table \ref{tbl:swell_results} presents the performance of our model in comparison to the uni-modal baselines, feature fusion, score fusion and state-of-the-art methods. The table shows that our best configuration of AttX connection with only ECG and EDA fusion achieves an accuracy of 65.20 and an f1 of 62.25, outperforming the state-of-the-art method \cite{koldijk2016detecting} by approximately 6.3\%. Also, we show that AttX connections perform better than the feature fusion and score fusion techniques by 4\% and 4.8\%, respectively.

{\renewcommand{\arraystretch}{1.4}
\begin{table}[!t]
\begin{center}
\caption{Classification results on SWELL-KW Stress vs. Non-Stress, in comparison to prior work using LOSO validation.}
\setlength
\tabcolsep{4.5pt}
\label{tbl:swell_results}
\scriptsize
\begin{tabular}{c|c|c|c|c}
\hline
\textbf{Ref.} & \textbf{Method}  & \textbf{Modality} & \textbf{Acc.} & \textbf{F1}\\
\hline \hline
\cite{koldijk2016detecting} & SVM & - & 58.90 & -- \\ \hline
\cite{behinaein2021transformer} & CNN + Transformer & ECG & 58.10 & 58.50 \\ \hline

\hline
\multirow{8}{*}{Baselines} & VGG & ECG & 58.19 & 47.75\\ \cline{2-5}
 & VGG & EDA & 60.33 & 58.07 \\  \cline{2-5}
 & ResNet & ECG & 52.65 & 46.72 \\ \cline{2-5}
 & ResNet & EDA & 59.03 & 54.23\\\cline{2-5}
 & VGG (feat. fus.) & ECG, EDA & 61.07 & 56.94\\\cline{2-5}
 & VGG (score fus.) & ECG, EDA & 60.34 & 50.52 \\\cline{2-5}
 & ResNet (feat. fus.) & ECG, EDA & 57.30 & 52.10\\\cline{2-5}
 & ResNet (score fus.) & ECG, EDA & 53.00 & 47.19 \\ \hline
 
\hline
Ours & VGG Type II [1] & ECG, EDA & 65.20 & 62.25 \\
\hline 
\end{tabular}
\end{center}
\end{table}}

{\renewcommand{\arraystretch}{1.4}
\begin{table}[!t]
\begin{center}
\caption{Classification results on CASE Arousal, in comparison to a number of baselines using LOSO validation (no prior works with LOSO were found). }
\setlength
\tabcolsep{4.5pt}
\scriptsize
\begin{tabular}{c|c|c|c|c}
\hline
\textbf{Ref.} & \textbf{Method}  & \textbf{Modality} & \textbf{Acc.} & \textbf{F1}\\
\hline \hline

\multirow{14}{*}{Baselines} & VGG & ECG & 50.44 & 47.38\\\cline{2-5}
 & VGG & EDA & 56.25 & 52.11 \\ \cline{2-5}
 & ResNet & ECG & 59.45 & 54.48 \\ \cline{2-5}
 & ResNet & EDA & 59.63 & 56.69\\\cline{2-5}
 & VGG & BVP & 53.34 & 50.93 \\ \cline{2-5}
 & ResNet & BVP & 60.48 & 53.26\\\cline{2-5}
 & VGG (feat. fus.) & ECG, EDA & 63.02 & 59.11\\\cline{2-5}
 & VGG (score fus.) & ECG, EDA & 52.20 & 49.21 \\\cline{2-5}
 & ResNet (feat. fus.) & ECG, EDA & 63.53 & 57.78\\\cline{2-5}
 & ResNet (score fus.) & ECG, EDA & 60.28 & 55.75 \\\cline{2-5}
 & VGG (feat. fus.) & BVP, EDA & 66.09 & 60.05\\\cline{2-5}
 & VGG (score fus.) & BVP, EDA & 52.57 & 49.36 \\\cline{2-5}
 & ResNet (feat. fus.) & BVP, EDA & 63.95 & 61.00\\\cline{2-5}
 & ResNet (score fus.) & BVP, EDA & 60.78 & 55.53 \\\hline
 
\hline
\multirow{4}{*}{Ours} & VGG Type II [1,2,3] & ECG, EDA & 66.72 & 61.00\\ \cline{2-5} 
& VGG Type II [2, 3] & BVP, EDA & 67.48 & 62.16 \\ \cline{2-5} 
& VGG Type II$^\dagger$ [2,3] & ECG, EDA, ST & 70.15 & 67.84 \\ \cline{2-5} 
& VGG Type II$^\dagger$ [2,3] & BVP, EDA, ST & 71.00 & 68.89 \\ \cline{2-5} 
\hline 
\end{tabular}
\label{tbl:case_results}
\end{center}
\footnotesize{Type II$^\dagger$ represents cross-connections from EDA, ST to ECG or BVP.}
\end{table}}

\subsubsection{CASE}
For the CASE dataset, we present the performance of our model using a combination of ECG, EDA, BVP, and ST in Table \ref{tbl:case_results}. We compare the performance of our model with the uni-model baselines, feature fusion, and score fusion techniques. No comparison to state-of-the-art is performed as seminal works on this dataset were not published at the time when our study was in progress. The original paper, which introduced and accompanied the dataset \cite{sharma2019dataset} also did not provide any baseline values. The table shows that using fusion techniques such as feature and score fusion, the performance of the model is increased compared to the uni-modal models of modalities ECG, EDA, and BVP. We observe that adding AttX connections further increases the performance. For instance, our model with ECG and EDA performs 3.7\% and 14\% better than the feature and the score fusion, respectively. Similarly, our model with BVP and EDA performs 1.4\% and 15\% better than feature and score fusion, respectively. Further, we show that adding ST to ECG and EDA or BVP and EDA enhances the performance of our model, achieving accuracy and f1 of 70.15 and 67.84 for ECG and EDA and 71.00 and 68.89 for BVP and EDA, respectively.

\section{Conclusion}
\label{sec:conclusion}
In this paper, we propose a novel attentive cross-modal connection, AttX, for learning representation from multimodal wearable data. These connections comprise an attentive feedforward network that can be integrated at any stage of the pipeline to create intermediate connections between individual streams for processing each modality. These intermediate cross-connections share weighted information (based on the importance) between the modality streams. To control the flow of information, we introduce different types of AttX connections that can share intermediate information in uni- and bi-directional formats. We perform extensive experiments to demonstrate the effectiveness of our proposed solution on three publicly available datasets using different backbones networks. The experiments show that sharing information from EDA to ECG or BVP enhances the model performance the most. Also, the optimum stage to fuse modalities is stage 2 (mid-fusion), closely followed by stage 3. Our experiments establish that the proposed method achieves superior or competitive performance by integrating AttX connections into simple CNN-based multimodal solutions compared to baseline uni-model, classical multimodal, and state-of-the-art methods. 

\section*{Acknowledgment}
The authors would like to thank Innovation for Defence Excellence and Security (IDEaS) program for funding this project. The authors would also like to thank Dirk Rodenburg for his help throughout the project.

\ifCLASSOPTIONcaptionsoff
  \newpage
\fi

\bibliographystyle{IEEEtran}
\small{
\bibliography{references}
}
\end{document}